\documentclass{article}
\usepackage{arxiv}
\usepackage[utf8]{inputenc} 
\usepackage[T1]{fontenc}    
\usepackage{hyperref}       
\usepackage{url}            
\usepackage{booktabs}       
\usepackage{amsfonts}       
\usepackage{nicefrac}       
\usepackage{microtype}      
\usepackage{lipsum}
\usepackage{enumitem}
\usepackage{graphicx}
\graphicspath{ {./images/} }

\usepackage[utf8]{inputenc}
\usepackage[T1]{fontenc}

\usepackage{csquotes}
\usepackage{amsmath}
\usepackage{bm}
\usepackage{amssymb}

\usepackage{todonotes}
\usepackage{tikz}
\usepackage{xcolor}
\usetikzlibrary{fit}
\usetikzlibrary{arrows.meta, positioning}
\usetikzlibrary{decorations.pathreplacing}
\usepackage{dirtytalk}
\usepackage{makecell}
\usepackage{subcaption}
\usepackage{authblk}

\title{When Counterfactual Reasoning Fails: Chaos and Real-World Complexity}

\author[1,3]{\textbf{Yahya Aalaila}}
\author[1]{\textbf{Gerrit Gro\ss{}mann}}
\author[1]{\textbf{Sumantrak Mukherjee}}
\author[2]{\textbf{Jonas Wahl}}
\author[1]{\textbf{Sebastian Vollmer}}

\affil[1]{ German Research Center for Artificial Intelligence (DFKI), Data Science and its Applications Research Group, Kaiserslautern, Germany}
\affil[2]{German Research Center for Artificial Intelligence (DFKI), Research Department Neuro-Mechanistic Modeling, Saarbrücken, Germany}
\affil[3]{Mohammed VI Polytechnic University, UM6P College of Computing, Benguerir, Morocco}
\affil[]{\textbf{Emails:} 
\texttt{\{yahya.aalaila, gerrit.grossmann, sumantrak.mukherjee, jonas.wahl, sebastian.vollmer\}@dfki.de},  \texttt{yahya.aalaila@um6p.ma}}

\renewcommand{\today}{}
\begin{document}
\maketitle
\begin{abstract}
Counterfactual reasoning, a cornerstone of human cognition and decision-making, is often seen as the 'holy grail' of causal learning, with applications ranging from interpreting machine learning models to promoting algorithmic fairness. While counterfactual reasoning has been extensively studied in contexts where the underlying causal model is well-defined, real-world causal modeling is often hindered by model and parameter uncertainty, observational noise, and chaotic behavior. The reliability of counterfactual analysis in such settings remains largely unexplored. In this work, we investigate the limitations of counterfactual reasoning within the framework of Structural Causal Models. Specifically, we empirically investigate \emph{counterfactual sequence estimation} and highlight cases where it becomes increasingly unreliable. We find that realistic assumptions, such as low degrees of model uncertainty or chaotic dynamics, can result in counterintuitive outcomes, including dramatic deviations between predicted and true counterfactual trajectories. This work urges caution when applying counterfactual reasoning in settings characterized by chaos and uncertainty. Furthermore, it raises the question of whether certain systems may pose fundamental limitations on the ability to answer counterfactual questions about their behavior.
\end{abstract}
\section{Introduction}
Imagine that a student, Alice, is thrilled to have passed her final exam. She wonders, what if she hadn't joined that study group last month? Would she have still succeeded? What if she had picked an easier path than taking this challenging course? Would she have successfully completed her studies at a different university too?
This example illustrates that imagining hypothetical realities to reason about \emph{\enquote{What if}} questions is a cornerstone of human cognition \cite{rafetseder2013counterfactual}. 
In the mathematical formalization of causality, this type of reflection is referred to as \emph{counterfactual reasoning} \cite{gundersen2023counterfactuals}. In Pearl's causal framework, counterfactual reasoning represents the third and highest level of the ladder of causation \cite{pearl2009causal}. 
Counterfactual reasoning is not only used by individuals in thought experiments, but are also common in science which remains controversial as counterfactual claims are typically not falsifiable \cite{dawid1999needs}. For instance, in Rubin's \emph{potential outcome framework} \cite{rubin2005causal}, the effect of a (e.g. medical) treatment is estimated by comparing the treated population to a counterfactual untreated one.
More concretely, \cite{noorbakhsh2022counterfactual} used a counterfactual study to estimate the number of Covid-19 related deaths assuming different transmission rates and \cite{marchezini2022counterfactual}
studied how (hypothetical) social support of a patient would have change their mental health outcomes. \cite{hizli2022joint} and \cite{zucker2021leveraging} investigate outcomes of hypothetical treatments to a patient while \cite{cinquini2024practical} explore counterfactual trajectories in physical systems. 

The framework of \emph{Structural Causal Models} (SCMs) offers a robust framework for the formalization of causality and provide a general recipe for computing counterfactuals based on three steps \cite{pearl2009causal}: abduction, action, and prediction. 
While this is widely accepted as an elegant and principled way of formalizing \emph{\enquote{What if}} questions, this method requires perfect world knowledge (i.e., a known SCM) and absence of measurement noise. 
In practice, SCMs are always a---more or less--coarse abstraction of the real world and only noisy observations are recorded. 
Consequently, even when we understand the laws governing a dynamical system, we rely on numerical methods to approximate unobserved parameters and variables. This challenge is further complicated by the fact that the systems we interact with in the real world are often highly complex.
Moreover, many natural and engineered systems, while deterministic, exhibit chaotic behavior so that small differences in initial conditions can lead to vastly different outcomes, a phenomenon known as sensitive dependence on initial conditions. 

When modeling real-life phenomena in a dynamic setting with hidden states, noise, partial observations, and uncertainty, it becomes exceedingly difficult to determine whether the underlying system exhibits chaotic or simply involves complex nonlinear behavior. Indeed, there is an ongoing debate around the identifiability of chaos in real‐world contexts. For example, the work in \cite{glass1988clocks} shows how biological feedback loops can appear chaotic, yet conclusive evidence for any specific system remains unattainable. Similarly, \cite{dawid1999needs} examines how cardiac processes may exhibit chaotic dynamics, while \cite{elowitz2000synthetic} demonstrates that even small gene regulatory networks can generate complex oscillations. Although not all of these works explicitly focus on chaos, they underscore that subtle feedback loops in biological systems might yield chaotic‐like effects. As a result, when we attempt to model real‐life phenomena, it is difficult to ascertain whether they are truly chaotic or highly complex—an uncertainty that carries serious implications for counterfactual reasoning. If even a slight parameter or measurement error can spur major deviations in a chaotic regime, then counterfactual predictions in such settings must be interpreted with caution.

In light of these debates on whether real‐world systems may exhibit chaotic or complex dynamics akin to chaotic behavior, our work examines what happens if we treat them as though they do—even under strong, seemingly generous assumptions. Specifically, we explore the case where a system truly exhibits sensitive dependence on initial conditions (e.g., a Lorenz-type model) alongside process noise, observational noise, and parameter uncertainty. By combining a state-space model (to capture low-level ODE-based evolution) with a structural causal framework (to reason about hypothetical interventions), we show that—even with full knowledge of the governing equations—small initial perturbations or slight parameter misestimates may still lead to vastly different trajectories. By comparing counterfactual trajectories when parameters are (1) perfectly known, (2) slightly misestimated, and (3) drawn from a posterior distribution, we show that while factual state estimation can remain accurate, counterfactual predictions can become severely unreliable. This highlights how, even with full knowledge of an ODE’s form and a sophisticated inference method, the mere possibility of chaotic behavior coupled with noisy observations and slight uncertainty can pose a fundamental obstacle for reliable counterfactual reasoning in real‐world settings.

\paragraph{Contributions}
\begin{itemize}
    \item To the best of our knowledge, this work is the first to demonstrate how to perform counterfactual reasoning in the context of dynamical systems without assuming (1) perfect observations of the system (i.e., no noise in the process or observations) and without (2) perfect knowledge of the underlying dynamical laws. To this end, we provide a conceptualization and mathematical formalization of the relationship between dynamical systems with inherent stochasticity and Structural Causal Models.
    \item Unlike previous work, we consider the computation of counterfactual predictions in chaotic systems with parameter uncertainty. We show that the reliability of counterfactual predictions cannot be derived (in a trivial way) from observing the original system.
\end{itemize}
The manuscript is organized as follows:
We lay out our notation and formalize dynamical models and (dynamic) SCMs in Section~\ref{sec:background}. 
Our central method is presented in Section~\ref{sec:our_method}, where we show how to compute counterfactuals in realistic scenarios using Bayesian filtering.
Section~\ref{sec:results} uses numerical experiments to illustrate the limitations of counterfactual reasoning in the context of noise, uncertainty, chaos, and abstraction. 
A discussion of related works in Section~\ref{sec:related_works} provides context for our contributions. Finally, a conclusion in Section~\ref{sec:conclusions} completes the manuscript.
\section{Related Work\label{sec:related_works}}
The term \say{counterfactual} has been used in various contexts, particularly in \textit{Counterfactual Explanations} \cite{tsirtsis2021counterfactual} to interpret the behavior of models by describing how changes in inputs would lead to different outputs, answering queries such as \emph{\say{Had this feature been different, the model would have predicted $x$ instead of $y$.}} \textit{Counterfactual Regret Minimization} is used for game strategies by considering alternative actions that could have been taken to minimize regret and improve outcomes. \textit{Counterfactual Fairness} \cite{creager2020causal} is a concept that reflects fairness in AI, stipulating that the model's prediction should be the same in both the observed and counterfactual scenarios where only sensitive attributes are altered.

While counterfactuals are ubiquitous in human reasoning, the reliability of counterfactual reasoning is often questioned. \cite{king2006dangers, king2007can, king2009empirical} and \cite{laugel2019dangers} underscore the risks of model dependence and post-hoc interpretability, where counterfactual explanations may be based on model artifacts rather than real data, potentially misleading users. \cite{king2007detecting} stress that counterfactual inferences often rely on speculative assumptions that lack empirical backing, making verification essential. They recommend methods to detect and control model dependence, particularly for extreme counterfactuals where the assumptions are most fragile. Recent work by \cite{slack2021counterfactual} highlights the instability of counterfactual predictions under perturbations, calling into question the reliability of counterfactual predictions, especially for real-world applications. \cite{keane2021if} highlighted key evaluation gaps in counterfactual explanations for XAI, noting the need for better validation, standardized benchmarks, and practical feasibility assessments to advance the field. A parallel line of work has explored counterfactual reasoning in human psychology. \cite{teigen2011going} explored how counterfactual thinking can make people's judgments more extreme, and how human biases and previous experiences play a key role in people’s counterfactual judgments.

While counterfactual thinking has been used in different contexts, we are particularly interested in counterfactuals in the context of dynamical systems. \cite{baradel2019cophy, janny2022filtered, yi2019clevrer, ding2021dynamic, weilbach2024counterfactual} focus on counterfactual learning of physical dynamics in mechanical systems, which allows predicting the outcomes of physical processes under counterfactual scenarios. The work in \cite{baradel2019cophy} introduces CoPhy, a framework that learns physically interpretable models by training on hypothetical interventions, allowing it to predict how a scene might evolve under altered initial conditions or forces. \cite{janny2022filtered} build on this idea with Filtered-CoPhy, extending counterfactual reasoning to pixel‐level reconstructions and emphasizing unsupervised learning of alternative physical trajectories. \cite{yi2019clevrer} propose CLEVReR, a benchmark specifically designed to test a model’s capacity for causal and counterfactual reasoning in collision‐based scenarios, thereby assessing whether models can correctly infer \say{what would happen if} after object interactions deviate from the observed course. Meanwhile, \cite{ding2021dynamic} develops a dynamic visual reasoning method that integrates visual inputs and language cues to learn differentiable physics models, enhancing the model’s ability to generate counterfactual outcomes for complex mechanical events. \cite{grossmann2024peculiarities} have shown counterintuitive behavior when computing counterfactuals on spatio-temporal point processes.
However, these approaches assume perfect observability of the system states, which is often unrealistic for many real-world scenarios, as shown by \cite{bica2020time}. To this end, \cite{haugh2023counterfactual} introduced an optimization-based framework for counterfactual analysis in a dynamic model with hidden states.
\section{Background}\label{sec:background}
This section introduces three core concepts and their notation: dynamical systems, represented through Ordinary Differential Equations (ODEs) and State-Space Models (SSMs); inference, applying Sequential Monte Carlo (SMC) methods to account for uncertainty and noise; and Structural Causal Models (SCMs), which formalize causality within these systems. We also outline the process of translating ODEs into SSMs, and further, into SCMs, enabling us to apply causal frameworks to well-established dynamical systems from the literature.
\subsection{Dynamical Systems}
\subsubsection{Ordinary Differential Equations}\label{subsec:ode}
Ordinary Differential Equations (ODEs) provide a universal language to describe deterministic systems via equations that determine how variables change in time as a function of other variables \cite{rubenstein2016deterministic}. Consider time-indexed state variables $X_{i}(t) \in \mathbb{R}$, for  $i = 1, \dots, d$. We denote $\mathbf{X}(t) = \left(X_1(t), X_2(t), \dots, X_d(t)\right)$ the vector of all $X_i(t)$. An ODE is defined by:
\begin{align}\label{eq:ODE}
    \frac{d}{dt} \mathbf{X}(t) = h_{\bm{\theta}}(\mathbf{X}(t)), \quad \mathbf{X}(0) = \mathbf{X}_0 \in \mathbb{R}^{d}.
\end{align}
where $\mathbf{h}_{\bm{\theta}} = \left(h_{1, \bm{\theta}},\dots ,h_{d, \bm{\theta}}\right))$ applies $h_{i, \bm{\theta}}$ on each component $X_i(t)$, specifying how each state component evolves over time. The function $\mathbf{h}_{\bm{\theta}}$ depends on parameters $\bm{\theta} = \left(\theta_1, \dots, \theta_p \right)$ that govern the evolution of the system. For each $\bm{\theta} \in\mathbb{R}^{p}$ and initial condition $\mathbf{X}_0\in\mathbb{R}^{d}$ equation \eqref{eq:ODE} describes a unique system.

While ODEs describe the continuous evolution of a system's state over time as $\mathbf{X}(t)$, numerical simulations often require discretizing the system to make it tractable for computational methods. To approximate the continuous behavior, we introduce a time step $\Delta$ and consider a set of discrete time points $t_i = i \cdot \Delta$, approximating the state of the system at these points as $\mathbf{X}_{t_i}$ at the discrete time $t_i$. For ease of notation, we will refer to the system state using the notation $\mathbf{X}_{t}$ to represent the state at discrete time step $t$. 
\subsubsection{State-Space Models}\label{subsec:SSM}
State-space models (SSMs) extend systems described by ODEs by modeling dynamical systems where the true state is hidden and only noisy observations are available at each discrete time point.
These models consist of two main components: the \emph{state equation}, which describes the evolution of the hidden state $\mathbf{X}_t \in \mathbb{R}^{d}$, and the \emph{observation equation}, which links the hidden state to the observed data $\mathbf{Y}_t\in \mathbb{R}^{d}$ \cite{giles2023particleda} (w.l.o.g., we assume the same dimension). The hidden states are assumed to follow a Markov process, and the observations depend only on the state at the corresponding time $t$. Both the hidden states and the observations are subject to additive Gaussian noise. The evolution of $\mathbf{X}_t$ is described by the following equations:
\begin{align} 
    \mathbf{X}_t &= F(\mathbf{X}_{t-1}, \bm{\theta}) + \mathbf{U}_t \label{eq:state}\\ 
   \mathbf{Y}_t &= \mathbf{H} \mathbf{X}_t + \mathbf{W}_t, \label{eq:obs} 
\end{align}
Here, $F(\mathbf{X}_{t-1}, \bm{\theta})$ is the \emph{forward operator}, representing the discretized deterministic part of the system's evolution.
The process noise $\mathbf{U}_t$ and the observation noise $\mathbf{V}_t$ are considered to be normally distributed with mean $\mathbf{0} \in \mathbb{R}^d$ and variance $\mathbf{R}\in \mathbb{R}^{d\times d}$ and $\mathbf{Q}\in \mathbb{R}^{d\times d}$, respectively.
The matrix $\mathbf{H} \in \mathbb{R}^{d \times d}$ is the observation model that linearly maps the hidden state space to the observation space. It defines how each component of the hidden state $\mathbf{X}_t$ contributes to the observed data $\mathbf{Y}_t$. 
\subsection{Sequential Monte Carlo Methods for State and Parameter Estimation} \label{sec:Filtering}
Estimating hidden states and unknown parameters in dynamical systems from noisy observations is a fundamental challenge in many scientific and engineering applications. Sequential Monte Carlo (SMC) methods, commonly known as particle filters, offer a powerful framework for tackling this problem, especially in the context of nonlinear and non-Gaussian state-space models.

In this work, we employ the \emph{nested particle filter} approach introduced by \cite{crisan2018nested} to jointly estimate the hidden states $\mathbf{X}_t$ and the system parameters $\bm{\theta}$. The NPF extends traditional particle filtering by incorporating an additional layer dedicated to parameter estimation. The algorithm consists of two nested layers of particle filters: an \textit{outer} filter that approximates the parameters posterior $\bm{\theta}$ given the observations and a set of \textit{inner} filters, one per sample generated in the outer filter, that yields approximations of the hidden state posterior that result for $\mathbf{X}_t$ conditional on the observations and each specific particle of $\bm{\theta}$. This approach can be broken down to the following steps (see Appendix \ref{sec:NPF}):
\begin{itemize}
    \item Generate $M$ parameter particles $\{\bm{\theta}^{(m)}\}_{1\leq m \leq M}$ from the prior distribution $\pi_0$ and $M\times N$ hidden states particles $\{\mathbf{x}_t^{(n,m)}\}_{1\leq n\leq N}^{1\leq m\leq M}$ from the prior distribution $\tau_0$.
    \item At each time step $t$, the state particles $\{\mathbf{x}_t^{(n,m)}\}$ are propagated using the system dynamics, and their weights are updated based on how they match the observations.
    \item The parameter particles $\{\bm{\theta}^{(m)}\}$ are then updated by aggregating the information from the inner filter (state particles) and computing new weights based on how well the states explain the observations.
\end{itemize}
Nested filtering approaches track the joint posterior  $(\mathbf{X}_t, \boldsymbol{\theta})$ using only observations up till time $t$. To enhance the estimates by incorporating future observations $\mathbf{Y}_{t+1:T}$, we apply a backward smoothing algorithm as detailed in \cite{doucet2009tutorial}. This procedure adjusts the particle weights $w_t^{(n,m)}$ post hoc to account for the entire observation sequence, producing smoothed weights $\tilde{w}_t^{(n,m)}$. Based on these smoothed weights, we obtain refined estimates of the states and parameters (see Appendix \ref{sec:smoothing}).
\subsection{Structural Causal Models}
A Structural Causal Model (SCM) describes a deterministic transformation from a set of exogenous (noise) variables to a set of endogenous (system) variables through a specific data-generating process. One of its core components is a directed acyclic graph (DAG) that represents the causal relationships between variables and can be used to answer causal queries.
We define an SCM as a tuple $(\mathbf{U}, \mathbf{V}, \mathbf{f}, P(\mathbf{U}), \mathbf{PA})$, where: $\mathbf{U} = \{U_1, \dots, U_m\}$ and $\mathbf{V} = \{V_{1}, \dots, V_n\}$ represent the sets of exogenous (noise) variables and endogenous (system) variables, respectively. The set of parent-child relationships, $\mathbf{PA} = \{PA_1, \dots, PA_n\}$, defines each, $V_i \in \mathbf{V}$ with a corresponding set of parent variables $PA_i \subseteq \mathbf{V} \cup \mathbf{U}$ while $U_i \in \mathbf{U}$ are root nodes. We assume an acyclic directed graph structure, allowing for endogenous variables without parents and those with multiple exogenous parents. The set of structural functions $\mathbf{f} = \{f_1(\cdot), \dots, f_n(\cdot)\}$ describes how each $V_i$ is generated by its causal parents, with $V_i := f_i(\text{PA}_i)$. The exogenous variables follow a probability distribution $P(\mathbf{U})$, typically assuming independence between noise variables.
\subsubsection{Computing Counterfactuals}\label{sec:cft}
SCMs make it possible to study the effects of modifications to the data generating process (e.g., by fixing the value of a specific $V_i$ or removing a dependency) and \enquote{imagining} how the output might look like. 
Specifically, we need to take an SCM and an observation as input, and hypothesize about how the observational data would have looked like under a different (modified) SCM.
Computing counterfactuals involves a three-step procedure \cite{pearl2009causal}: \textbf{Abduction}, where exogenous variables are inferred from observed data, computing the posterior distribution of noise variables; \textbf{Action}, which modifies the SCM by fixing certain variables or relationships; and \textbf{Prediction}, solving the modified SCM using the inferred noise posterior rather than the original distribution.

The abduction step is computationally the most intensive, as it requires solving an inverse problem. While we defined an SCM as a transformation from noise variables to endogenous variables, the abduction step reverses this process by identifying the noise variables that lead to a specific assignment of endogenous variables. In Bayesian terms, the observations update the prior distribution $P(\mathbf{U})$ to a posterior distribution conditioned on the evidence. 
Typically, this posterior distribution is more complex than the prior, as it lacks a closed analytical form and loses the independence between noise variables.

The intuitive reasoning behind this approach is as follows: We assume that all inherent stochastic and environmental influences—conceptually considered as noise—are encapsulated within the noise variables of the SCM. By fixing the noise (as accurately as possible), we essentially stabilize all environmental effects and other disturbances, allowing the system to be re-evaluated under specific modifications. The underlying assumption of this process is that sufficient knowledge about the noise can be captured through observations and that the noise variables in the modified SCM correspond meaningfully to those in the original model.
\subsubsection{Dynamical SCMs}
SCMs are typically acyclic and primarily used to model static systems. However, generalizing SCMs to dynamical systems presents multiple possibilities. One of the simplest approaches is to unfold the SCM across discrete time steps. This method involves creating a copy of the SCM for each time step $t$, ensuring that the parents of any variable $V_i$ are confined to the present or past time steps, but not the future. This approach was essentially employed by \cite{oberst2019counterfactual} to handle temporal data within the SCM framework.
Alternatively, \cite{bongers2018causal} introduced dynamical SCMs, which are similar but incorporate additional parametric assumptions about how the unfolded SCM evolves over time. However, to avoid unnecessary assumptions and design choices, we adopt the simpler unfolding method. This approach is principled and aligns well with the SSM framework, making it suitable for modeling dynamical systems over discrete time steps.
\section{Methodology}\label{sec:our_method}
In this section, we outline how we model the dynamical system as an SSM to infer the system's parameters and hidden states. Importantly, we establish how SSMs can be interpreted as an SCM unfolded over time (Section~\ref{sec:SSM-SCM}), shifting the focus from merely describing system dynamics to the underlying causal relationships between variables and external influences. While the SSM provides a low-level, quantitative description of the system's evolution, the SCM framework offers a higher-level perspective that captures how changes in variables or external conditions propagate through the system. 
This perspective allows us to compute counterfactuals by first performing an abduction step—where we infer the latent variables and noise terms from observations (Section~\ref{sec:abduct}). The subsequent prediction step depends on the type of counterfactual intervention being considered, such as perturbing initial conditions (Section~\ref{sec:prediction}). 
\subsection{Translating SSMs to SCMs}\label{sec:SSM-SCM}
Converting an SSM into an SCM is simple by treating all $\mathbf{X}_t$ and $\mathbf{Y}_t$ as endogenous variables, while all $\mathbf{U}$ and $\mathbf{W}$ are considered noise variables. 
The hidden state $\mathbf{X}_t$ depends on the previous state $\mathbf{X}_{t-1}$ (with $t=0$ as user-specified) and the process noise. Likewise, the observation $\mathbf{Y}_t$ is determined by $\mathbf{X}_t$ and the observational noise. Thus, the state equation \eqref{eq:state} $\mathbf{X}_{t} = F(\mathbf{X}_{t-1}, \bm{\theta}) + \mathbf{U}_t$ and observation equation \eqref{eq:obs} $\mathbf{Y}_t = \mathbf{H} \mathbf{X}_t + \mathbf{W}_t$ are considered as the structural equation from the SCM perspective. Handling the parameter $\bm{\theta}$ is less straightforward. In principle, it can be hard-coded into the structural functions of the SCM. However, since the system parameters are typically unknown and need to be inferred, a more principled approach is to represent them as nodes in the SCM.

\begin{figure}
\centering
\begin{tikzpicture}
  \node[draw, circle, fill=red!30] (X0) {$\mathbf{X_0}$};
 \node[draw, circle, fill=red!30, right=of X0] (X1) {$\mathbf{X_1}$};
  \node[right= 0.5cm of X1] (dots1) {$\cdots$};
  \node[draw, circle, fill=red!30, right= 0.5cm of dots1] (Xt) {$\mathbf{X_t}$};
   \node[right= 0.5 cm of Xt] (dots2) {$\cdots$};
  \node[draw, circle, fill=red!30, right= 0.5cm of dots2] (XT) {$\mathbf{X_T}$};

  \node[draw, circle, fill=blue!30, below=1cm of X0] (Y0) {$\mathbf{Y_0}$};
  \node[draw, circle, fill=blue!30, below=1cm of X1] (Y1) {$\mathbf{Y_1}$};

    \node[draw, circle, fill=blue!30, below=1 cm of Xt] (Yt) {$\mathbf{Y_t}$};
  \node[draw, circle, fill=blue!30, below=1 cm of XT] (YT) {$\mathbf{Y_T}$};

\node[draw, circle, fill=gray!30, above =0.5cm of X1, , scale=0.7] (U1) {$\mathbf{U_1}$};
\node[draw, circle, fill=gray!30, above =0.5cm of Xt, , scale=0.7] (Ut) {$\mathbf{U_t}$};
\node[draw, circle, fill=gray!30, above =0.5cm of XT, , scale=0.7] (UT) {$\mathbf{U_T}$};

\node[draw, circle, fill=gray!30, above left=0.5cm and 0.5cm of Y0, , scale=0.7] (W0) {$\mathbf{W_0}$};
\node[draw, circle, fill=gray!30, above left=0.5cm and 0.5cm of Y1, , scale=0.7] (W1) {$\mathbf{W_1}$};
\node[draw, circle, fill=gray!30, above left=0.5cm and 0.5cm of Yt, , scale=0.7] (Wt) {$\mathbf{W_t}$};
\node[draw, circle, fill=gray!30, above left=0.5cm and 0.5cm of YT, , scale=0.7] (WT) {$\mathbf{W_T}$};

  \node[draw, circle, fill=olive!20, above = 1.5cm of Xt, scale=1] (theta1) {$\bm{\theta}$};
  

  \draw[->, >=stealth, line width=1pt] (X0) -- (X1);
  \draw[->, >=stealth, line width=1pt] (X1) -- (dots1);
  \draw[->, >=stealth, line width=1pt] (dots1) -- (Xt);
  \draw[->, >=stealth, line width=1pt] (Xt) -- (dots2);
  \draw[->, >=stealth, line width=1pt] (dots2) -- (XT);
  
  \draw[->, >=stealth, line width=0.4pt] (U1) -- (X1);
  \draw[->, >=stealth, line width=0.4pt] (Ut) -- (Xt);
  \draw[->, >=stealth, line width=0.4pt] (UT) -- (XT);

  \draw[->, >=stealth, line width=0.8pt] (theta1) -- (X1);
  \draw[->, >=stealth, line width=0.8pt] (theta1) -- (XT);
\draw[->, >=stealth, bend left=40, line width=0.8pt] (theta1) to (Xt);
  \draw[->, >=stealth, line width=0.8pt] (X0) -- (Y0);
  \draw[->, >=stealth, line width=0.8pt] (X1) -- (Y1);
  \draw[->, >=stealth, line width=0.8pt] (Xt) -- (Yt);
  \draw[->, >=stealth, line width=0.8pt] (XT) -- (YT);

  \draw[->, >=stealth, line width=0.4pt] (W0) -- (Y0);
  \draw[->, >=stealth, line width=0.4pt] (W1) -- (Y1);
\draw[->, >=stealth, line width=0.4pt] (Wt) -- (Yt);
  \draw[->, >=stealth, line width=0.4pt] (WT) -- (YT);

\end{tikzpicture}
\caption{Graphical representation that highlights the equivalence between SSMs and SCMs.} 
\label{fig:graph_SCM}
\end{figure}
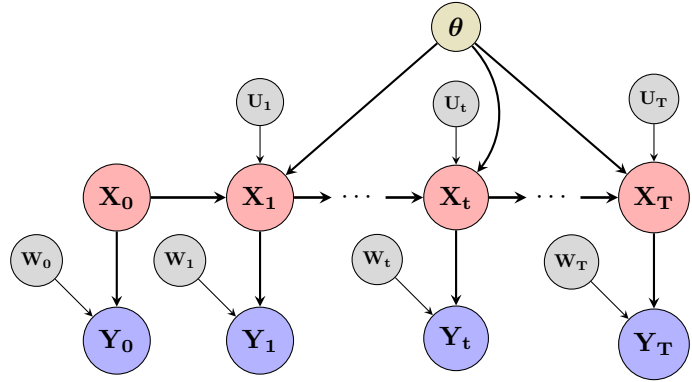

Figure~\ref{fig:graph_SCM} provides a causal graph illustrating the evolution of the hidden states under the influence of process noise. Each node represents a causal relationship at time $t$, demonstrating how hidden states evolve sequentially and influence the observed outputs. This causal interpretation of SSMs allows us to leverage the SCM framework for reasoning about interventions and counterfactual scenarios, where each time step becomes a causal node governed by the previous state and the noise perturbations. Having established this interpretation, we now proceed to describe how counterfactual trajectories are generated. 
\subsection{Estimating the Noise Posterior With NPF Smoothing}\label{sec:abduct}
It is virtually infeasible to provide a closed-form expression for the joint posterior distribution of the hidden state and parameters, but we can jointly estimate the hidden states and system parameters using a nested particle filter-smoothing approaches. To this end, we follow the conventional NPF approach mentioned in \ref{sec:Filtering} and detailed in \cite{crisan2018nested}.

\subsection*{Initialization}
\begin{itemize}[leftmargin=*, label={--}]
    \item \textbf{Parameter Particles:} Draw \(M\) samples \(\{\bm{\theta}^{(m)}\}_{m=1}^{M}\) from the parameter prior \(\pi_0\). This captures our initial uncertainty about \(\theta\).
    \item \textbf{State Particles:} For each parameter \(\bm{\theta}^{(m)}\), generate \(N\) state particles \(\{\mathbf{x}_0^{(n,m)}\}_{n=1}^{N}\) from the state prior. In our work, the state evolution is modeled by the forward operator \(F\) defined above.
\end{itemize}

\subsection*{Recursive Update (for each time step \(t\))}
\begin{enumerate}
    \item \textbf{Parameter Jittering (Mutation):}  
    For each parameter particle \(\bm{\theta}^{(m)}_{t-1}\), apply a jittering kernel \(\kappa_N\) to obtain a new candidate
    \[
    \bar{\bm{\theta}}^{(m)}_t \sim \kappa_N(\cdot \mid \bm{\theta}^{(m)}_{t-1}).
    \]
    \item \textbf{State Propagation:}  
    For each jittered parameter \(\bar{\bm{\theta}}^{(m)}_t\) and its associated state particles \(\{\mathbf{x}_{t-1}^{(n,m)}\}_{n=1}^{N}\), update the state by
    \[
    \mathbf{x}_{t}^{(n,m)} = F_t(\mathbf{x}_{t-1}^{(n,m)}, \bm{\theta}^{(m)}) + \mathbf{U}_t, \quad \mathbf{U}_t \sim \mathcal{N}(\mathbf{0}, \mathbf{R}), \; 1\leq n \leq N.
    \]
    \item \textbf{Inner Weighting (State Update):}  
    For each state particle \(\mathbf{x}_t^{(n,m)}\), compute the weight $w_t^{(n,m)} \propto p\bigl(\mathbf{Y}_t \mid \mathbf{x}_t^{(n,m)}\bigr),$
    \item \textbf{Outer Weighting (Parameter Update):}  
    For each candidate \(\bar{\bm{\theta}}^{(m)}_t\), approximate its marginal likelihood by averaging the inner weights:
    \[
    u_t\bigl(\bar{\bm{\theta}}^{(m)}_t\bigr) \approx \frac{1}{N}\sum_{n=1}^{N} p\bigl(\mathbf{Y}_t \mid \mathbf{x}_t^{(n,m)}\bigr), \qquad     v_t^{(m)} \propto u_t\bigl(\bar{\bm{\theta}}^{(m)}_t\bigr).
    \]
    \item \textbf{Parameter Resampling:}  
    Resample the parameter particles (and their associated state particles) according to \(v_t^{(m)}\) to yield the updated set $\{\bm{\theta}_t^{(m)},\, \{\mathbf{x}_t^{(n,m)}\}_{n=1}^{N}\}_{m=1}^{M}$.

    This collection approximates the joint posterior distribution of \(\mathbf{X}_t\) and \(\theta\).
\end{enumerate}
 
However, NPF algorithm tracks the states and parameters $(\mathbf{X}_t,\bm{\theta})$ using observation only up till time $t$. As detailed in Section \ref{sec:cft},  the noise abduction step requires the full set of observations (i.e., up to time $T$). Therefore, once the forward filtering is complete, we apply a backward smoothing pass to refine the state and parameter estimates by incorporating information from future observations $\mathbf{Y}_{t:T}$. 
\paragraph{Inferring the Exogenous Noise}
Once the hidden states are estimated, abducting the noise variables are computed using the structural equation in \eqref{eq:state}. Specifically, with every state and parameter particle $(\mathbf{x}_{t}^{(n, m)}, \bm{\theta}^{(m)})$ the corresponding noise $\bm{\mu}_{t}^{(n,m)}$ is calculated as the residual of the following form:
\begin{align}
    \bm{\mu}_{t}^{(n,m)} = \mathbf{x}_{t}^{(n, m)} - F\left(\mathbf{x}_{t-1}^{(n, m)}, \bm{\theta}^{(m)}\right).
\end{align}
Then the abducted noise ${\mathbf{U}_t^{\text{cf}}}$ can be approximated with a normal distribution with mean $\bm{\mu}_t$ and variance $\bm{\sigma}_t$. The mean $\bm{\mu}_t$ is calculated as the weighted average of the residuals $\bm{\mu}_{t}^{(n, m)}$ at each time $t$. Concretely, ${\mathbf{U}_t^{\text{cf}}}$ is sampled from $\mathcal{N}(\bm{\mu}_t, \bm{\sigma}_t)$, with
\begin{align*}
     \bm{\mu}_t = \sum_{n, m = 1}^{N, M} \tilde{w}_t^{(m,n)} \left( \mathbf{x}_{t}^{(n, m)} - F(\mathbf{x}_{t-1}^{(n, m)}, \bm{\theta}^{(m)})\right) \quad \text{and} \quad
     \bm{\sigma}_t = \sum_{n, m = 1}^{N, M} \tilde{w}_t^{(m,n)}\left( \bm{\mu}_t^{(n,m)} - \bm{\mu}_t\right)^2,
\end{align*}
where  $\tilde{w}_t^{(m,n)}$ are the normalized smoothed weights returned by the backward smoothing layer. Here, $\bm{\sigma}_t$ is the weighted weighted variance.
\subsection{Generating Counterfactual Trajectories}\label{sec:prediction}
To generate the counterfactual sequence, we first define the counterfactual Structural Causal Model (CF-SCM), which -given the abducted noise $\mathbf{U}_t^{\text{cf}}$- is specified according to the type of intervention. In this work, we mainly focus on interventions on the initial conditions: how the sequence of hidden states would evolve if the initial state had been different. Specifically, if the initial state was slightly perturbed with an insignificant value $\delta$, ($\mathbf{X}_0^{cf} = \mathbf{X}_0 + \delta\mathbf{e}_j$), how would the rest of the sequence propagate $\mathbf{X}_{1:t}^{cf}$? This is particularly relevant in the context of chaotic systems, as a slight change in initial conditions would lead to vastly different sequences. 
\begin{align}\label{eq:CSCM1}
\begin{cases}
        \mathbf{X}^{\text{cf}}_{t} &:=
        F_{t}\left(\mathbf{X}^{\text{cf}}_{t-1}, \Tilde{\bm{\theta}} \right) + \mathbf{U}_t^{\text{cf}}, \quad  \mathbf{U}_t^{\text{cf}} \sim \mathcal{N}\left(\ \bm{\mu}_t,  \bm{\sigma}_t\right).\\ 
        \mathbf{X}_0^{cf} &= \mathbf{X}_0 + \delta\mathbf{e}_j,
\end{cases}
\end{align}
with $\delta>0$ is a small perturbation and $\mathbf{e}_j$ is $j$-th unit vector.
Note that if the governing parameters in the CF-SCM reflect the true underlying dynamics, $\Tilde{\bm{\theta}} = \bm{\theta}_{true}$, the counterfactual trajectory should closely resemble the deterministic counterfactual projection, especially when the system is endowed with a structure a priori. However, if the parameter estimates are slightly inaccurate, $\Tilde{\bm{\theta}} = \hat{\bm{\theta}}$, the counterfactual predictions can become unreliable. This unreliability arises because chaotic systems exhibit unpredictable behavior for certain parameters while being completely predictable for others. To this end, we consider three options for $\Tilde{\bm{\theta}}$ in equation \eqref{eq:CSCM1}:
\begin{itemize}
    \item  $\Tilde{\bm{\theta}}$ is set to the \textit{true} parameter values ($\Tilde{\bm{\theta}} = \bm{\theta}_{true}$). In this configuration of the CF-SCM, there is no uncertainty in the underlying dynamics of the counterfactual scenario. The reliability of CF trajectories is influenced solely by the estimated posterior distribution of the noise.
    \item $\Tilde{\bm{\theta}}$ is set to the \textit{estimated} parameter ($\Tilde{\bm{\theta}} = \hat{\bm{\theta}}$), here $\hat{\bm{\theta}}$ is obtained using the filtering-smoothing algorithm described in Section \ref{sec:Filtering}. This setting allows us to assess counterfactual trajectories when a slight inaccuracy is introduced into the underlying dynamics of the system.
    \item $\Tilde{\bm{\theta}}$ is sampled from the full posterior distribution of the parameters $\mathcal{N}(\hat{\bm{\theta}}, \bm{\sigma}_{\bm{\theta}})$. This posterior distribution captures the uncertainty around the parameter estimates, reflecting the range of plausible values given the full length of observations and model assumptions. This approach assesses the reliability of the counterfactual reasoning while accounting for parameter uncertainty, simulating a more realistic scenario where the true parameter values are not known precisely.
\end{itemize}
\section{Case Studies}\label{sec:results}
\begin{figure}[h]
\centering
\begin{tikzpicture}
\draw[->, line width=0.4mm] (0, 1) -- (0, -6) coordinate[midway] (mid) node[midway, left] {};
\node[draw, circle, fill=blue!20, above right  = 3cm and 4.5cm of mid] (Y0) {$\mathbf{Y}_0$};
\node[right=of Y0] (dotsy1) {$\cdots$};
\node[draw, circle, fill=blue!40, right=of dotsy1] (Yt) {$\mathbf{Y}_t$};
\node[right=of Yt] (dotsyt) {$\cdots$};
\node[draw, circle, fill=blue!20, right=of dotsyt] (YT) {$\mathbf{Y}_T$};

\node[above right = 1 cm and -0.7cm of Y0] (a) {(a) Hidden state and parameters estimation};

\node[draw, circle, fill=red!40, below = 5.2cm of Y0, scale = 0.8] (Xcf0) {$\mathbf{X}^{cf}_0$};

\node[right=of Xcf0] (dotsx1) {$\cdots$};
\node[draw, circle, fill=red!50, below=5.2cm of Yt, scale = 0.8] (Xcft) {$\mathbf{X}^{cf}_t$};
\node[right=of Xcft] (dotsx2) {$\cdots$};
\node[draw, circle, fill=red!40, below=5.2cm of YT, scale = 0.8] (XcfT) {$\mathbf{X}^{cf}_T$};
\node[above = 0.3cm of Xcf0] (act) {$\mathbf{X}_{0}^{cf} = \mathbf{X}_{0} + \delta \mathbf{e}_j$};

\node[below  =0.6cm of Yt] (estxt) {
$\{\hat{\mathbf{x}}_t, \hat{\bm{\theta}}\} = NPFS(\mathbf{Y}_{0:T})$
};

\node[draw, circle, fill = gray!30, above left = 2 cm and 1.6cm of Xcft, , scale=0.7] (abdUt) {${\mathbf{U}}_t^{\text{cf}}$};
\node[draw, circle, fill = olive!30, above right = 0.8 cm and 1cm of Xcft, , scale=0.8] (abdTheta) {${\bm{\theta}}$};

\node[right = 0.4 cm of abdTheta] (thet) {
${\bm{\theta}} \sim \mathcal{N}\left(\hat{\bm{\theta}}, \bm{\sigma}_{\bm{\theta}} \right)$
};
\node[right = 0.1cm of abdUt] (cond) {
${\mathbf{U}}_t^{\text{cf}} = \mathbf{U}_t | (\mathbf{X}_t = \hat{\mathbf{x}}_t,\bm{\theta} = \hat{\bm{\theta}})$
};

\draw[decorate,decoration={brace,amplitude=15pt,mirror}, line width=1pt] ([xshift=0cm, yshift=-0.2cm]Y0.south west) -- ([xshift=0cm, yshift=-0.2cm]YT.south east) ;
\node[above right = 3.2 cm and 1cm of Xcf0] (b) {(b) Counterfactual SCM};

\node (Obs) [anchor=west, text = blue!50] at ([xshift=0.3cm, yshift=3.3cm]mid.west) {Observations};
\node (NPF) [anchor=west, text = black] at ([xshift=0.3cm, yshift=2cm]mid.west) {Smoothing};
\node (Abd) [anchor=west, text = gray] at ([xshift=0.3cm, yshift=-0.2cm]mid.south) {Abduction};
\node (int) [anchor=west, text = gray] at ([xshift=0.3cm, yshift=-1.6cm]mid.south) {Action};
\node (pred) [anchor=west, text = red!60] at ([xshift=0.3cm, yshift=-2.8cm]mid.south) {Prediction};

\draw[dashed, gray!60, line width=1pt] ([xshift=-2.6cm, yshift=0.5cm]abdUt.west) -- ([xshift=7.1cm, yshift=0.5cm]abdUt.east);
\draw[dashed, gray!60, line width=1pt] ([xshift=-2.6cm, yshift=-0.5cm]abdUt.west) -- ([xshift=7.1cm, yshift=-0.5cm]abdUt.east);
  
\draw[->, >=stealth, line width=0.4pt] (Xcf0) -- (dotsx1);
\draw[->, >=stealth, line width=0.7pt] (dotsx1) -- (Xcft);
\draw[->, >=stealth, line width=0.8pt] (Xcft) -- (dotsx2);
\draw[->, >=stealth, line width=0.4pt] (dotsx2) -- (XcfT);

\draw[->, >=stealth, line width=0.8pt] (abdUt) -- (Xcft);

\draw[->, >=stealth, line width=0.8pt] (abdTheta) -- (Xcft);
\draw[->, >=stealth, line width=0.8pt] (abdTheta) -- (XcfT);
\draw[->, dashed, dash pattern=on 3pt off 2pt, line width=0.4pt] (abdTheta) -- (dotsx1);
\draw[->, dashed, dash pattern=on 3pt off 2pt, line width=0.4pt] (abdTheta) -- (dotsx2);

\draw[draw=gray!60, line width=1pt, rounded corners=10pt] ([xshift= -2 cm, yshift=-1.8cm]Y0.east) rectangle ([xshift=1.1cm, yshift=0.8cm]YT.east);
\draw[draw=gray, line width=1pt, rounded corners=10pt] ([xshift=-2cm, yshift=-0.6cm]Xcf0.east) rectangle ([xshift=1.1cm, yshift=3.6cm]XcfT.east);
\end{tikzpicture}
\caption{Illustration of the proposed methodology to generate counterfactuals. Panel (a) shows the estimation of states and parameters, using the filtering/smoothing (NPFS) method, given a sequence of observations $\mathbf{Y}_{0:T}$. Panel (b) CF-SCM: After abducting the noise $\mathbf{U}_t^{\text{cf}}$, an intervention is applied to the initial state $\mathbf{X}_{0}$, resulting in a counterfactual trajectory $\mathbf{X}_{0:T}^{\text{cf}}$.} 
\label{fig:graph_flow}
\end{figure}
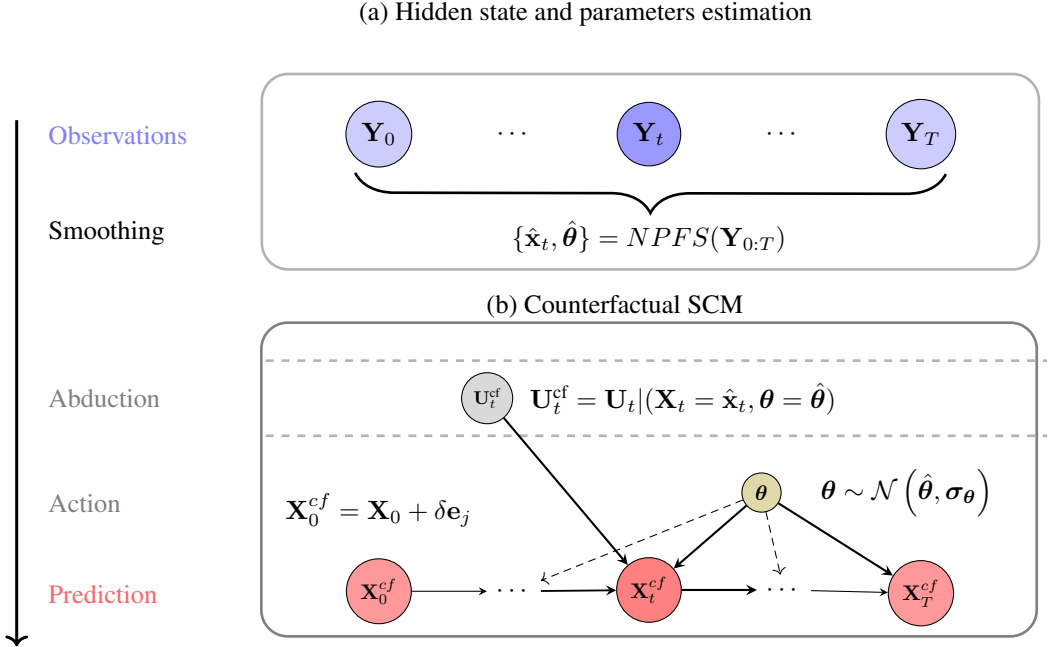

This section presents experimental results that illustrate the sensitivity of counterfactual reasoning in dynamical systems. The experiments are intentionally simple, designed to approximate real-world scenarios under controlled conditions, and are performed on systems that are nearly at a toy level. Specifically, we investigate how common assumptions in the mathematical modeling of real-world systems---such as noisy observations, model uncertainty, and chaotic dynamics---affect the quality and reliability of counterfactual reasoning within these systems. The flow of our approach is summarized in Figure~\ref{fig:graph_flow}. We generate a noisy trajectory by simulating a dynamical 'ground truth' model from the literature with known parameters. Next, we apply the filtering/smoothing technique from Section \ref{sec:our_method} to estimate both hidden states and system parameters. Using these estimates, we approximate the posterior noise distribution. The CF-SCM is then constructed by incorporating the abducted noise and performing an intervention at the initial conditions. We generate multiple counterfactual trajectories by sampling from the estimated posterior noise, with the parameter node $\Tilde{\bm{\theta}}$ in the CF-SCM following the three options described in Section \ref{sec:prediction}.
Due to space constraints, we provide a brief overview of our experimental setup here, with full details available in Appendix \ref{appendix:expsetup}. We evaluate counterfactual reliability on several dynamical systems---including the Lorenz, R\"{o}ssler, and Logistic Growth models---using an NPF for state and parameter estimation. The specifics of the models, initial conditions, parameter priors, and evaluation metrics are thoroughly described in the appendix. 

To evaluate the reliability of counterfactual sequences, we compare the estimated counterfactual sequences, $\mathbf{X}^{\text{cf}_i}_{1:T}$, and the deterministic counterfactual sequence, $\mathbf{X}_{1:T}^{\text{cf}}$. This comparison is conducted through one- and two-dimensional plots, along with a line plot of the Root Mean Squared Error across time (RMSE$_t$; see Appendix \ref{appendix:RMSEt}). Note that, $\mathbf{X}^{\text{cf}}_{1:T}$ represents the \textit{deterministic} counterfactual trajectory. This means that starting from a counterfactual initial condition, $\mathbf{X}_0^{\text{cf}}$, the subsequent sequence is computed using purely deterministic dynamics. In this setting, where the experiments aim to mimic real-world scenarios in a controlled manner, having access to $\mathbf{X}_{1:T}^{\text{cf}}$ is useful for comparison with the generated counterfactual sequences, providing a reference point for evaluating the accuracy of the generated counterfactuals.
\subsection{Experimental Results}
In this section, we present the experimental results based on the counterfactual sequence generation process outlined in Section~\ref{sec:prediction}. Specifically, the parameter $\Tilde{\bm{\theta}}$ in the CF-SCM, can be set to the true parameters $\bm{\theta}_{true}$, the point estimate $\hat{\bm{\theta}}$, and sampled from the approximated posterior parameters $\mathcal{N}(\hat{\bm{\theta}}, \bm{\sigma}_{\bm{\theta}})$. Observational noise and process noise are sampled from a normal distribution with mean $\mathbf{0}$ and variances $\sigma_{\mathbf{W}}\mathbf{I}$ and $\sigma_{\mathbf{U}}\mathbf{I}$, respectively, with $\sigma_{\mathbf{W}}$ and $\sigma_{\mathbf{U}}$ taking values in $\left\{(0.01,4), (0.01,9), (1,2), (4,1)\right\}$. The initial conditions for the Lorenz, R\"{o}ssler and logistic growth systems are set to $(1, 1, 1)$ and $(1, 1, 0)$, and $10$ respectively. The counterfactual initial conditions are defined for Lorenz and R\"{o}ssler systems as $\mathbf{X}^{\text{cf}}_0 = \mathbf{X}_0 + 10^{-4}\mathbf{e}_1$, where $\mathbf{e}_1$ represents a unit perturbation along the first coordinate. For logistic growth, the initial condition is set to $X^{\text{cf}} = X_0 + 10$.
\begin{figure}
    \centering
        \begin{minipage}{\textwidth}
        \centering
        \begin{minipage}{0.333\textwidth}
            \centering
            $\Tilde{\bm{\theta}} = {\bm{\theta}}_{\text{true}}$
        \end{minipage}%
        \hfill
        \begin{minipage}{0.333\textwidth}
            \centering
            $\Tilde{\bm{\theta}} = \hat{\bm{\theta}}$
        \end{minipage}%
        \hfill
        \begin{minipage}{0.333\textwidth}
            \centering
            $\Tilde{\bm{\theta}} \sim \mathcal{N}\left(\hat{\bm{\theta}}, \bm{\sigma}_{\bm{\theta}}\right)$
        \end{minipage}
    \end{minipage}
    \begin{minipage}[b]{0.32\textwidth}
        \centering
        \includegraphics[width=\textwidth]{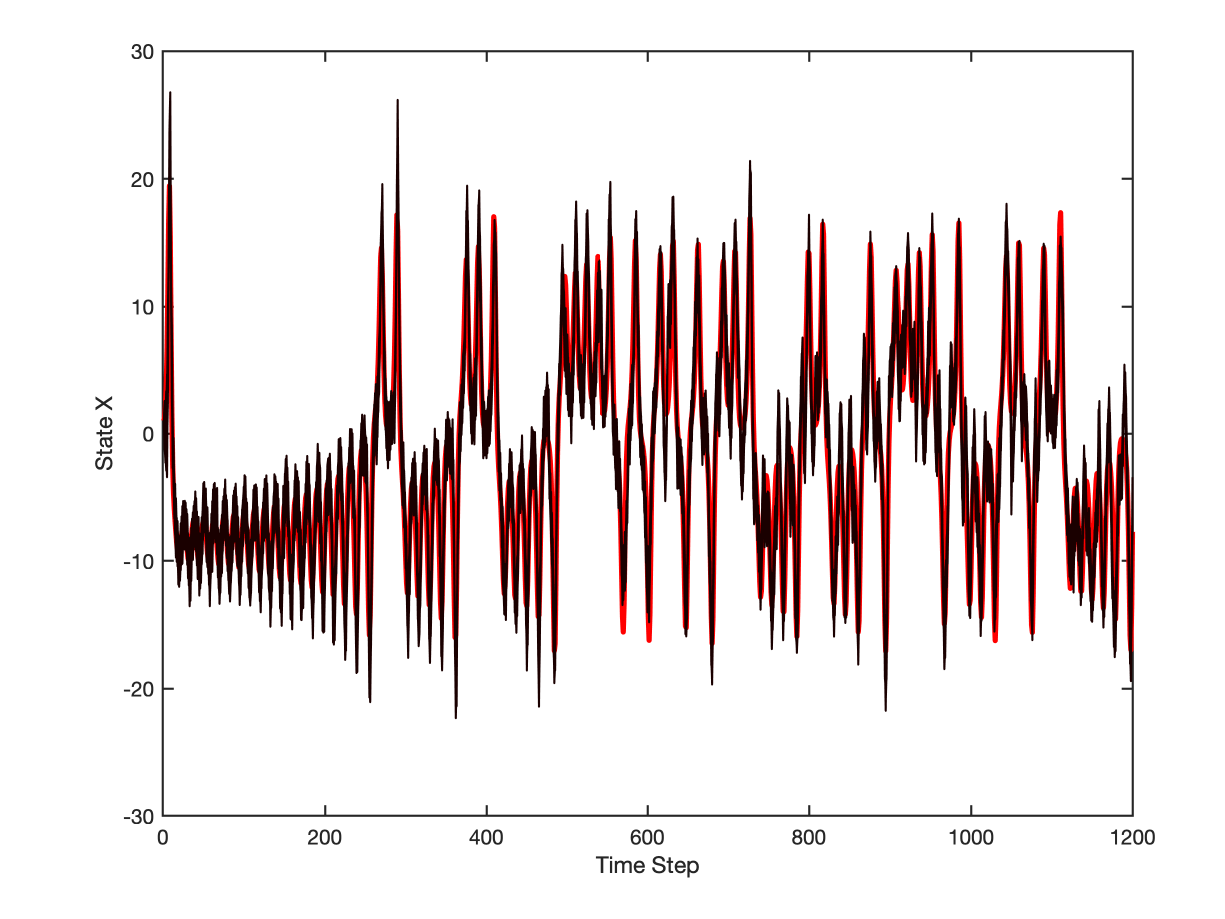} %
    \end{minipage}
    \begin{minipage}[b]{0.32\textwidth}
        \centering
        \includegraphics[width=\textwidth]{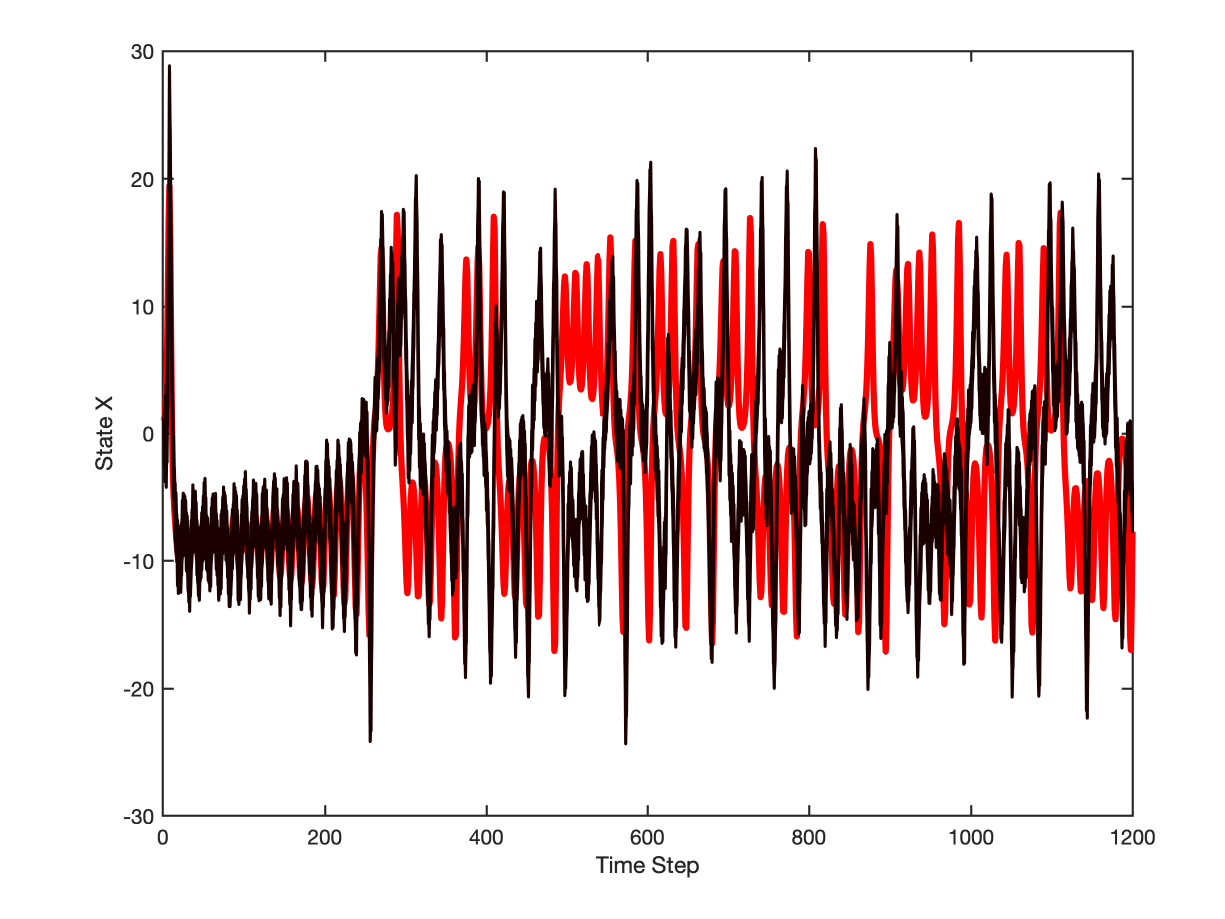} %
    \end{minipage}
        \begin{minipage}[b]{0.32\textwidth}
        \centering
        \includegraphics[width=\textwidth]{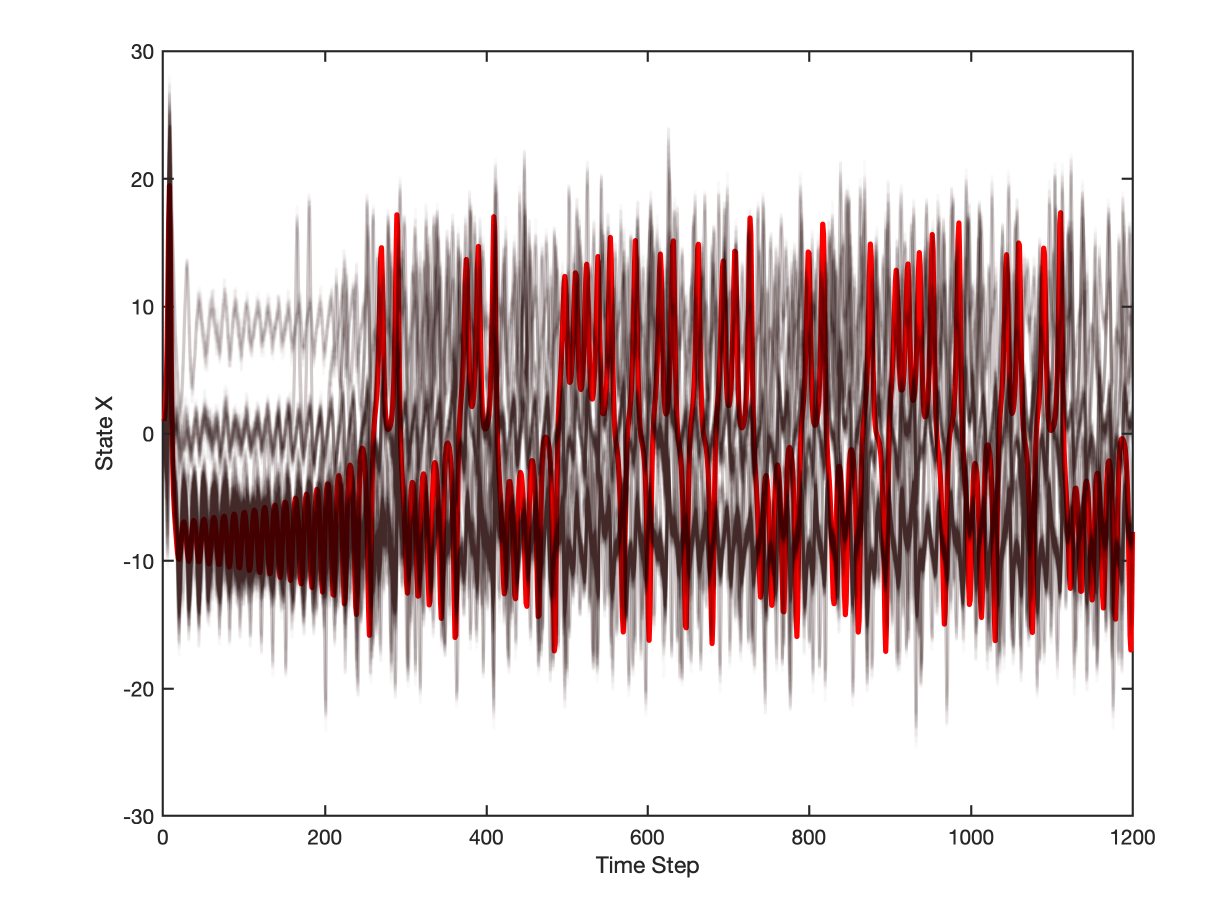} 
    \end{minipage}
    \begin{minipage}[b]{0.32\textwidth}
        \centering
        \includegraphics[width=\textwidth]{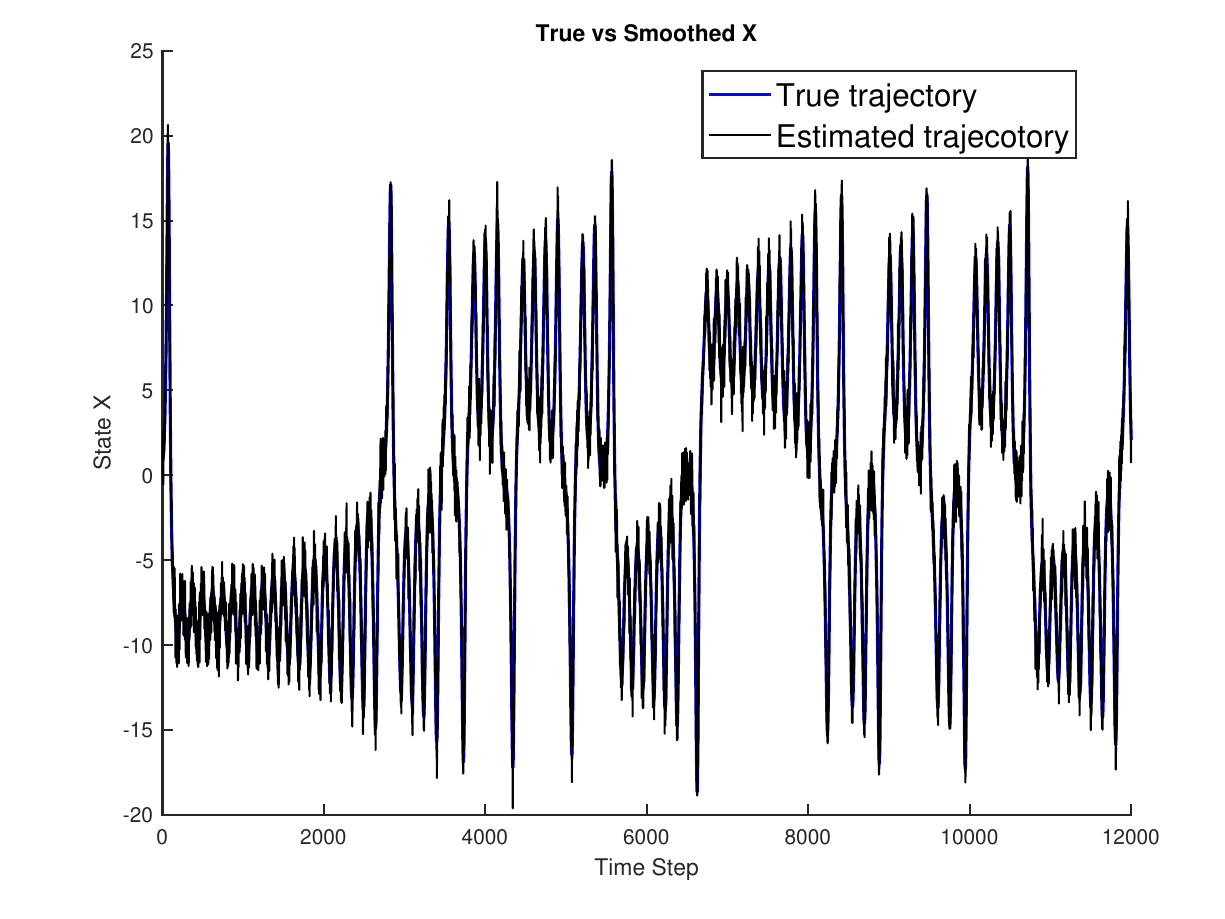} %
    \end{minipage}
    \begin{minipage}[b]{0.32\textwidth}
        \centering
        \includegraphics[width=\textwidth]{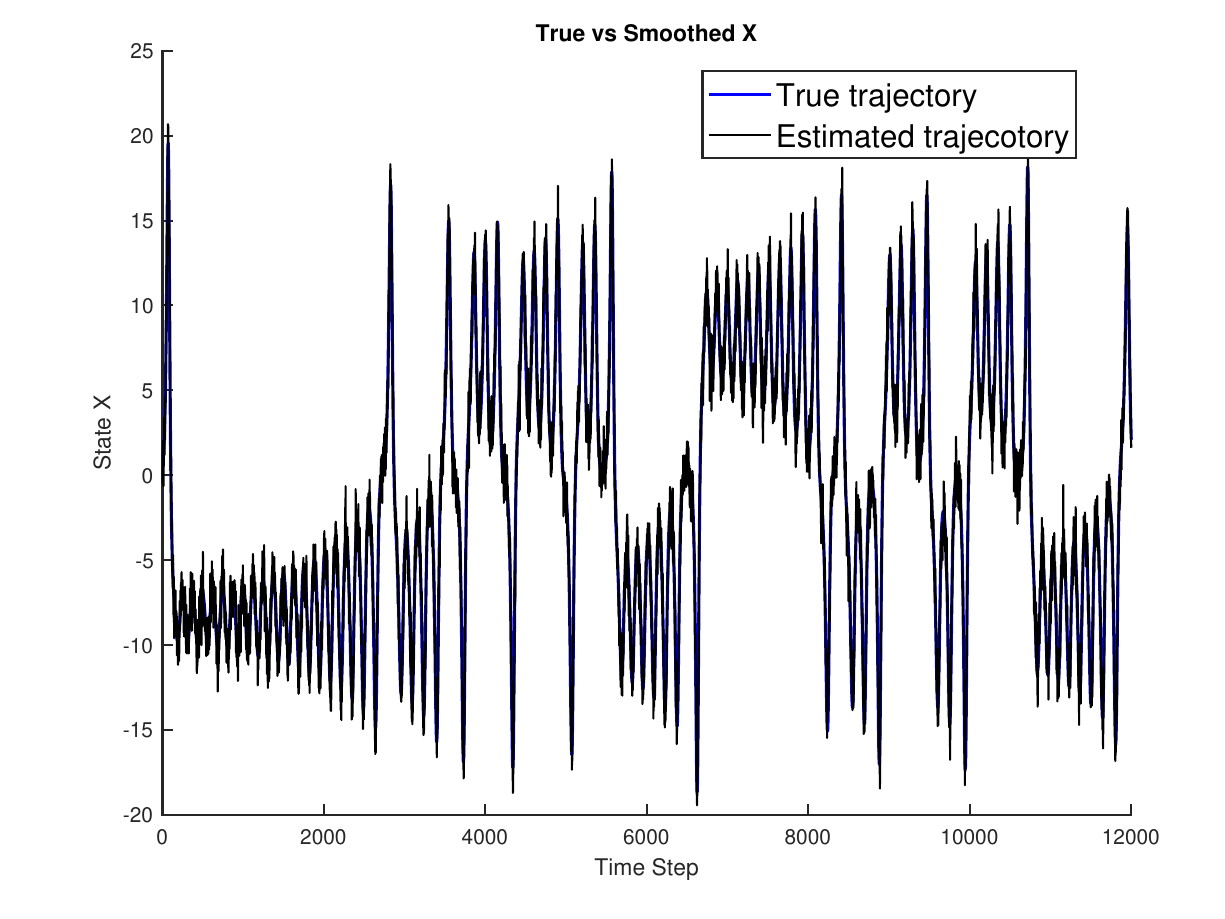} %
    \end{minipage}
        \begin{minipage}[b]{0.32\textwidth}
        \centering
        \includegraphics[width=\textwidth]{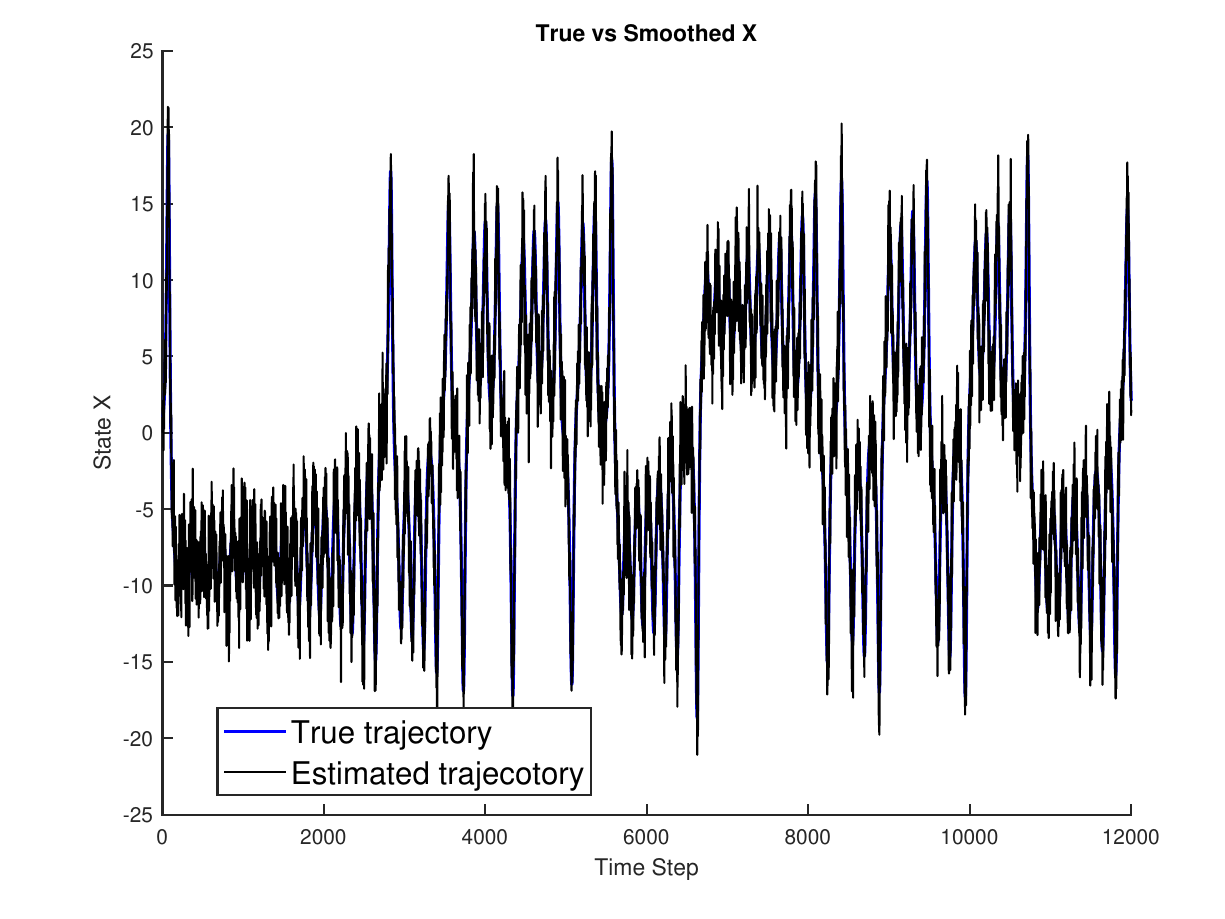} 
    \end{minipage}
    \caption{Top row: Deterministic counterfactual trajectory (in red) compared to the generated counterfactual trajectories (in black) for the Lorenz system under three parameter assumptions: $\Tilde{\bm{\theta}} = {\bm{\theta}}_{true}$, $\Tilde{\bm{\theta}} = \hat{\bm{\theta}}$, and $\Tilde{\bm{\theta}} \sim \mathcal{N}(\hat{\bm{\theta}}, \bm{\sigma}_{\bm{\theta}})$. Bottom row: Observed sequence (black) alongside the estimated factual hidden sequence (blue), illustrating how even an accurate factual estimation may fail to produce reliable counterfactual trajectories once initial conditions are altered or parameter uncertainty is introduced.}
    \label{fig:LineplotLor}
\end{figure}
\begin{figure}
    \centering
        \begin{minipage}{\textwidth}
        \centering
        \begin{minipage}{0.333\textwidth}
            \centering
            $\Tilde{\bm{\theta}} = {\bm{\theta}}_{\text{true}}$
        \end{minipage}%
        \hfill
        \begin{minipage}{0.333\textwidth}
            \centering
            $\Tilde{\bm{\theta}} = \hat{\bm{\theta}}$
        \end{minipage}%
        \hfill
        \begin{minipage}{0.333\textwidth}
            \centering
            $\Tilde{\bm{\theta}} \sim \mathcal{N}\left(\hat{\bm{\theta}}, \bm{\sigma}_{\bm{\theta}}\right)$
        \end{minipage}
    \end{minipage}
    \begin{minipage}[b]{0.32\textwidth}
        \centering
        \includegraphics[width=\textwidth]{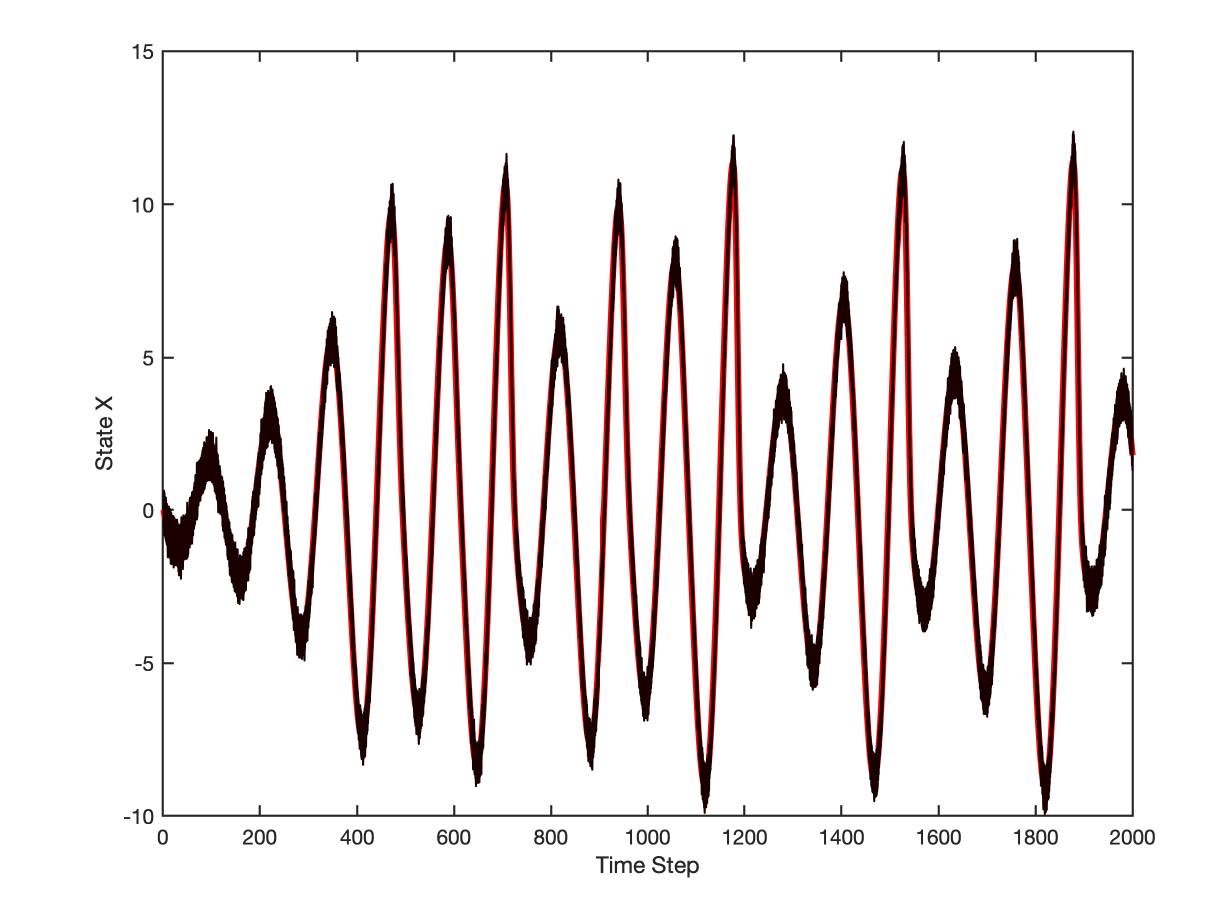} %
    \end{minipage}
    \begin{minipage}[b]{0.32\textwidth}
        \centering
        \includegraphics[width=\textwidth]{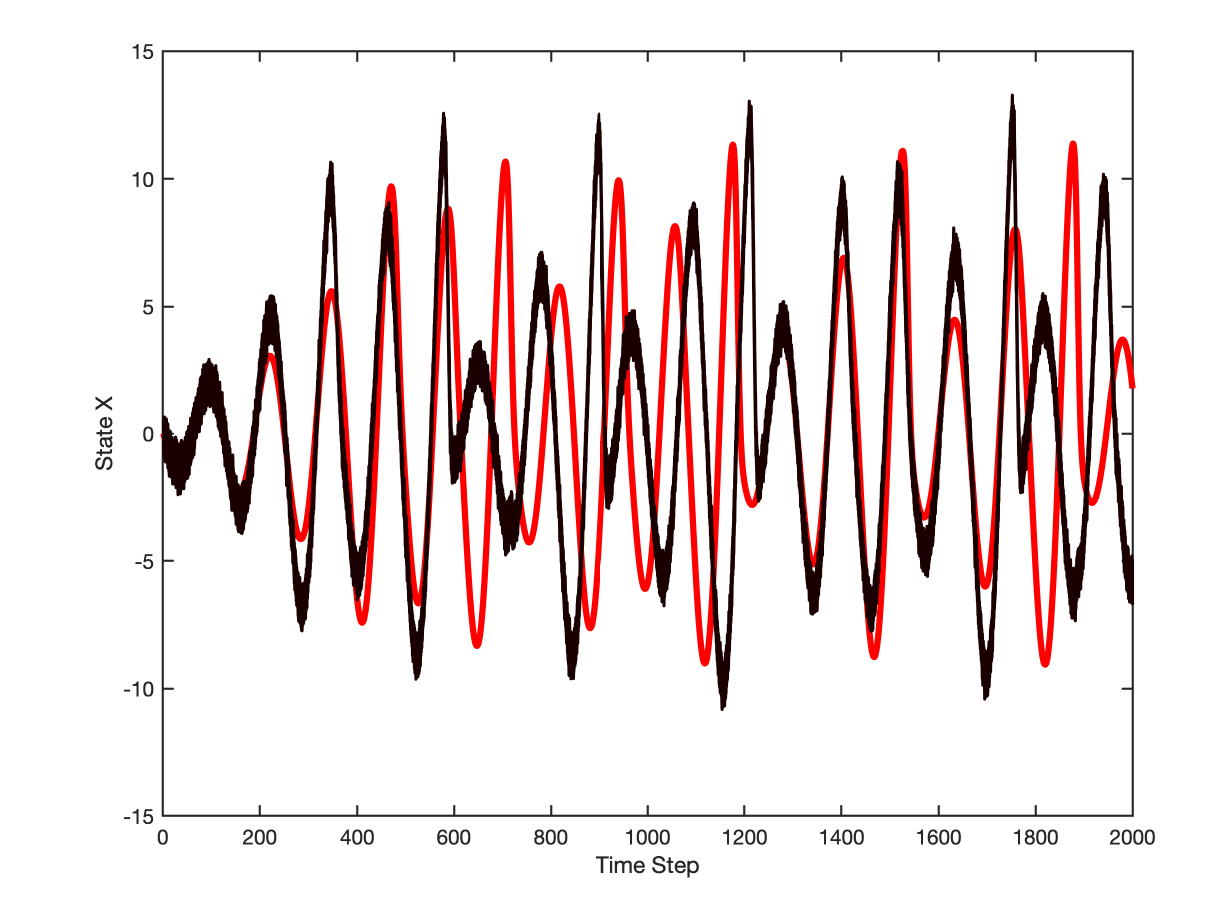} %
    \end{minipage}
        \begin{minipage}[b]{0.32\textwidth}
        \centering
        \includegraphics[width=\textwidth]{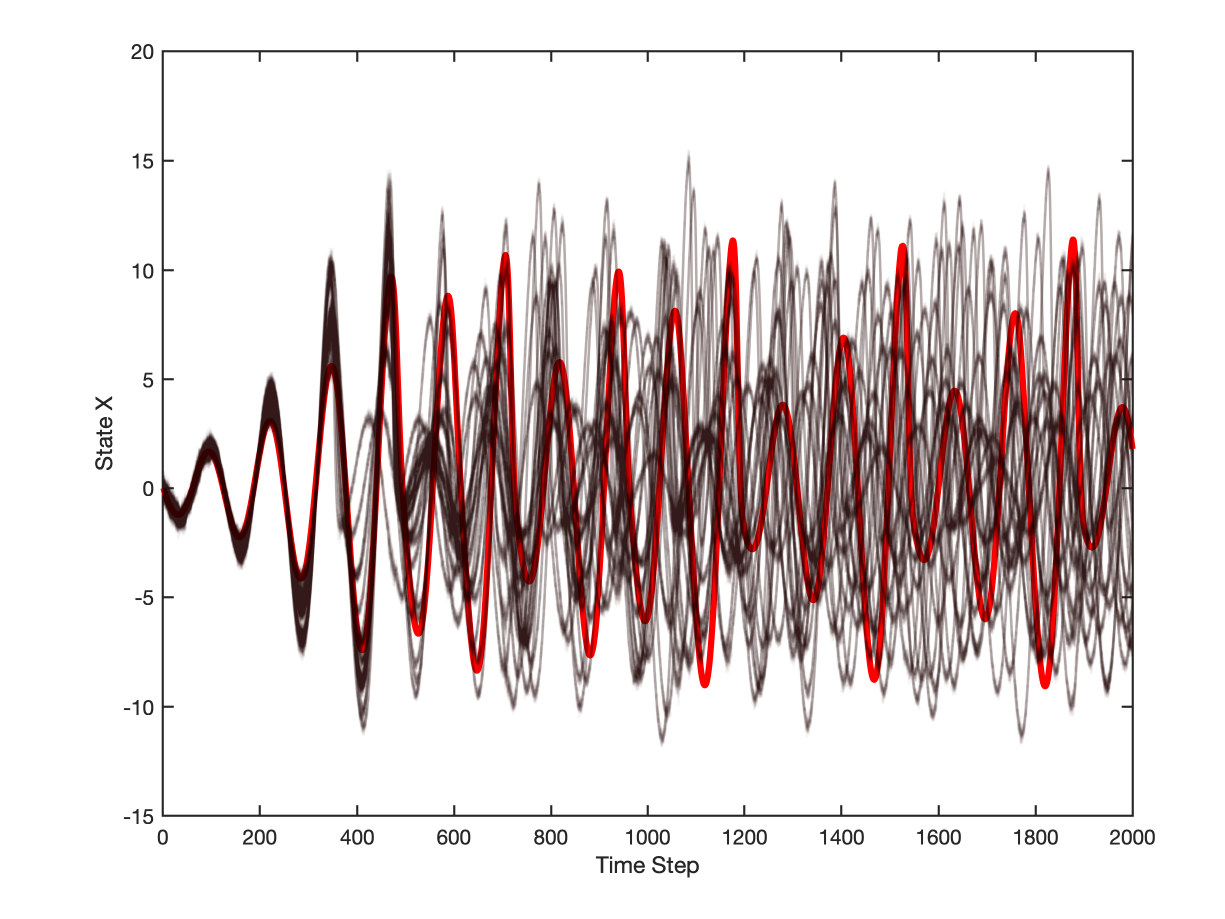} 
    \end{minipage}
    \begin{minipage}[b]{0.32\textwidth}
        \centering
        \includegraphics[width=\textwidth]{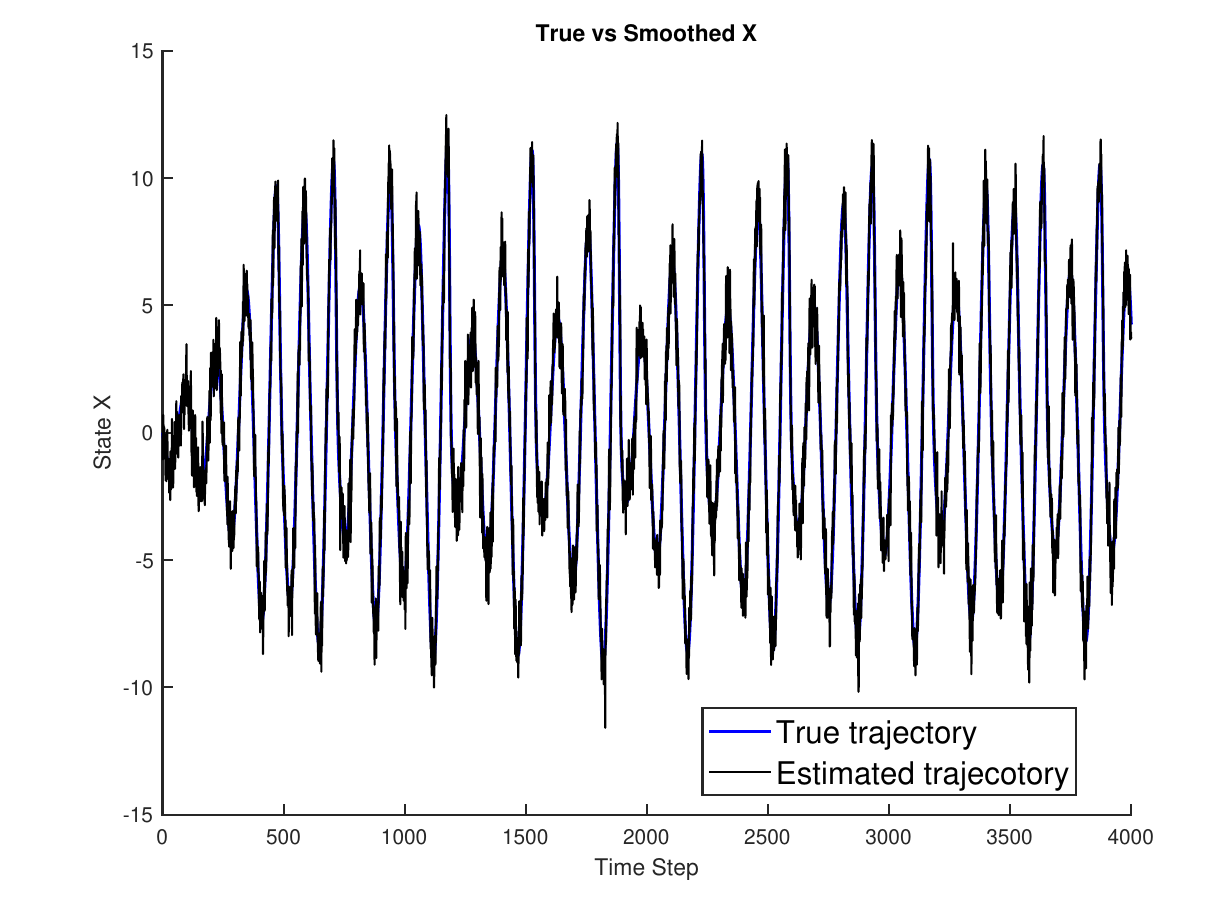} %
    \end{minipage}
    \begin{minipage}[b]{0.32\textwidth}
        \centering
        \includegraphics[width=\textwidth]{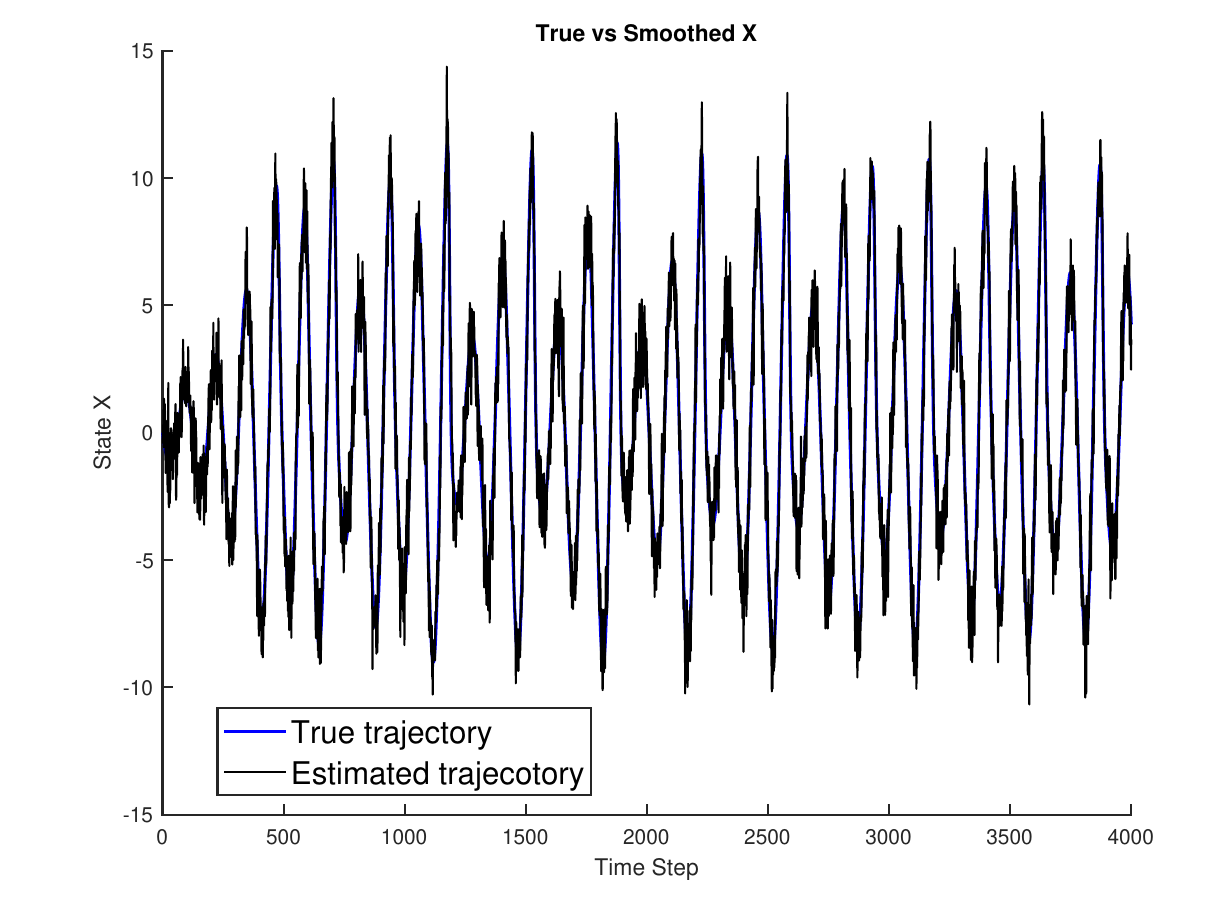} %
    \end{minipage}
        \begin{minipage}[b]{0.32\textwidth}
        \centering
        \includegraphics[width=\textwidth]{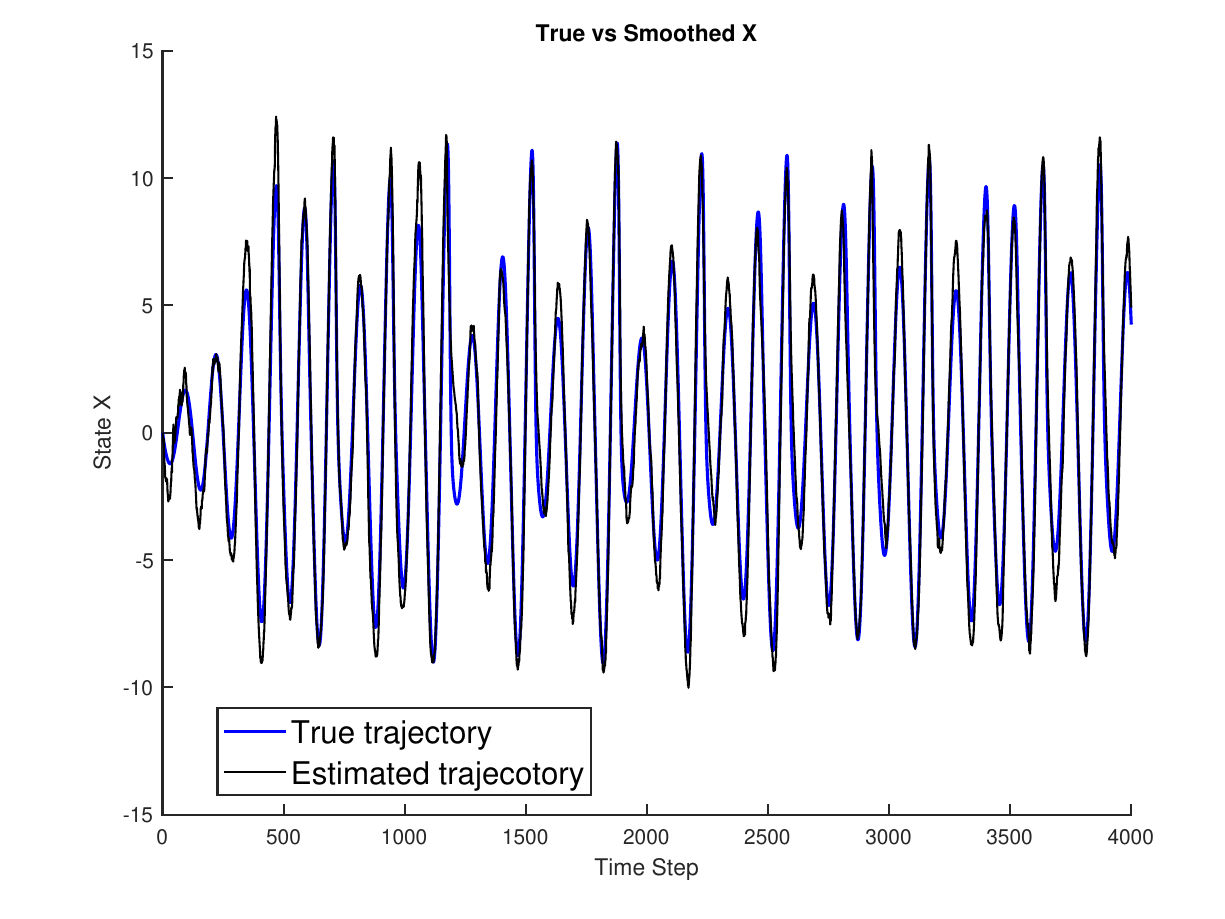} 
    \end{minipage}
    \caption{Top row: Deterministic counterfactual trajectory (in red) compared to the generated counterfactual trajectories (in black) for the R\"{o}ssler system under three parameter assumptions: $\Tilde{\bm{\theta}} = {\bm{\theta}}_{true}$, $\Tilde{\bm{\theta}} = \hat{\bm{\theta}}$, and $\Tilde{\bm{\theta}} \sim \mathcal{N}(\hat{\bm{\theta}}, \bm{\sigma}_{\bm{\theta}})$. Bottom row: Observed sequence (black) alongside the estimated factual hidden sequence (blue), illustrating how even an accurate factual estimation may fail to produce reliable counterfactual trajectories once initial conditions are altered or parameter uncertainty is introduced.}
    \label{fig:LineplotRoss}
\end{figure}
\begin{figure}
    \centering
        \begin{minipage}{\textwidth}
        \centering
        \begin{minipage}{0.333\textwidth}
            \centering
            $\Tilde{\bm{\theta}} = {\bm{\theta}}_{\text{true}}$
        \end{minipage}%
        \hfill
        \begin{minipage}{0.333\textwidth}
            \centering
            $\Tilde{\bm{\theta}} = \hat{\bm{\theta}}$
        \end{minipage}%
        \hfill
        \begin{minipage}{0.333\textwidth}
            \centering
            $\Tilde{\bm{\theta}} \sim \mathcal{N}\left(\hat{\bm{\theta}}, \bm{\sigma}_{\bm{\theta}}\right)$
        \end{minipage}
    \end{minipage}
    \begin{minipage}[b]{0.32\textwidth}
        \centering
        \includegraphics[width=\textwidth]{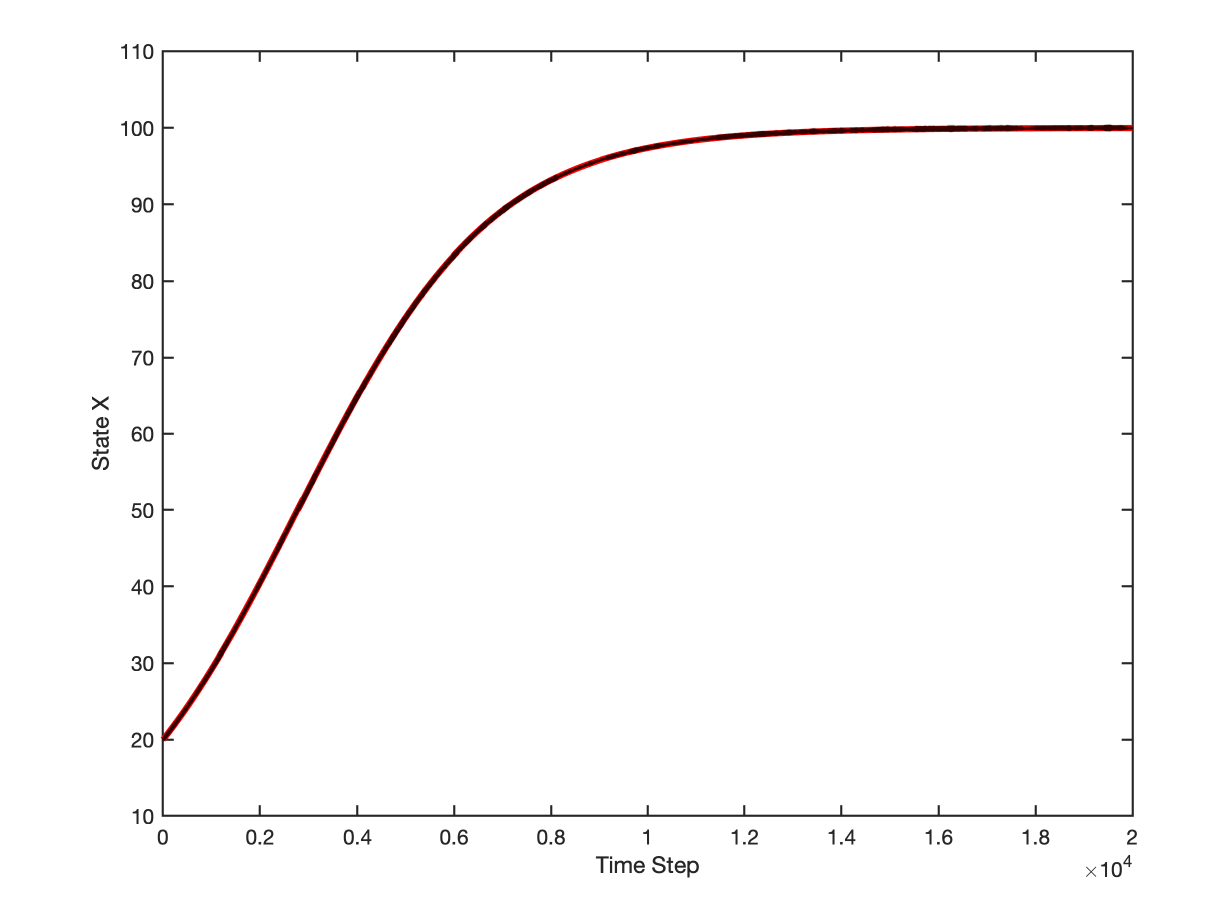} %
    \end{minipage}
    \begin{minipage}[b]{0.32\textwidth}
        \centering
        \includegraphics[width=\textwidth]{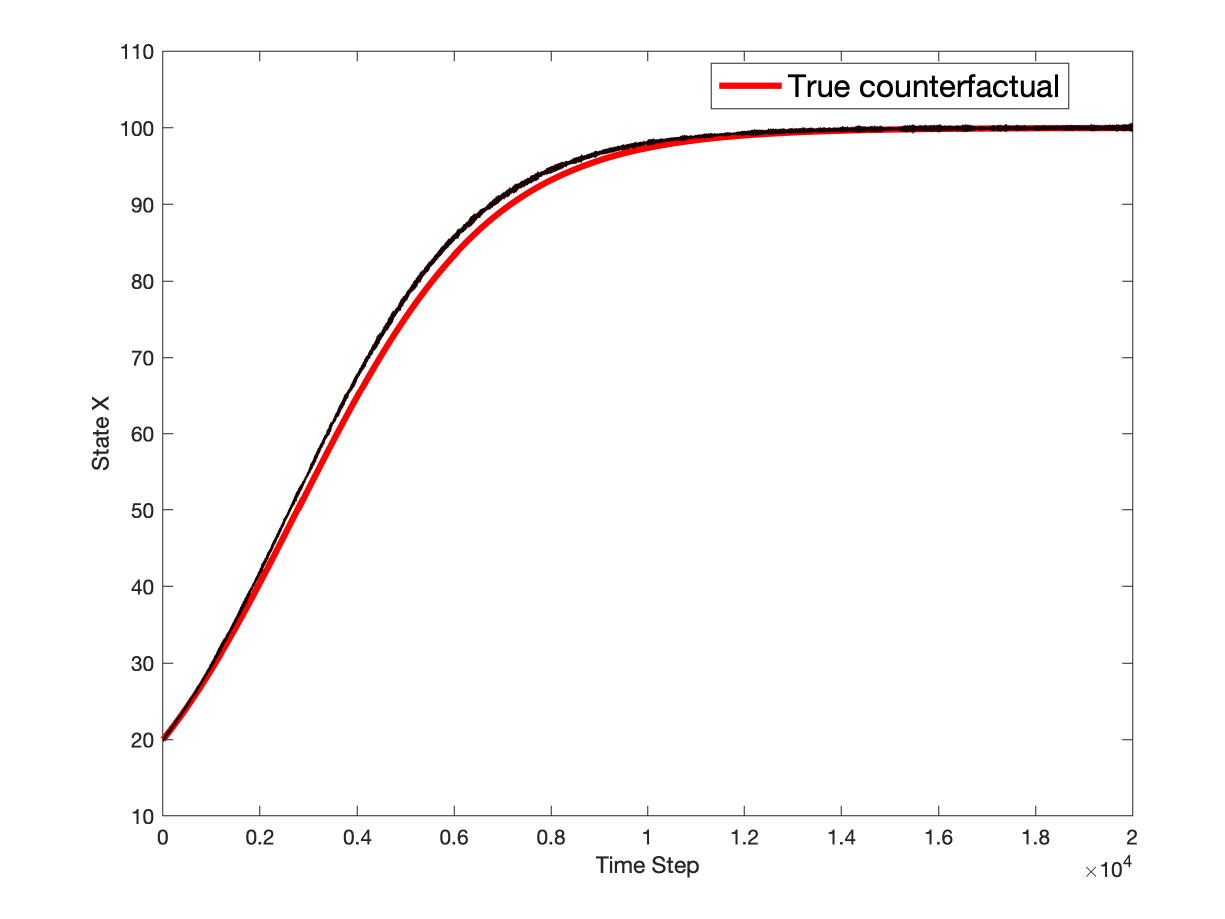} %
    \end{minipage}
        \begin{minipage}[b]{0.32\textwidth}
        \centering
        \includegraphics[width=\textwidth]{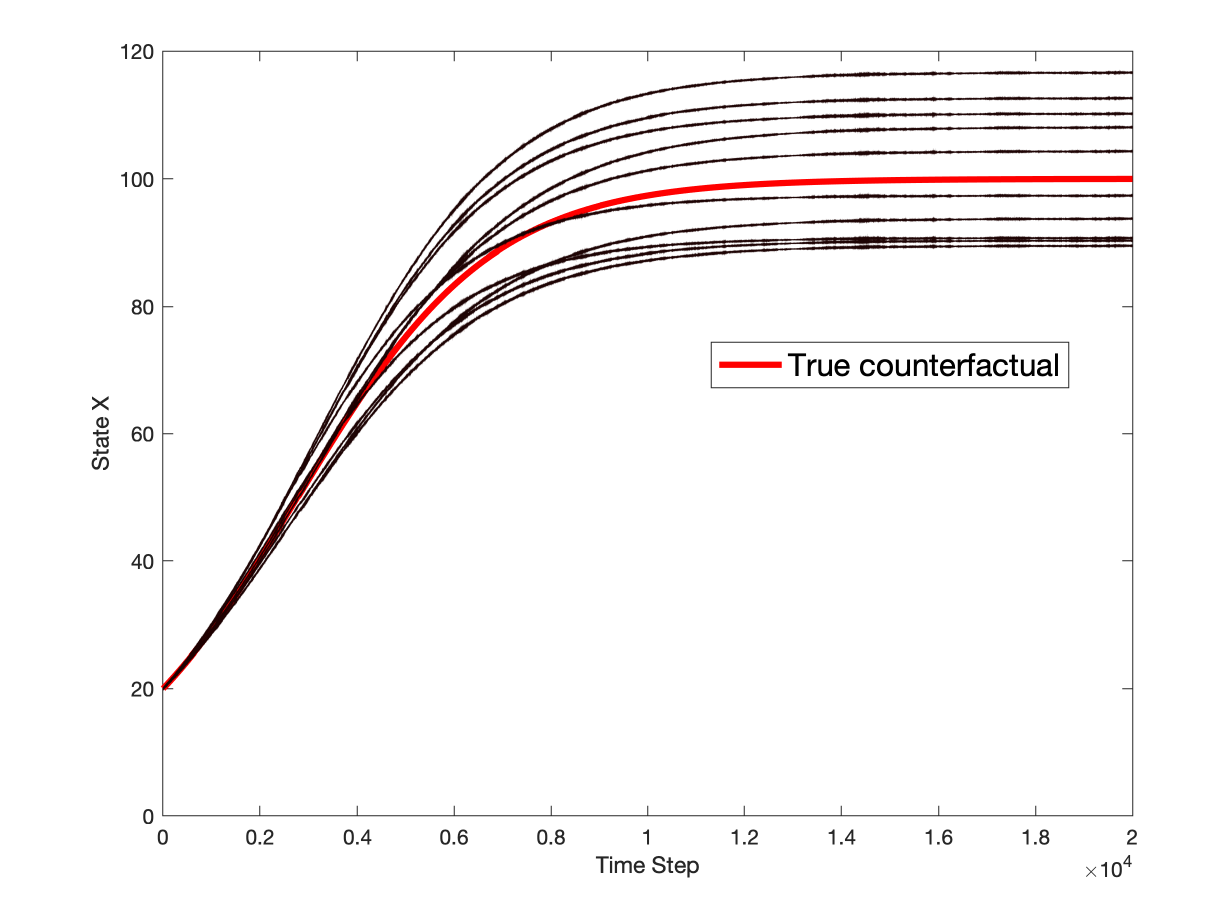} %
    \end{minipage}
    \begin{minipage}[b]{0.32\textwidth}
        \centering
        \includegraphics[width=\textwidth]{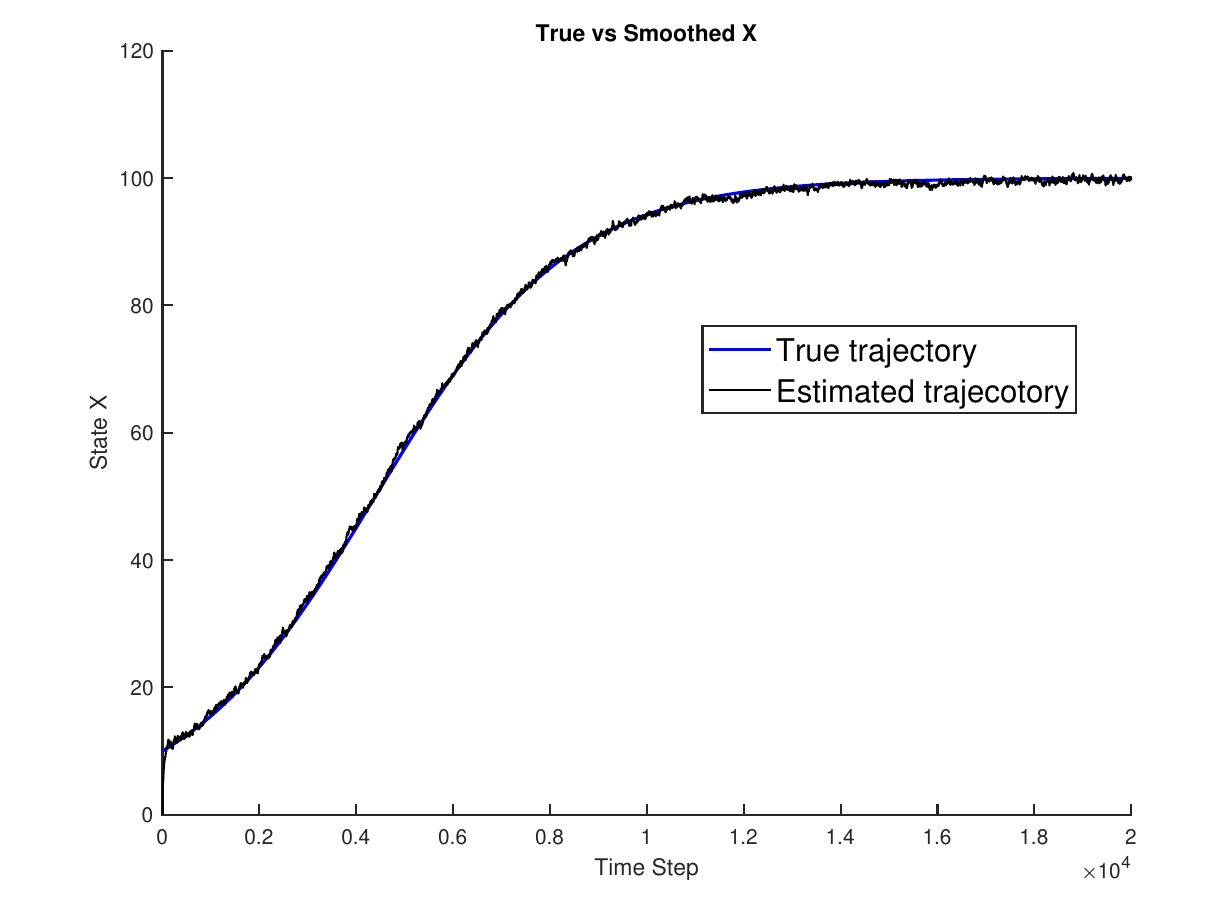} %
    \end{minipage}
    \begin{minipage}[b]{0.32\textwidth}
        \centering
        \includegraphics[width=\textwidth]{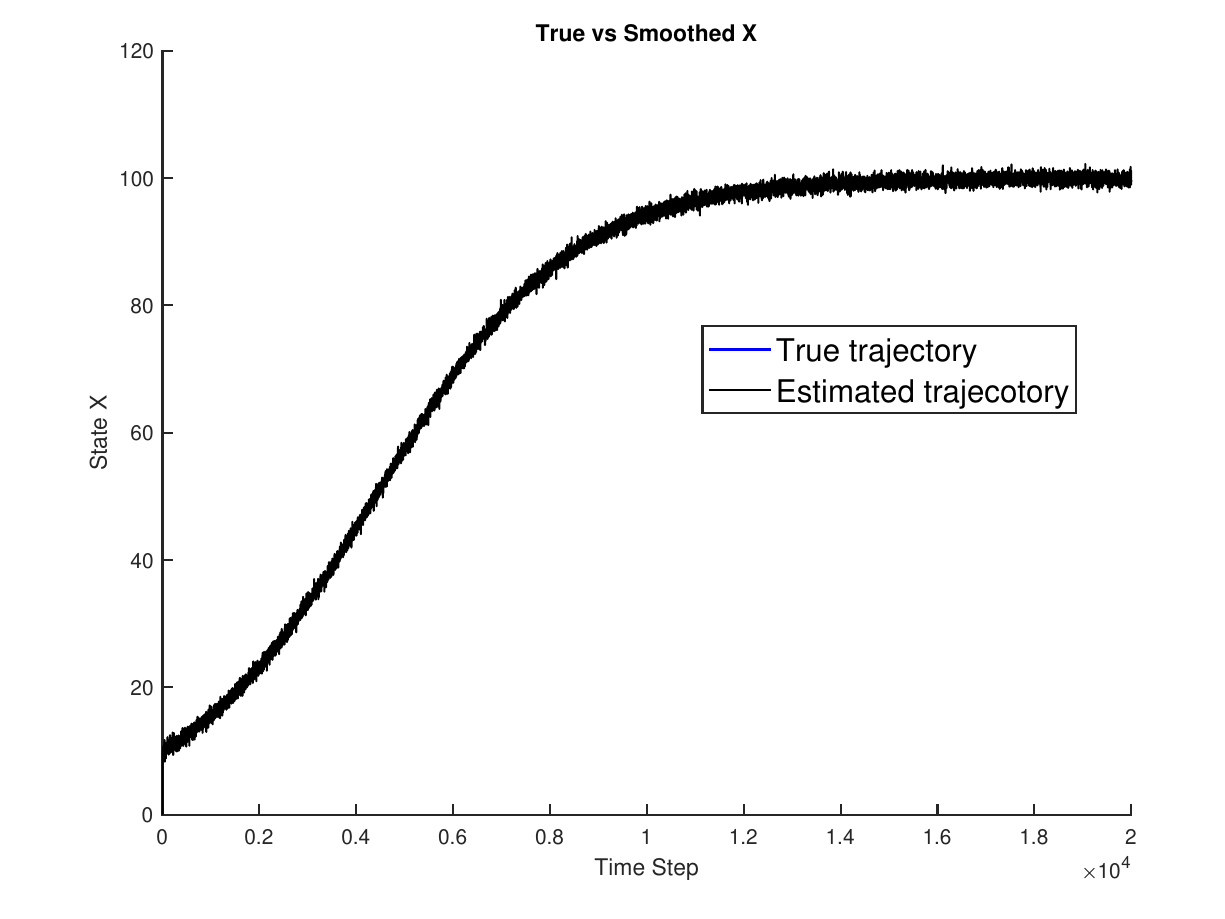} %
    \end{minipage}
        \begin{minipage}[b]{0.32\textwidth}
        \centering
        \includegraphics[width=\textwidth]{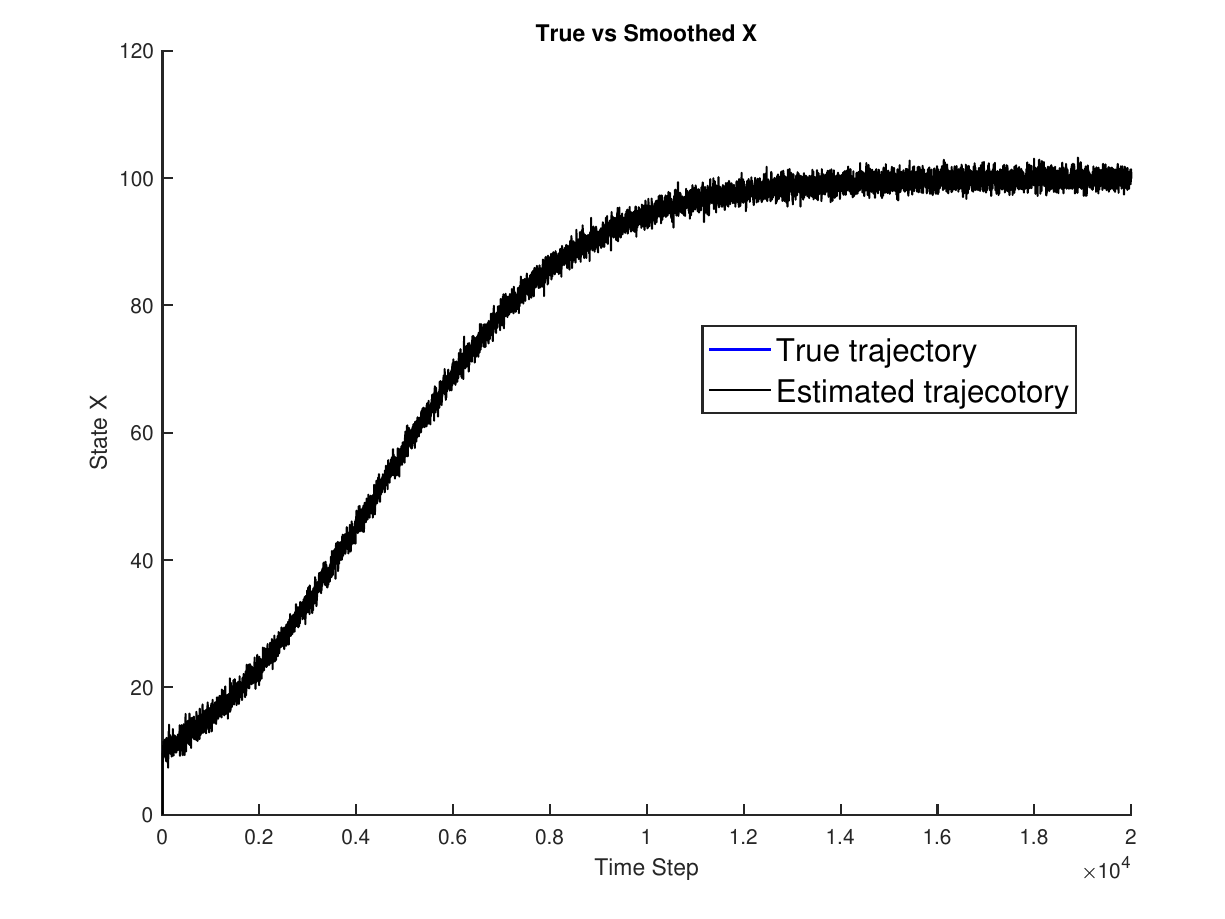} 
    \end{minipage}
    \caption{Top row: Deterministic counterfactual trajectory (in red) compared to the generated counterfactual trajectories (in black) for the logistic growth system under three parameter assumptions: $\Tilde{\bm{\theta}} = {\bm{\theta}}_{true}$, $\Tilde{\bm{\theta}} = \hat{\bm{\theta}}$, and $\Tilde{\bm{\theta}} \sim \mathcal{N}(\hat{\bm{\theta}}, \bm{\sigma}_{\bm{\theta}})$. Bottom row: Observed sequence (black) alongside the estimated factual hidden sequence (blue).}
    \label{fig:LineplotLogit}
\end{figure}

Figure~\ref{fig:LineplotLor}, Figure~\ref{fig:LineplotRoss}, and  represent Figure~\ref{fig:LineplotLogit} results for Lorenz, R\"{o}ssler and Logistic growth systems. The second row displays one-dimensional line plots of the estimated factual trajectory from the observational trajectory, while the first row displays one-dimensional line plots of the generated and deterministic counterfactuals. The first, second and third columns show how system parameters are incorporated in the CF-SCM namely, $\Tilde{\bm{\theta}} = {\bm{\theta}}_{\text{true}}$, $\Tilde{\bm{\theta}} = \hat{\bm{\theta}}$ and $\Tilde{\bm{\theta}} \sim \mathcal{N}(\hat{\bm{\theta}}, \bm{\sigma}_{\bm{\theta}})$, respectively.
\begin{figure}
    \centering
    \begin{minipage}{\textwidth}
        \centering
        \begin{minipage}{0.333\textwidth}
            \centering
            $\Tilde{\bm{\theta}} = {\bm{\theta}}_{\text{true}}$
        \end{minipage}%
        \hfill
        \begin{minipage}{0.333\textwidth}
            \centering
            $\Tilde{\bm{\theta}} = \hat{\bm{\theta}}$
        \end{minipage}%
        \hfill
        \begin{minipage}{0.333\textwidth}
            \centering
            $\Tilde{\bm{\theta}} \sim \mathcal{N}\left(\hat{\bm{\theta}}, \bm{\sigma}_{\bm{\theta}}\right)$
        \end{minipage}
    \end{minipage}
    \begin{minipage}[b]{0.333\textwidth}
        \centering
        \includegraphics[width=\textwidth]{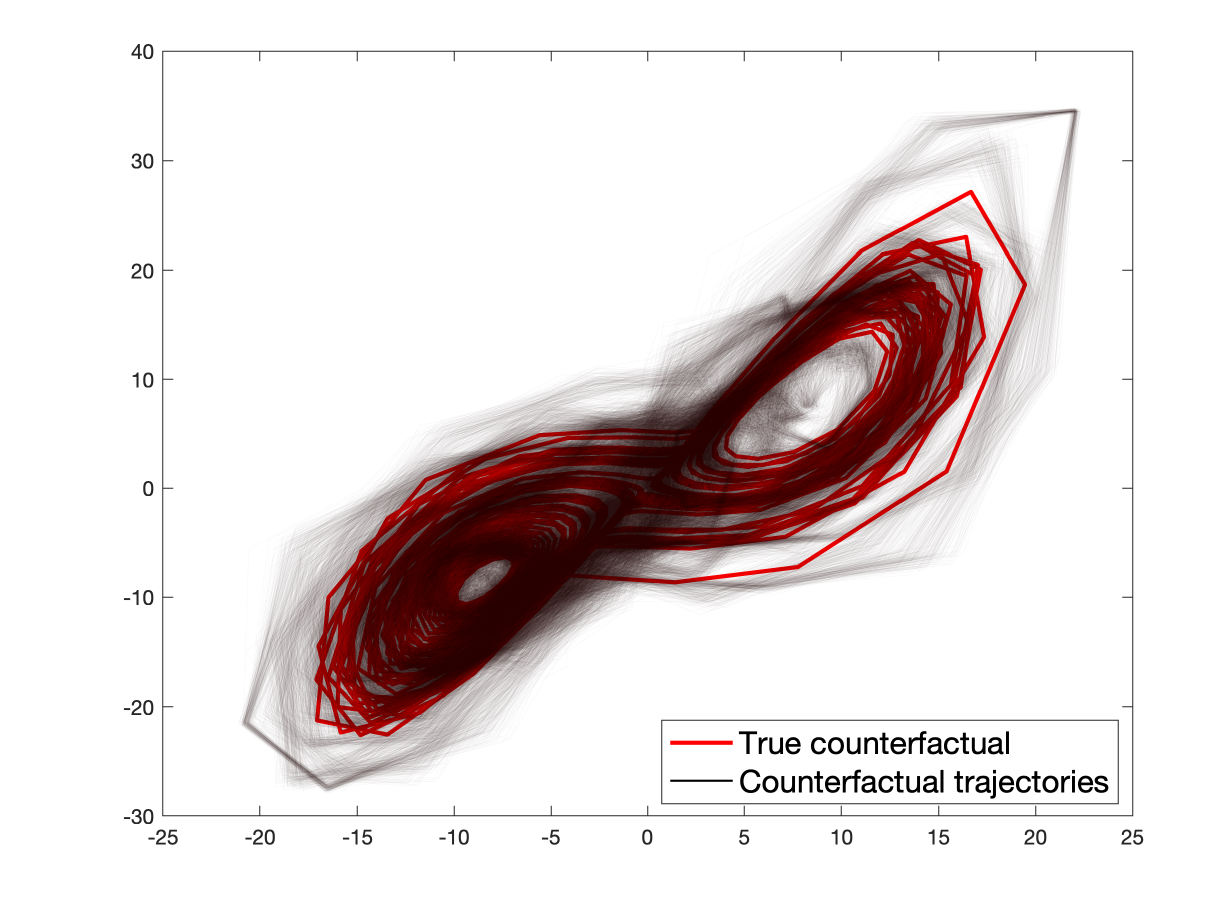}
    \end{minipage}%
    \hfill
    \begin{minipage}[b]{0.333\textwidth}
        \centering
        \includegraphics[width=\textwidth]{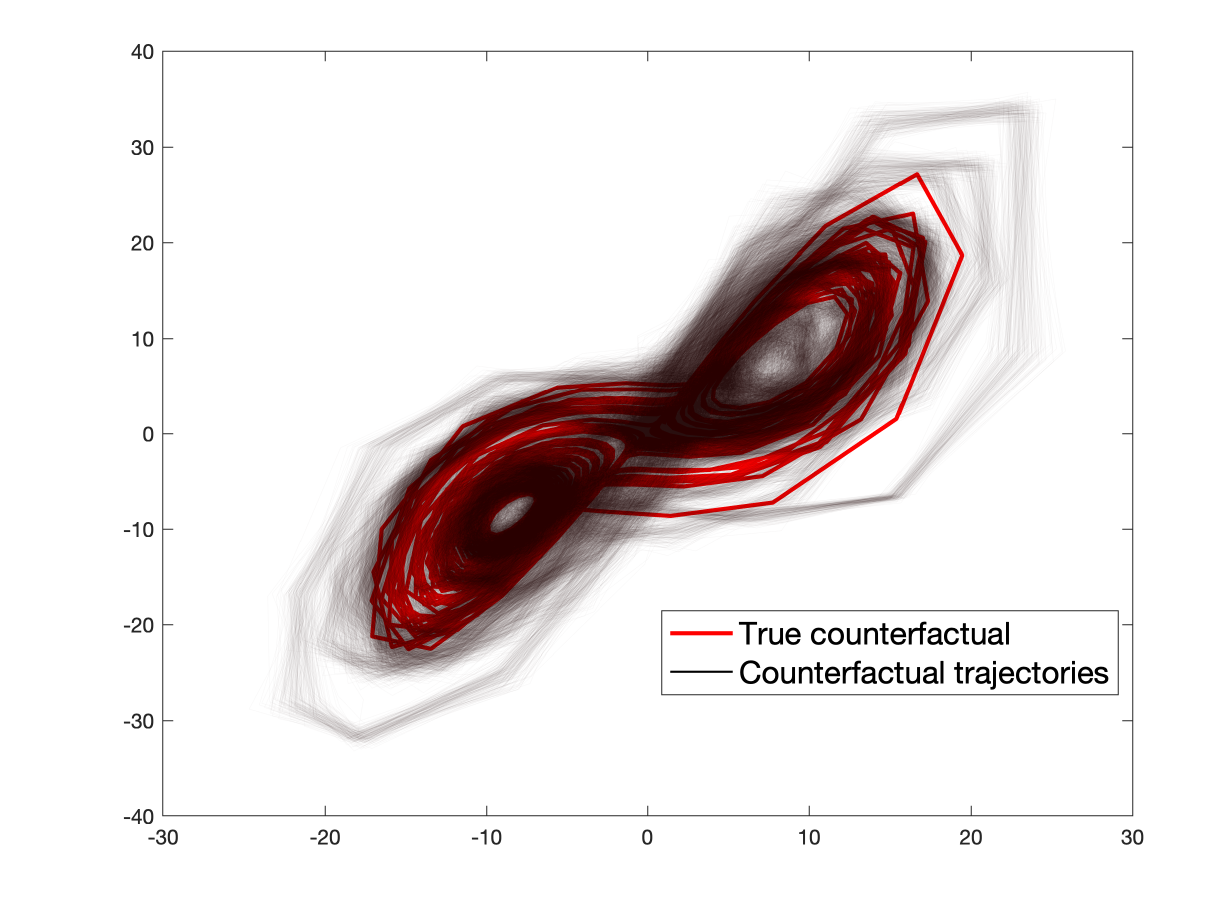}
    \end{minipage}%
    \hfill
    \begin{minipage}[b]{0.333\textwidth}
        \centering
        \includegraphics[width=\textwidth]{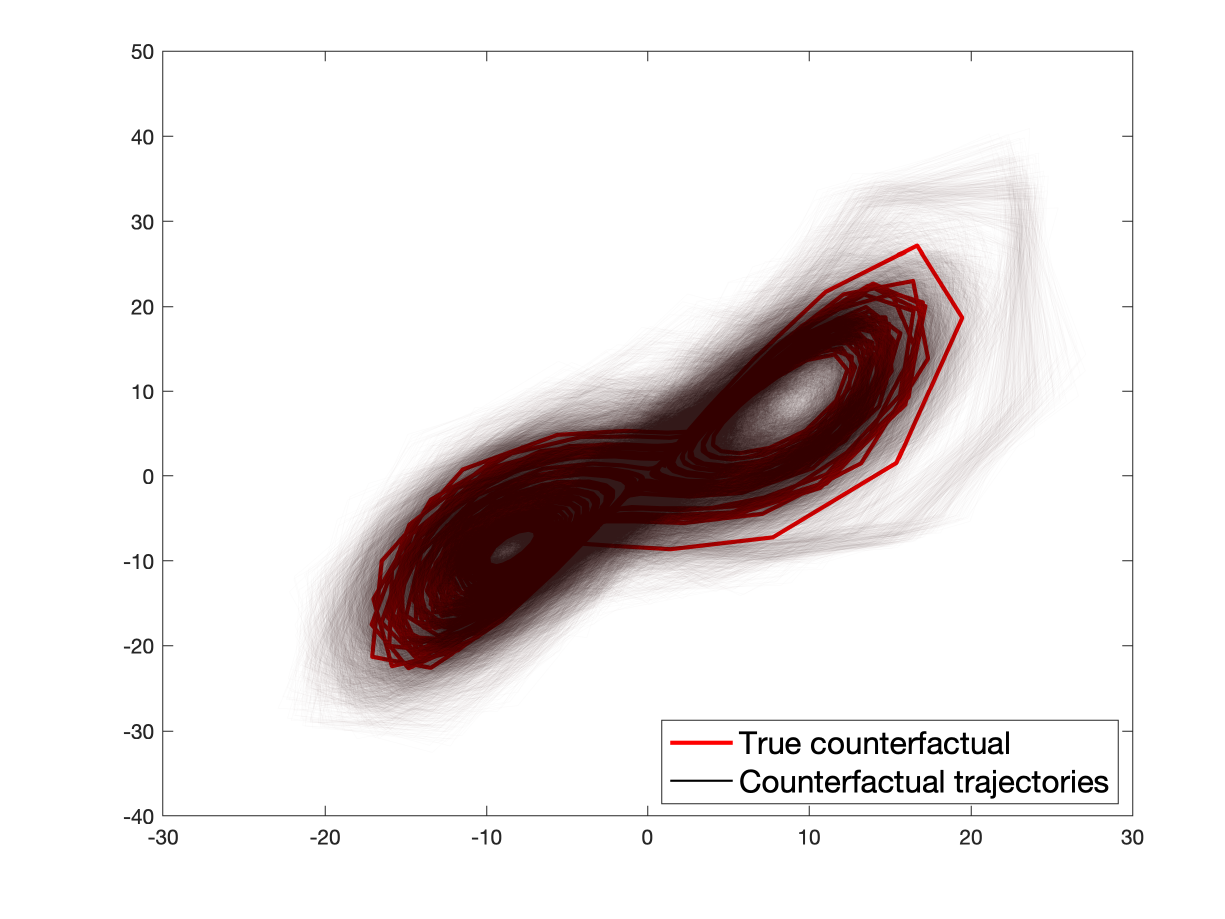}
    \end{minipage}
    \begin{minipage}[b]{0.333\textwidth}
        \centering
        \includegraphics[width=\textwidth]{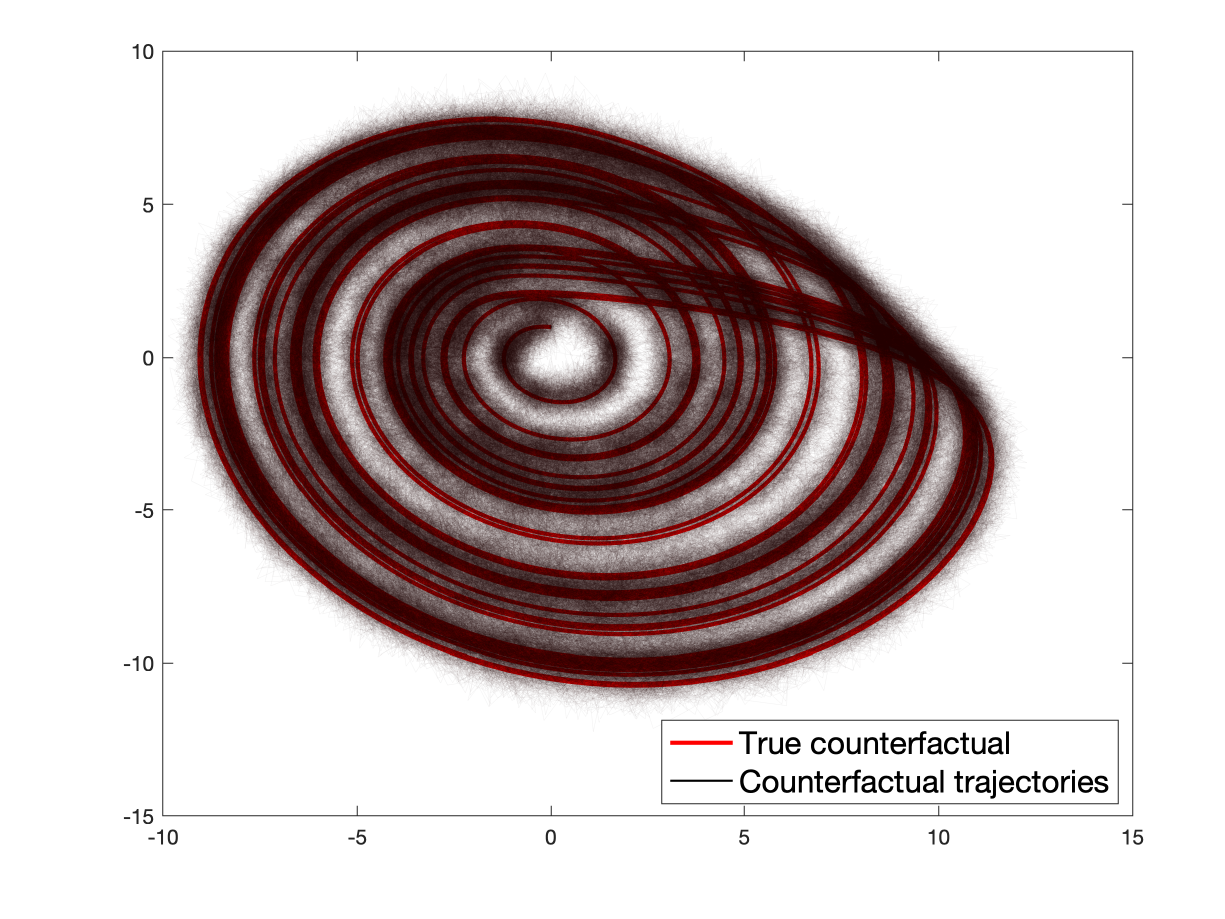}
    \end{minipage}%
    \hfill
    \begin{minipage}[b]{0.333\textwidth}
        \centering
        \includegraphics[width=\textwidth]{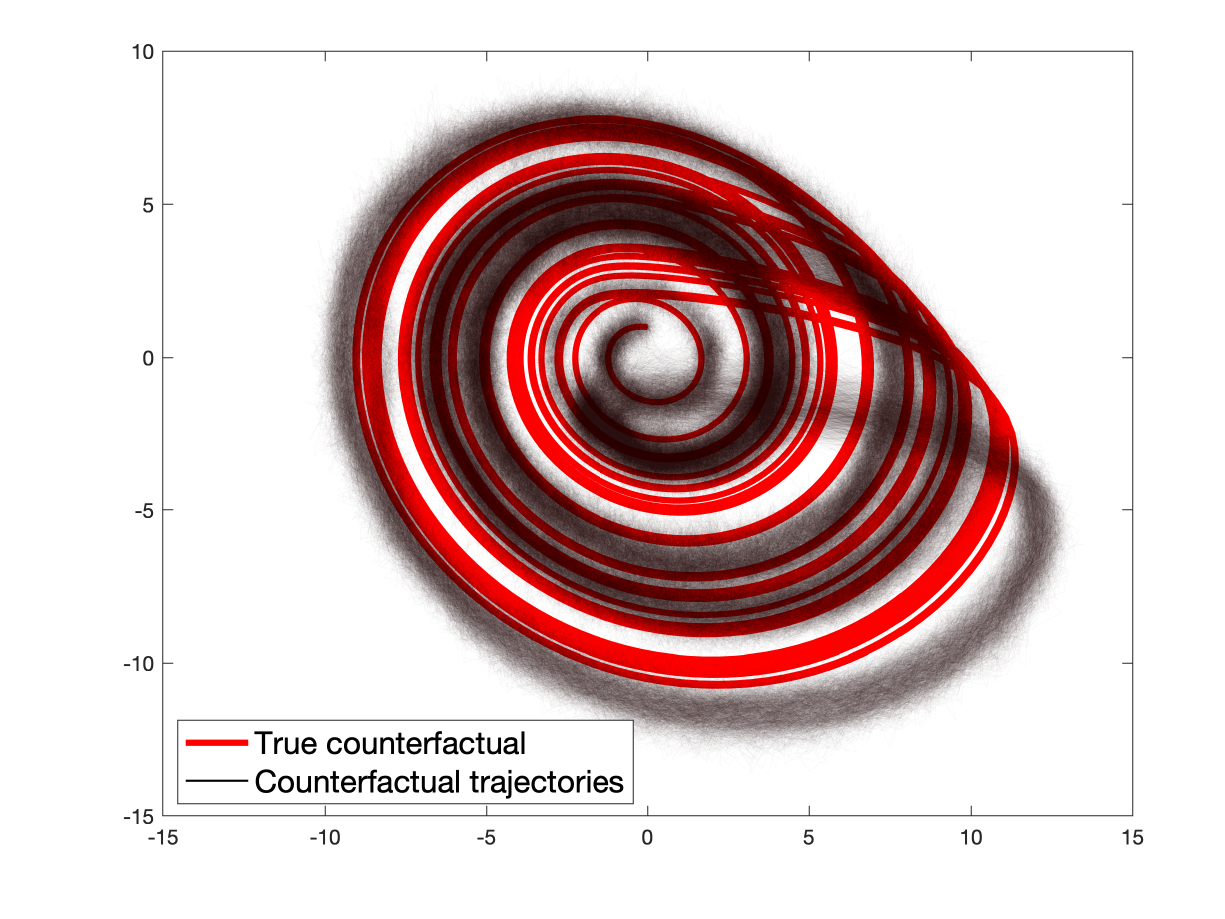}
    \end{minipage}%
    \hfill
    \begin{minipage}[b]{0.333\textwidth}
        \centering
        \includegraphics[width=\textwidth]{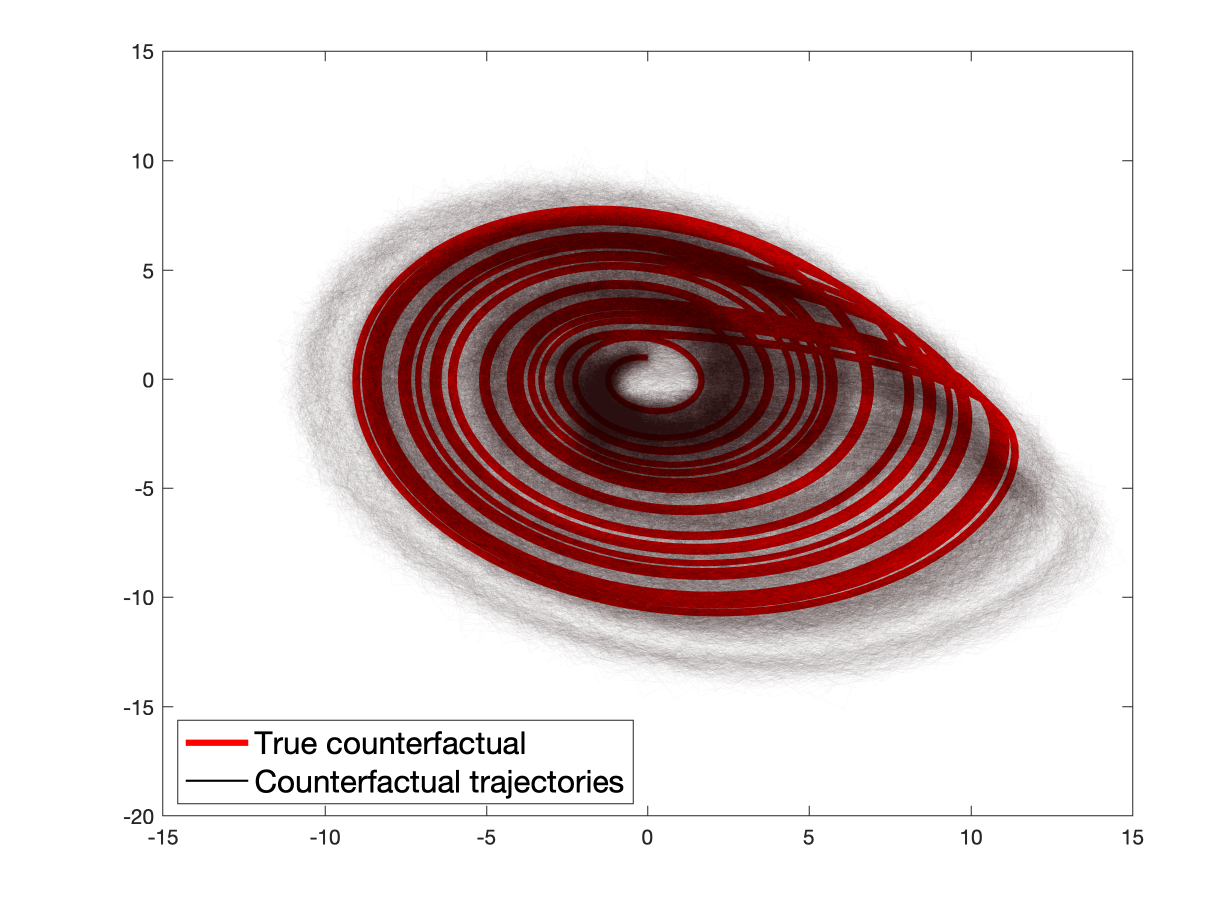}
    \end{minipage}
    \caption{2D plot of counterfactual trajectories (in red) and the deterministic counterfactual trajectory (in black) for Lorenz (first row) and R\"{o}ssler (second row) systems.}
    \label{fig:2d}
\end{figure}
\begin{figure}
    \centering
    \begin{minipage}{\textwidth}
        \centering
        \begin{minipage}{0.333\textwidth}
            \centering
            $\Tilde{\bm{\theta}} = {\bm{\theta}}_{\text{true}}$
        \end{minipage}%
        \hfill
        \begin{minipage}{0.333\textwidth}
            \centering
            $\Tilde{\bm{\theta}} = \hat{\bm{\theta}}$
        \end{minipage}%
        \hfill
        \begin{minipage}{0.333\textwidth}
            \centering
            $\Tilde{\bm{\theta}} \sim \mathcal{N}\left(\hat{\bm{\theta}}, \bm{\sigma}_{\bm{\theta}}\right)$
        \end{minipage}
    \end{minipage}
    \begin{minipage}[b]{0.32\textwidth}
        \centering
        \includegraphics[width=\textwidth]{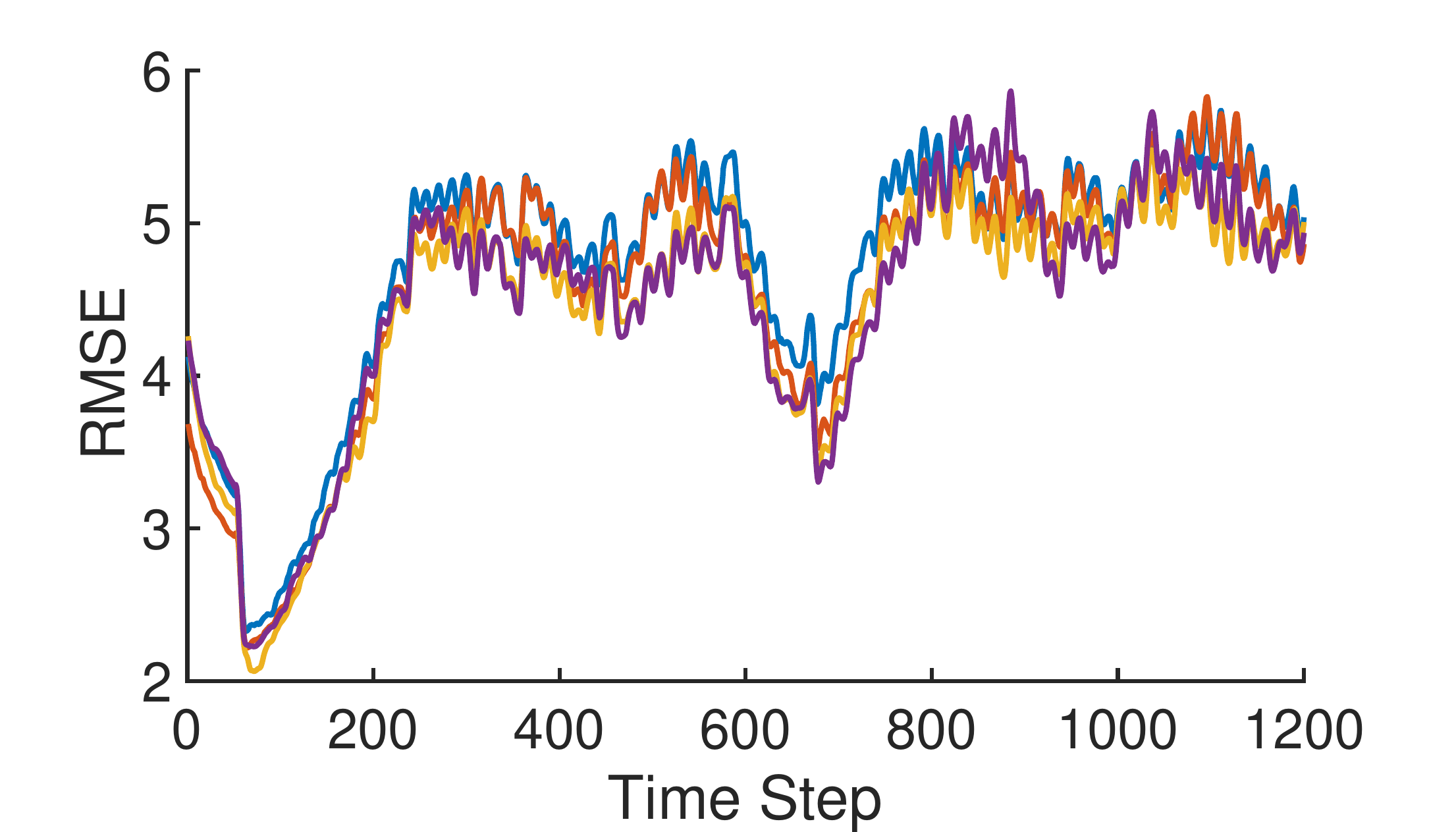} %
    \end{minipage}
    \hfill
    \begin{minipage}[b]{0.32\textwidth}
        \centering
        \includegraphics[width=\textwidth]{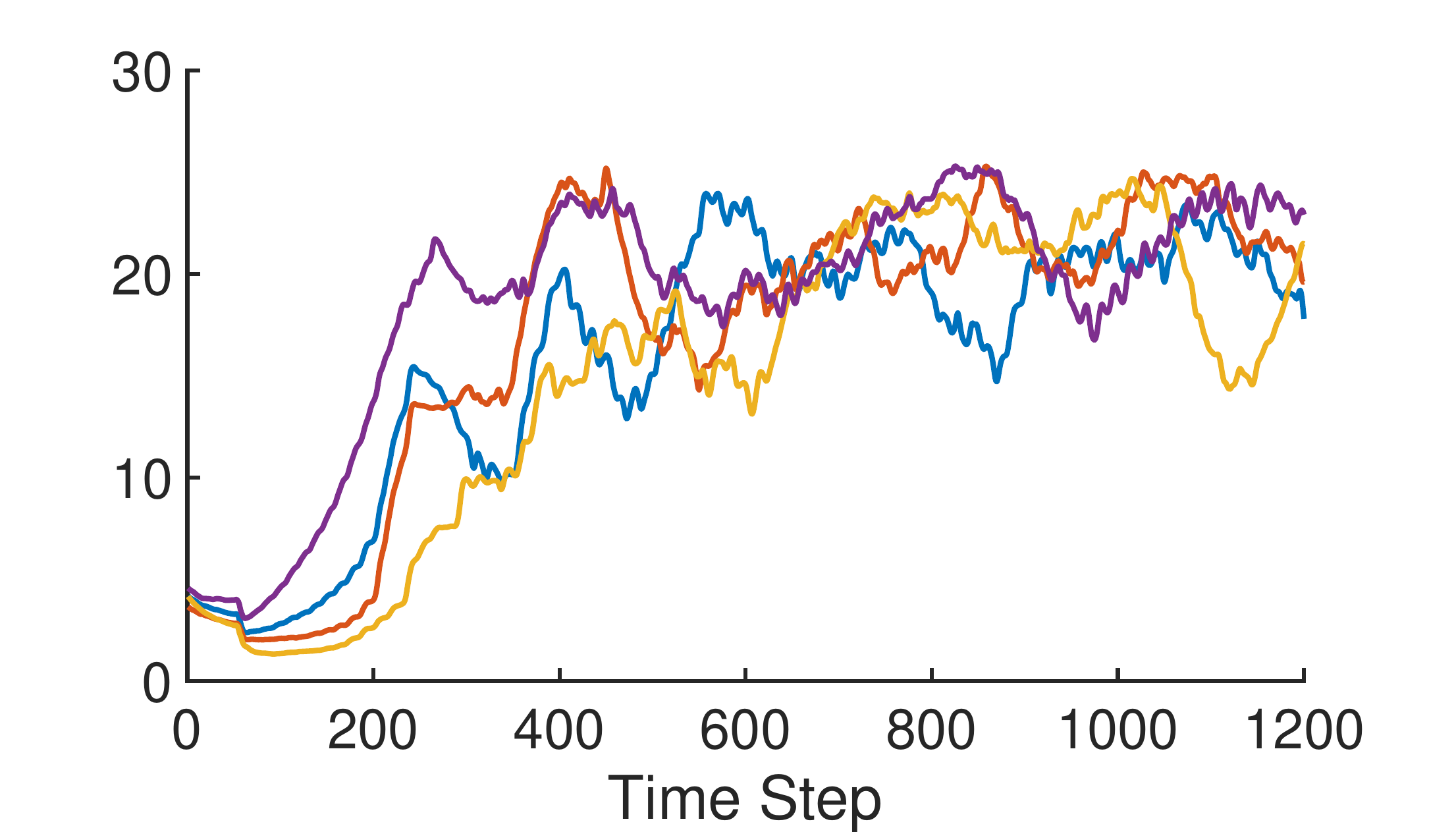} %
    \end{minipage}
    \hfill
    \begin{minipage}[b]{0.32\textwidth}
        \centering
        \includegraphics[width=\textwidth]{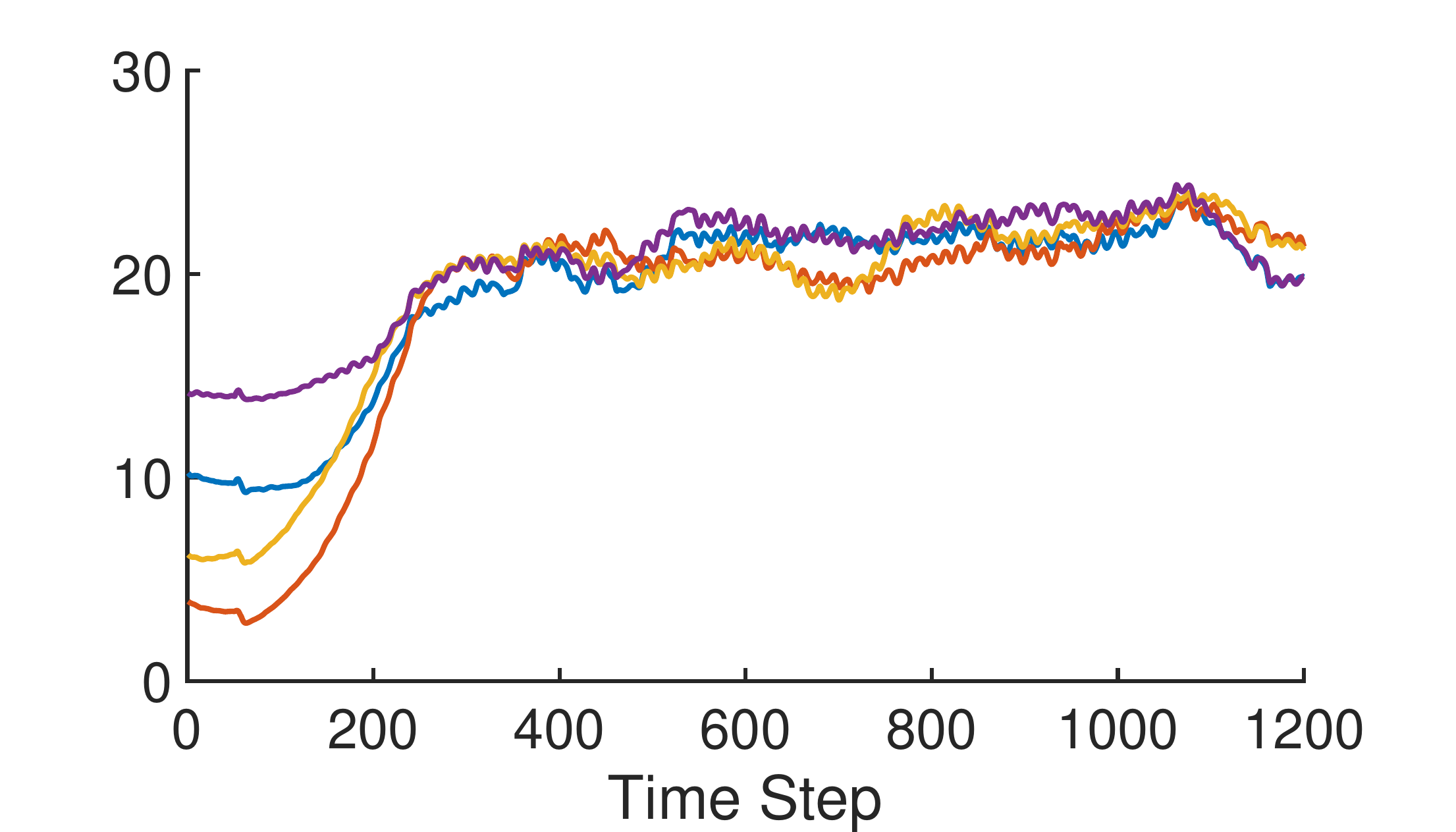} %
    \end{minipage}
    \begin{minipage}[b]{0.32\textwidth}
        \centering
        \includegraphics[width=\textwidth]{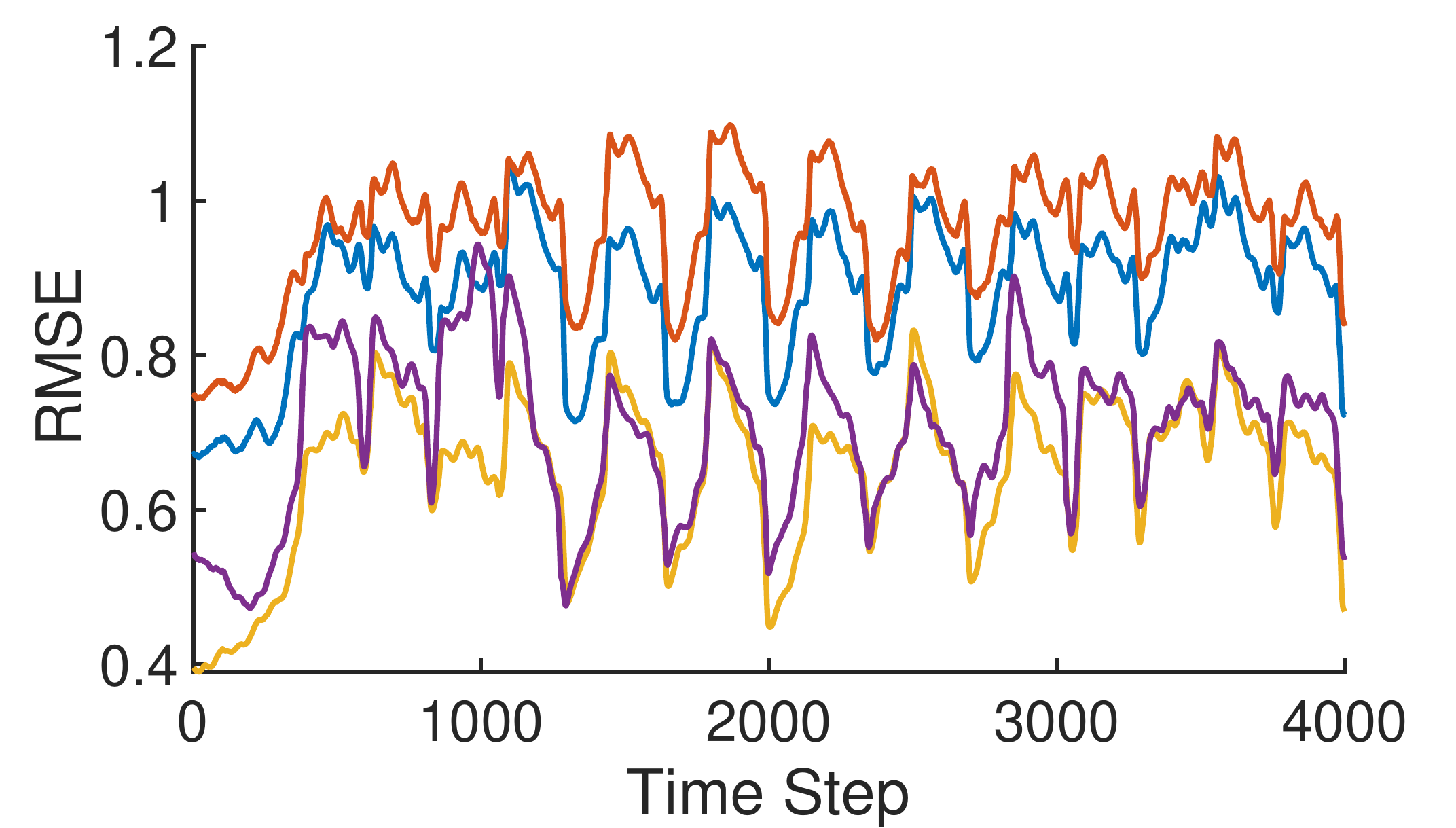} 
    \end{minipage}
    \hfill
    \begin{minipage}[b]{0.32\textwidth}
        \centering
        \includegraphics[width=\textwidth]{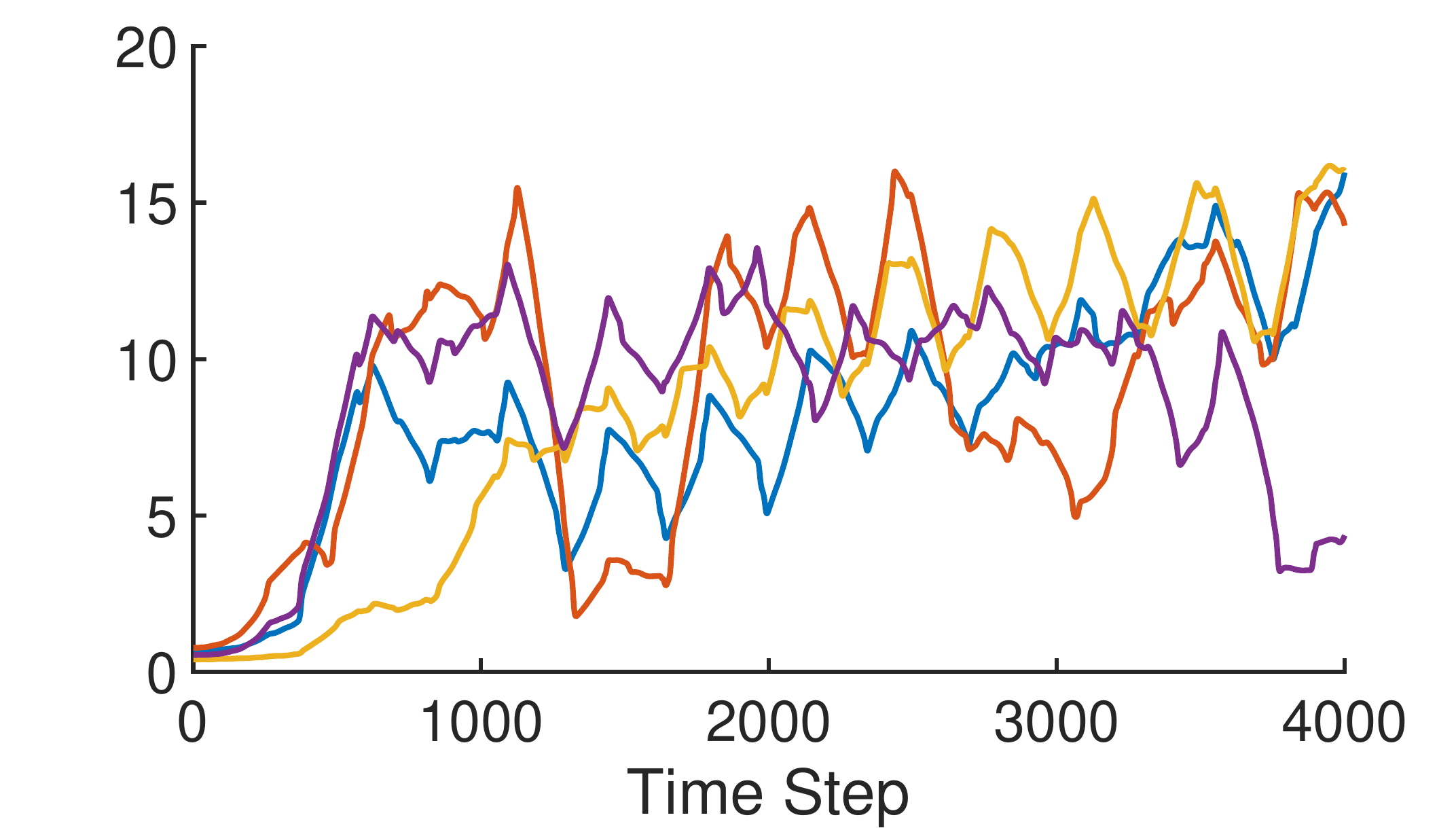} 
    \end{minipage}
    \hfill
    \begin{minipage}[b]{0.32\textwidth}
        \centering
        \includegraphics[width=\textwidth]{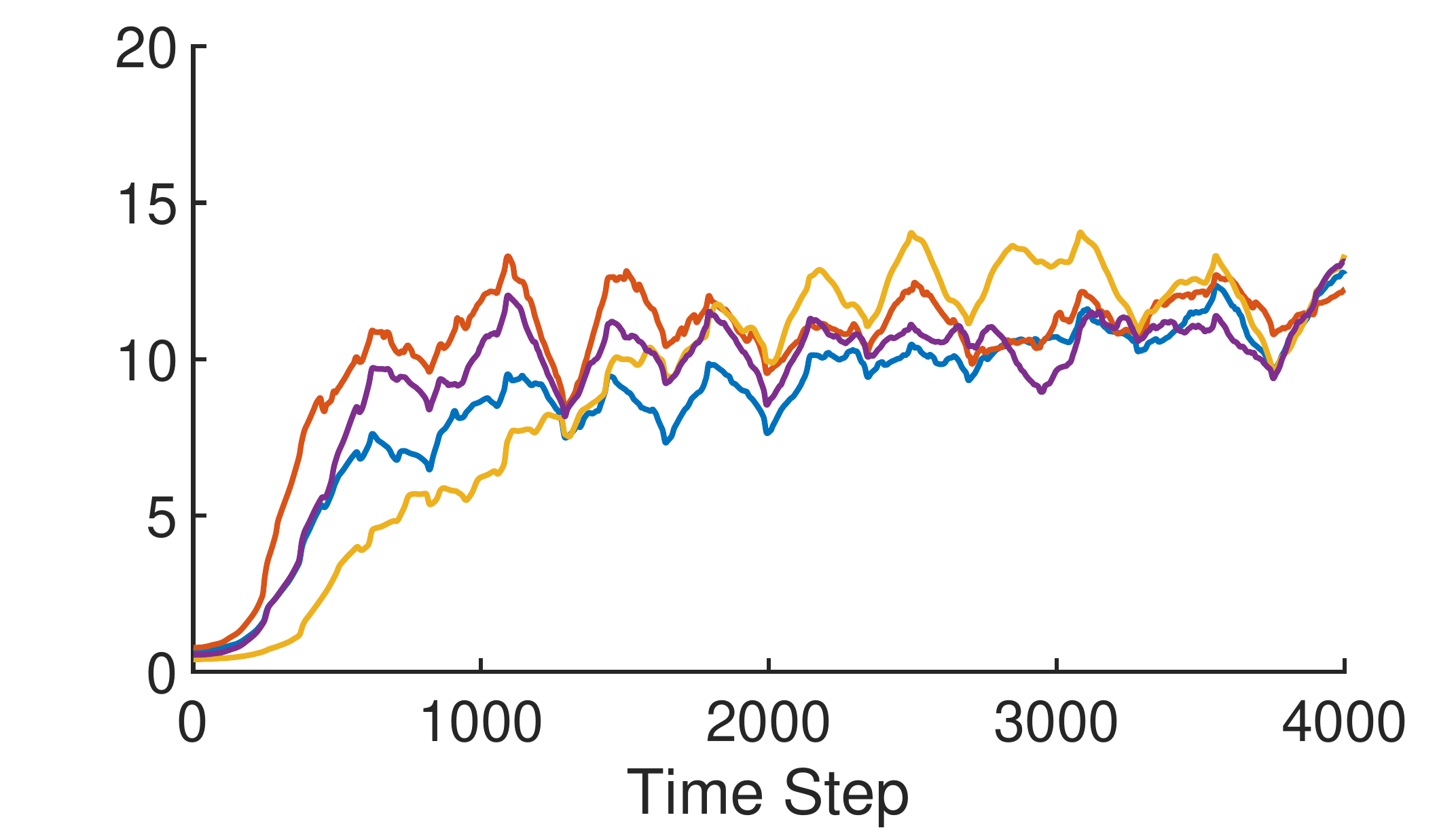} %
    \end{minipage}
    \begin{minipage}[b]{0.32\textwidth}
        \centering
        \includegraphics[width=\textwidth]{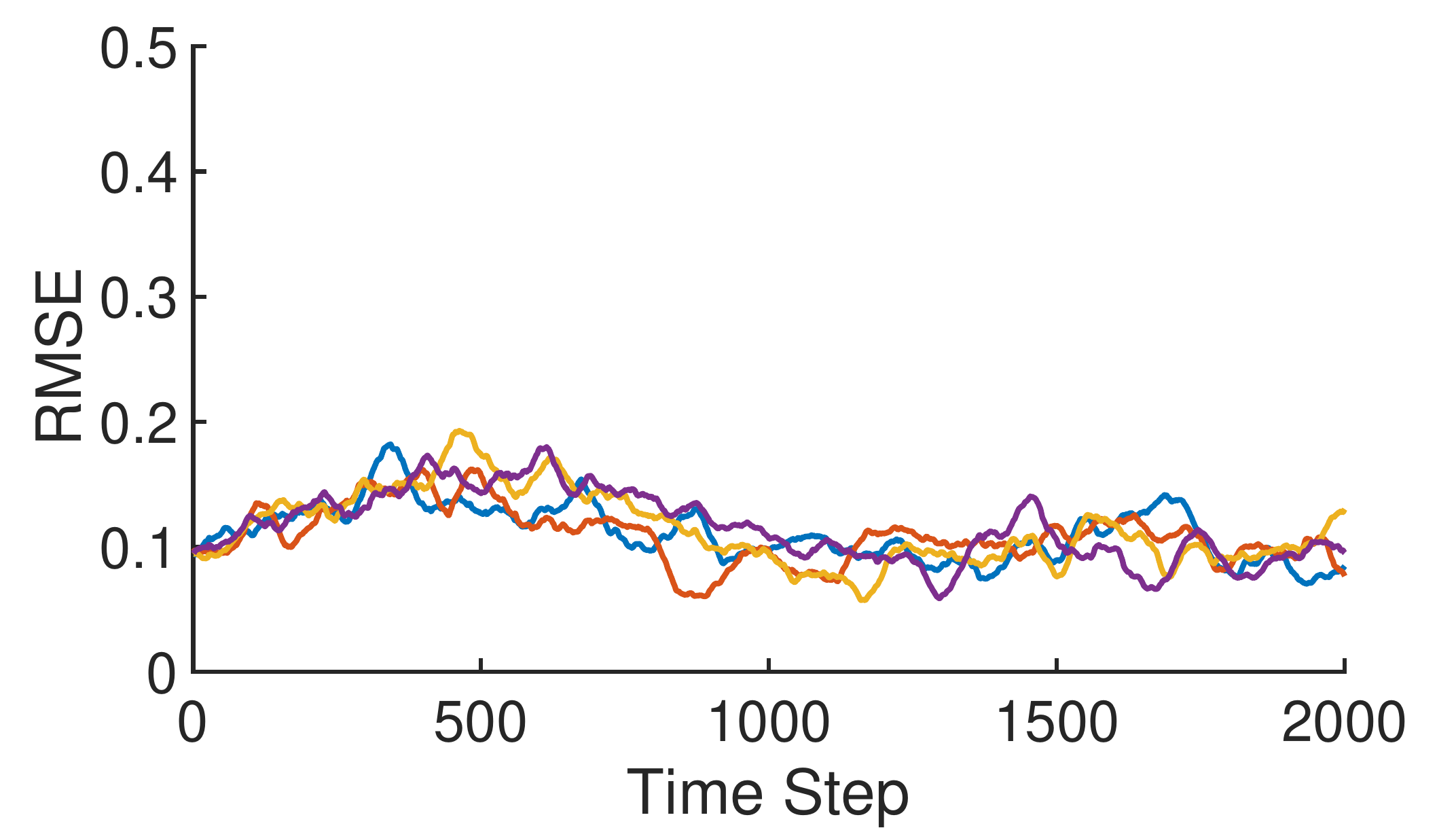} 
    \end{minipage}
    \hfill
    \begin{minipage}[b]{0.32\textwidth}
        \centering
        \includegraphics[width=\textwidth]{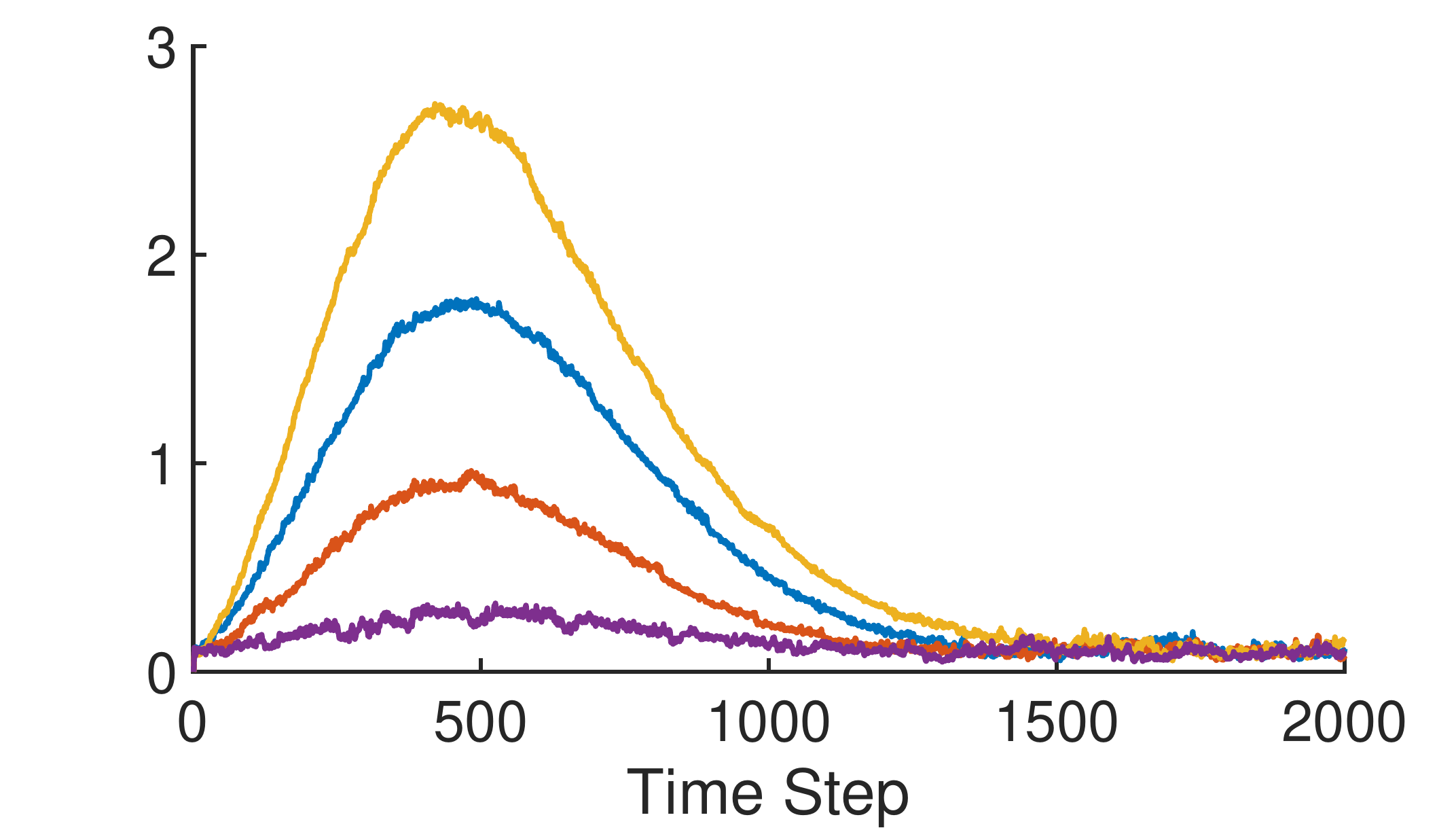} 
    \end{minipage}
    \hfill
    \begin{minipage}[b]{0.32\textwidth}
        \centering
        \includegraphics[width=\textwidth]{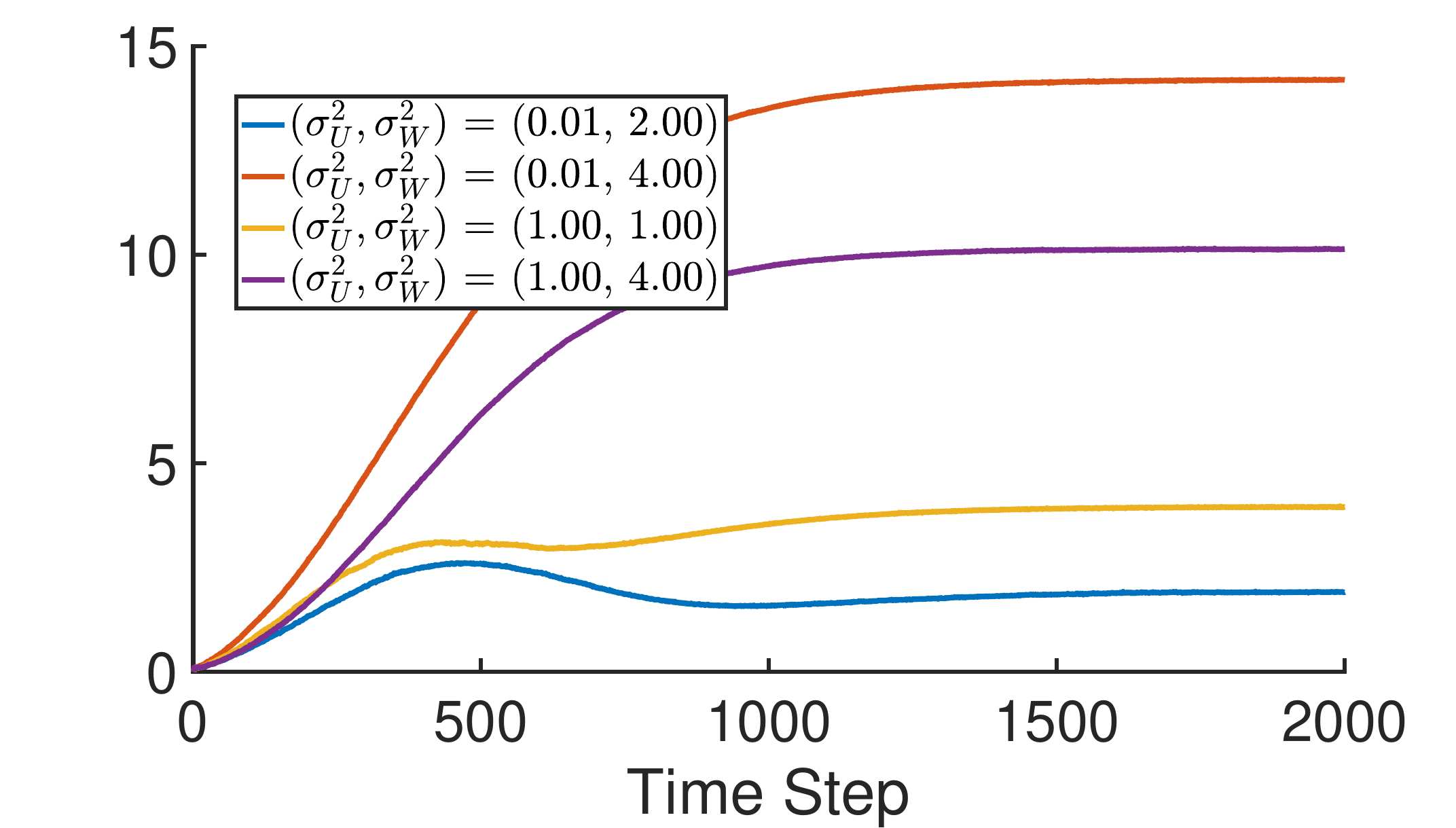} %
    \end{minipage}
    \caption{RMSE$_t$ of counterfactual trajectories generated for four different noise values, corresponding to the Lorenz (first row), R\"{o}ssler (second row), and Logistic growth (third row) systems.}
    \label{fig:rmse}
\end{figure}
Figure~\ref{fig:2d} displays the two-dimensional plot of the deterministic and generated counterfactual trajectories of Lorenz and R\"{o}ssler in the first and second row, respectively. The first, second and third columns show how system parameters are incorporated in the CF-SCM namely, $\Tilde{\bm{\theta}} = {\bm{\theta}}_{\text{true}}$, $\Tilde{\bm{\theta}} = \hat{\bm{\theta}}$ and $\Tilde{\bm{\theta}} \sim \mathcal{N}(\hat{\bm{\theta}}, \bm{\sigma}_{\bm{\theta}})$, respectively. Figure~\ref{fig:rmse} represents the line plot of RMSE$_t$ error as time $t$ progresses. As outlined in Section~\ref{appendix:RMSEt}, the error is calculated based on the generated versus deterministic counterfactuals. Note that RMSE reported is smoothed with a window size of $200$ discrete time steps. In the Lorenz and R\"{o}ssler systems, both known for their chaotic dynamics, we observed a pronounced sensitivity to initial conditions, as shown in the one-dimensional line plots in Figure~\ref{fig:LineplotLor} and Figure~\ref{fig:LineplotRoss}. With parameters set to the true values $\Tilde{\bm{\theta}} = \bm{\theta}_{\text{true}}$, the counterfactual trajectories closely align with the deterministic counterfactual in the initial time steps but soon diverge substantially, which is more pronounced in the two-dimensional representation in Figure~\ref{fig:2d}. This behavior underscores the “butterfly effect,” where small deviations lead to vastly different outcomes over time, even if the underlying parameters are considered to be known a priori.

When parameters are set to point estimates $\Tilde{\bm{\theta}} = \hat{\bm{\theta}}$ or sampled from an approximated posterior distribution $\mathcal{N}(\hat{\bm{\theta}}, \bm{\sigma}_{\bm{\theta}})$, Figures~\ref{fig:LineplotLor}, ~\ref{fig:LineplotRoss} (first line), and \ref{fig:2d} reveal a substantial increase in trajectory divergence. In Figure~\ref{fig:rmse}, the RMSE over time indicates a significant and persistent error for chaotic systems, particularly in the second and third columns, where parameter uncertainties are introduced. This behavior is indeed attributed to sensitive dependence on initial conditions, a key attribute in chaotic systems. However, the second line of Figure~\ref{fig:LineplotLor} and Figure~\ref{fig:LineplotRoss} explicitly shows the corresponding factual trajectories estimated from observational sequences. This reinforces that, while parameter inference techniques can effectively approximate system dynamics, even slight inaccuracies in parameter estimates can drastically alter the reliability of counterfactual sequences in chaotic systems. Figure~\ref{fig:rmse} also illustrates that as noise levels increase, the RMSE remains consistently high throughout the trajectory, especially in the Lorenz and R\"{o}ssler systems. This high RMSE reflects the compounding effect of noise in a chaotic system, where small errors in state estimation quickly amplify, thereby reducing counterfactual reliability. 
Here, the logistic growth system serves as a baseline for comparison with the chaotic Lorenz and R"{ossler} models, precisely because it is not inherently chaotic. The counterfactual trajectories in the first row of Figure~\ref{fig:LineplotLogit} and the recorded errors in Figure~\ref{fig:rmse} show that when the parameters are known or shifted ever so slightly, the counterfactual sequences remain aligned with the deterministic trajectory. In the third row, uncertainty and noise are evident in the spread of counterfactuals, yet the system moves into a stable state by around time step~(1000), keeping deviations within acceptable bounds. However, we stress that these experiments rely on strong assumptions of additive Gaussian noise and perfect structural knowledge of the ODE, made intentionally to isolate and highlight how chaos interacts with noise and uncertainty. While non‐chaotic systems may appear to yield more robust counterfactual predictions under these conditions, real‐world settings that lack such idealized knowledge—and that may involve additional confounders or more complex noise processes—remain challenging for counterfactual reasoning.

Notably, the Lorenz system’s characteristic sensitivity appears around time step 500 in Figure~\ref{fig:LineplotLor}, even when parameters are set to their true values, because slight differences in initial conditions rapidly accumulate. However, once parameter uncertainty is introduced, these minor shifts in parameter space exacerbate the system’s natural divergence, causing counterfactual trajectories to deviate significantly from the deterministic baseline. This effect emerges even earlier, near time steps 300 and 10 in the second and third columns of Figure~\ref{fig:LineplotLor}—underscoring how parameter uncertainty can trigger divergence well before the 500-step mark that would otherwise be observed under true parameter settings.
\section{Discussion, Conclusions and Future Work}\label{sec:conclusions}
We use chaotic systems like Lorenz and R\"{o}ssler models precisely to illustrate the fragility of counterfactual reasoning in dynamic settings where uncertainty, hidden states, and chaos are involved. In many real-world scenarios, the true states of systems are hidden, observations are noisy, and system parameters are uncertain. One might not even be aware that a system exhibits chaotic behavior. Therefore, we focus on using chaotic dynamical systems even under strong assumptions such as (1) known structure, (2) access to the full observation sequence, and (3) controlled prior distributions of the parameters. When counterfactual reasoning is applied even in such controlled contexts, the inherent unpredictability and sensitivity to initial conditions in chaotic systems can lead to significant deviations in the outcomes of counterfactual sequences, occurring well before any divergences appear in the predicted factual sequences. This demonstrates that counterfactual reasoning can be particularly fragile in the presence of chaos and uncertainty, highlighting the need for caution when applying these methods to complex dynamical systems.
All models, including Structural Causal Models (SCMs), are approximations of reality. Computing counterfactuals relies on SCMs. A key insight is that while an SCM may suffice for prediction or causal inference, it may not be reliable for counterfactual estimation. Common approximations—like parameter uncertainties, complex underlying dynamics, or simplifying abstractions—can undermine the reliability of counterfactual estimations.

Our results indicate that there are fundamental theoretical and practical limitations to computing counterfactuals, which require further refinement in future work. These limitations could be particularly relevant in fields such as medicine, where counterfactual insights hold great promise for personalized treatment. Additionally, our findings highlight the difficulty of formalizing certain aspects of human counterfactual reasoning within SCMs. Returning to our initial example of Alice passing her final exam, it might be straightforward to compute specific counterfactuals that would make the event impossible (e.g., Alice opening a bakery instead of attending college). However, we currently lack the formalism to adequately study such abstract forms of reasoning and their likelihoods. We hope future work will shed more light on this important aspect of causal research.
\section*{Acknowledgements}
JW received support from the German Federal Ministry of Education and Research (BMBF) as part of the project MAC-MERLin (Grant Agreement No. 01IW24007) and from the European Research Council (ERC) Starting Grant CausalEarth under the European Union’s Horizon 2020 research and innovation program (Grant Agreement No. 948112).
\bibliographystyle{unsrt}  
\bibliography{references} 
\appendix
\section{Running Example}
\subsection{The Lorenz System as an ODE} Consider the Lorenz system, a set of three coupled, nonlinear differential equations originally developed to model atmospheric convection. The system is known for its chaotic behavior under certain parameter values and initial conditions. Let $\bm{X}(t) = \left(X_1(t), X_2(t), X_3(t)\right)$ denote the state variables at time $t$, representing the convective fluid velocity, horizontal temperature variation, and vertical temperature variation, respectively. Let $h_{\bm{\theta}}(\bm{X}(t)) = \left(h_{1,\bm{\theta}}(\bm{X}(t)), h_{2,\bm{\theta}}(\bm{X}(t)), h_{3,\bm{\theta}}(\bm{X}(t))\right)$. The system is described by the following ODEs:

\begin{align} 
\begin{cases} \frac{d}{dt} X_1(t) = h_{1, \bm{\theta}}\big(\bm{X}(t)\big) = \sigma \left( X_2(t) - X_1(t) \right), \\
\frac{d}{dt} X_2(t) = h_{2, \bm{\theta}}\big(\bm{X}(t)\big) = X_1(t) \left( \rho - X_3(t) \right) - X_2(t), \\
\frac{d}{dt} X_3(t) = h_{3, \bm{\theta}}\big(\bm{X}(t)\big) = X_1(t) X_2(t) - \beta X_3(t), \\
\left(X_1(0), X_2(0), X_3(0)\right) = \left(X_1^0, X_2^0, X_3^0\right), 
\end{cases} 
\end{align} where $\bm{\theta} = \left( \sigma, \rho, \beta \right)$, with $\sigma, \rho, \beta > 0$ being constants.

This system models the dynamics of convection rolls in the atmosphere and is notable for its chaotic solutions under certain conditions. The functions $h_{1, \bm{\theta}}$, $h_{2, \bm{\theta}}$, and $h_{3, \bm{\theta}}$ represent the rate of change for each state variable.

\subsection{Lorenz system and SSM}
The forward operator $F(\mathbf{X}_{t-1}, \bm{\theta})$ represents the deterministic dynamics of the Lorenz system, modeling the interactions among the convective fluid velocity and the horizontal and vertical temperature variations. We can define it the fourth-order Runge-Kutta, as:
\begin{align}\label{eq:RK4}
    F(\mathbf{X}_{t-1}, \bm{\theta}) = \mathbf{X}_{t-1} + \frac{\Delta}{6} \left( k_{1, \bm{\theta}} + 2k_{2, \bm{\theta}} + 2k_{3, \bm{\theta}} + k_{4, \bm{\theta}} \right).
\end{align}
where \(\Delta \in \mathbb{R}_{>0}\) is the size of the discrete time step and $\boldsymbol{\theta} = (\sigma, \rho, \beta)$. Here, $k_{1, \bm{\theta}}$, $k_{2, \bm{\theta}}$, $k_{3, \bm{\theta}}$ and $k_{4, \bm{\theta}}$ are defined as
\begin{align}
    k_{1, \bm{\theta}} &= \mathbf{h}_{\bm{\theta}}(\mathbf{X}_{t-1}), \\
    k_{2, \bm{\theta}} &= \mathbf{h}_{\bm{\theta}}\left(\mathbf{X}_{t-1} + \frac{\Delta}{2} k_{1, \bm{\theta}}\right), \\
    k_{3, \bm{\theta}} &= \mathbf{h}_{\bm{\theta}}\left(\mathbf{X}_{t-1} + \frac{\Delta}{2} k_{2, \bm{\theta}}\right), \\
    k_{4, \bm{\theta}} &= \mathbf{h}_{\bm{\theta}}\left(\mathbf{X}_{t-1} + \Delta k_{3, \bm{\theta}}\right).
\end{align}
\subsection{Lorenz system as an SCM}
We aim to represent the State Space Model describing the Lorenz system as an SCM. To achieve this, we define the following variables:
\begin{itemize}
    \item \(V^{X_{1,t}}_t \in \mathbb{R} \), \( V^{X_{2,t}}_t  \in \mathbb{R} \), and \( V^{X_{3,t}}_t  \in \mathbb{R} \) denote the true state variables of the Lorenz system at time at time \( t \in \{0, \dots, T\} \).
    \item \(V^{Y_{1,t}}_t \in \mathbb{R} \), \( V^{Y_{2,t}}_t  \in \mathbb{R} \), and \( V^{Y_{3,t}}_t  \in \mathbb{R} \) represent the corresponding noisy observations at time \( t \).
\end{itemize}
For each time point \( t \), we introduce two noise variables.  \( \mathbf{U}_t \in \mathbb{R}^3 \) accounts for process noise and \( \mathbf{W}_t \in \mathbb{R}^3 \) accounts for observational noise. Variables at the initial time \( t=0 \) have no parent dependencies. For \( t > 0 \), the dependencies are structured as follows:
\begin{itemize}
    \item Each observational variable \( V^{Y_{i,t}}_t \) is influenced by the true state variable  \( V^{X_{i,t}}_t \) and the observational noise \( \mathbf{W}_t\).
    \item Each true state variable \( V^{X_{i,t}}_t \) depends on the previous true state variables \( V^{X_{1,t}}_t \), \( V^{X_{2,t}}_t \), and \( V^{X_{3,t}}_t \), as well as the process noise \( \mathbf{U}_t \).
\end{itemize}
The functional relationships are defined by the RK4 functions defined in equation \eqref{eq:RK4}. To incorporate the parameter vector \( \boldsymbol{\theta} = (\sigma, \rho, \beta) \) into the model, we introduce an additional node representing \( \boldsymbol{\theta} \). 

\subsection{Nested Particle Filter}\label{sec:NPF}
\begin{itemize}
    \item Generate $M$ parameter particles $\{\bm{\theta}^{(m)}\}_{1\leq m \leq M}$ from the prior distribution $\pi_0$ and $M\times N$ hidden states particles $\{\mathbf{x}^{(n,m)}\}_{1\leq n\leq N,\;1\leq m\leq M}$ from the prior distribution $\tau_0$.
    \item For each time step $t$, and parameter particle $\bm{\theta}^{(m)}$:
    \begin{itemize}
        \item Propagate the hidden state particles $\{\mathbf{x}^{(n,m)}\}_{1\leq n\leq N}$ using the state equation \eqref{eq:state} as follows:
            \begin{align}
                \mathbf{x}_{t}^{(n,m)} = F_t(\mathbf{x}_{t-1}^{(n,m)}, \bm{\theta}^{(m)}) + \mathbf{U}_t, \quad \mathbf{U}_t \sim \mathcal{N}(\mathbf{0}, \mathbf{R}), \; 1\leq n \leq N.
            \end{align}
        \item Compute the inner filter estimate as a weighted average, with weights calculated based on how well the particles explain the observations:
            \begin{align}
            \mathbf{x}_t^{(m)} = \sum_{n = 1}^{N} w_t^{(n)} \mathbf{x}_{t}^{(n,m)} \quad
                w_t^{(n)} = p(Y_t|X_t^{(n,m)}) 
            \end{align}
    \end{itemize}
    \item Compute the outer filter state and parameter estimate as a weighted average, with weights calculated based on how well the estimated inner estimates explain the observations:
        \begin{align}
            \hat{\mathbf{x}}_t = \sum_{m = 1}^{M} v_t^{(m)} \mathbf{x}_{t}^{(m)} \quad \text{and} \quad \hat{\bm{\theta}} = \sum_{m = 1}^{M} v_t^{(m)} \bm{\theta}^{(m)}, \; \text{with }
            v_t^{(m)} = p(Y_t|X_t^{(m)})
        \end{align}
\end{itemize}
The outer filter directly approximates the marginal posterior of $\boldsymbol{\theta}$, while the combination of the outer and inner filters yields an approximation of the joint posterior of $\mathbf{X}(t)$ and $\boldsymbol{\theta}$. This nested approach is fully recursive, with the computational cost at each time step remaining constant, as the method does not require reprocessing of previous observations. The resampling mechanism ensures that particles representing unlikely parameter values or state trajectories are discarded, focusing computational effort on the most probable estimates.
\subsection{Backwards Smoothing}\label{sec:smoothing}
On the other hand, NPF approaches to track the joint posterior  $(\mathbf{X}(t), \boldsymbol{\theta})$ using only observations up till time $t$. To refine the posterior incorporating future observations as well $\mathbf{Y}_{t:T}$, a backward smoothing layer can be applied recursively for $t = T-1: 1$, in the following steps:
\begin{itemize}
    \item Smoothing for Hidden States:  
    For each outer particle $\bm{\theta}_t^{(m)}$, the hidden state particles $\mathbf{X}_t^{(n,m)}$ are smoothed by adjusting their weights based on the transition probability between consecutive time steps and the future observations:
    \begin{equation}
    \tilde{w}_t^{(n,m)} = w_t^{(n,m)} \sum_{k=1}^{N} p(\mathbf{X}_{t+1}^{(k,m)} \mid \mathbf{X}_t^{(n,m)}, \bm{\theta}_t^{(m)}) \tilde{w}_{t+1}^{(k,m)}
    \end{equation}
    where $\tilde{w}_t^{(n,m)}$ represents the smoothed weights for the hidden state particles, and $p(\mathbf{X}_{t+1}^{(k,m)} \mid \mathbf{X}_t^{(n,m)}, \bm{\theta}_t^{(m)})$ is the transition probability.
\item The smoothed estimated states and parameters are thus calculated: 
\begin{align}
    \hat{\mathbf{x}}_t = \sum_{m = 1}^{M} \tilde{w}_t^{(n,m)} \mathbf{x}_{t}^{(n,m)} \quad \text{and} \quad \hat{\bm{\theta}} = \sum_{m = 1}^{M} \tilde{v}_t^{(m)} \bm{\theta}^{(m)},
\end{align}
where $\tilde{v}_t^{(m)} = 1/M \sum_{m= 1}^{M}\tilde{w}_t^{(n,m)} $ are the smoothed weights.
\end{itemize}
\section{Experiments}
\subsection{Experimental Set-up}\label{appendix:expsetup}
We use different dynamical systems described by an ODE (Section $\ref{subsec:ode}$). Table $\ref{tab:systems}$ summarizes these systems with the corresponding setup. 
We use the state space model (Section $\ref{sec:SSM-SCM}$) to model the evolution of system's state over time.
The forward deterministic pass $F(\mathbf{X}_{t-1}, \bm{\theta})$ in equation $\eqref{subsec:SSM}$ is set to be a RK4 approximation method.

True states and noisy observations of the system are generated using a user-defined initial condition $\mathbf{X}_0$ and true parameters $\bm{\theta}_{true}$. Based on the generated observations, we estimate the hidden states and system parameters using a forward-nested filter followed by a backward-smoothing pass. 
We generate counterfactual trajectories based on interventions on the initial conditions. We abduct the noise posterior based on the estimated states and parameters. The sequences of counterfactual hidden states are generated based on equation \eqref{eq:CSCM1}.

The Number of particles in the filtering process is set to $N = M = 200$. 
Initial particles $\theta^{(m)}$ are generated from a prior distribution $\mathbf{\pi}_0$ that depends on the dynamical system.
Process noise $\mathbf{U}_t$ is sampled from normal distribution with mean $\mathbf{0}$ and variance $a\mathbf{I}$.
Observational noise $\mathbf{V}_t$ is sampled from normal distribution with mean $\mathbf{0}$ and variance $b\mathbf{I}$.
The step size $\Delta$ in RK4 is set to $5 \times 10^{-2}$. 
All experiments were conducted using Matlab R2024b. 

\subsubsection{Dynamical systems}
Table $\ref{tab:systems}$ details the dynamical systems considered in the experiments, mainly Lorenz, R\"{o}ssler, and Logistic Growth systems. For each system, we specify the initial conditions, the 'ground truth' parameters, the prior distributions used in the particle filter (PF), and the regions of chaotic behavior. 

The Lorenz and Rössler systems are classic examples of chaotic dynamical systems, both characterized by a set of three coupled, nonlinear differential equations. The Lorenz system, originally developed to model atmospheric convection, is famous for its chaotic behavior, where small changes in initial conditions can lead to vastly different outcomes—an effect commonly referred to as the "butterfly effect." Similarly, the Rössler system was introduced as a simpler model of chaos and exhibits comparable sensitivity to initial conditions, where even slight perturbations can cause divergent trajectories over time. Both systems serve as important case studies in the field of chaos theory, providing insight into how chaotic dynamics manifest in different mathematical models.

We use the RK4 discrete-time approximation for both systems (see Appendix). Both systems are described by a 3-dimensional state vector $\mathbf{X}_t = (X_{1,t}, X_{2,t}, X_{3,t})$, with parameters $\bm{\theta} = (\sigma, \rho, \beta) = (10, 28, 8/3)$ and $\bm{\theta} = (a, b, c) = (0.2, 0.2, 5.7)$. for Lorenz system. the initial particles are sampled from uniform prior distributions: $\sigma \sim \mathcal{U}(5, 20)$, $\rho \sim \mathcal{U}(15, 50)$, and $\beta \sim \mathcal{U}(1, 8)$. The initial condition is set as $\mathbf{X}_0 = (1, 1, 1)$, and the counterfactual initial condition is defined as $\mathbf{X}^{\text{cf}}_0 = \mathbf{X}_0 + 10^{-4}\mathbf{e}_1$. Similarly, for the Rössler system, the prior distributions for the parameters are given by $a \sim \mathcal{U}(0.1, 0.3)$, $b \sim \mathcal{U}(0.1, 0.3)$, and $c \sim \mathcal{U}(4, 8)$. The initial condition is $\mathbf{X}_0 = (1, 1, 0)$, and the corresponding counterfactual initial condition is $\mathbf{X}^{\text{cf}}_0 = \mathbf{X}_0 + 10^{-4}\mathbf{e}_1$.

In contrast, the logistic growth model serves as a benchmark example of a non-chaotic system. It describes the evolution of a population over time, where growth is initially exponential but eventually stabilizes as the population approaches a carrying capacity. Unlike the Lorenz and Rössler systems, the logistic growth model does not exhibit sensitivity to initial conditions in the same way, making it a suitable reference for studying counterfactual reliability in the absence of chaotic dynamics. The logistic growth process is described by the discrete-time RK4 approximation of the logistic differential equation (Appendix), with parameters $\bm{\theta} = (r, K) = (0.5, 100)$, where $r$ is the intrinsic growth rate and $K$ is the carrying capacity. The parameter priors are given by $r \sim \mathcal{U}(0, 1)$ and $K \sim \mathcal{U}(85, 110)$. The initial condition is set to $X_0 = 10$, and the counterfactual initial condition is $X^{\text{cf}}_0 = X_0 + 10$.
\begin{table}
\centering
\caption{Dynamical systems considered in the experiments, along with their corresponding ground truth parameters, prior distributions, and chaotic behavior regions.}
\begin{tabular}{|c|c|c|c|}
\hline
\textbf{System} & \textbf{Ground Truth $\bm{\theta}$} & \textbf{Prior Distribution (Uniform)} & \textbf{Chaotic Region} \\
\hline
Lorenz System   & $\left(\sigma, \rho, \beta\right)  = \left(10, 28,8/3 \right)$ & \makecell{$\sigma \sim U(5, 15)$ \\ $\rho \sim U(20, 35)$ \\  $\beta \sim U(2, 4)$} & $\rho \gtrsim 24.74$ \\
\hline
Rössler System  & $ \left(a,b,c \right) = \left(0.2, 0.2, 5.7\right)$ & \makecell{$a \sim U(0.1, 0.3)$ \\ $b \sim U(0.1, 0.3)$ \\ $c \sim U(4, 7)$} & $c \gtrsim 5.7$ \\
\hline
Logistic Growth & $\left(r, K \right) = \left(3.9, 1\right)$ & \makecell{$r \sim U(2, 4)$ \\ $K \sim U(0.8, 1.2)$} & Na \\
\hline
\end{tabular}
\label{tab:systems}
\end{table}

\subsubsection{Evaluation metrics}\label{appendix:RMSEt}
We calculate the root mean squared error (RMSE$_t$) over time to evaluate the accuracy of the sampled counterfactual trajectories compared to the projected counterfactual trajectory. This is done by computing the Euclidean distance at each time step in the \( d \)-dimensional phase space. Given \( d \) as the state dimension, \( T \) as the number of time steps, and \( N_{\text{cf}} \) as the number of counterfactual trajectories, the phase space distance between the deterministic counterfactual state \( \mathbf{X}^{\text{cf}}_t \) and the \( i \)-th counterfactual state \( \mathbf{X}_t^{\text{cf}_i} \) at time \(t\) is calculated as:

\[
d(t, i) = \sqrt{ \sum_{k=1}^{d} \left( \mathbf{X}_{k, t}^{\text{cf}_i} - \mathbf{X}_{k, t}^{\text{cf}} \right)^2 }
\]

where $\mathbf{X}_t^{\text{cf}}$ is the true counterfactual state computed using the deterministic ODE and $\mathbf{X}_t^{\text{cf}_i}$ the $i$-th counterfactual generated at time $t$, using equation \eqref{eq:CSCM1}. Here $\mathbf{X}_t^{\text{cf}} = [X_{1, t}^{\text{cf}}, X_{2, t}^{\text{cf}}, \dots, X_{d, t}^{\text{cf}}]$ and $\mathbf{X}_t^{\text{cf}_i} = [X_{1, t}^{\text{cf}_i}, X_{2, t}^{\text{cf}_i}, \dots, X_{d, t}^{\text{cf}_i}]$.
The RMSE over all counterfactual trajectories at time \( t \) is then computed as:
\[
\text{RMSE}_t = \sqrt{\frac{1}{N_{\text{cf}}} \sum_{i=1}^{N_{\text{cf}}} d(t, i) ^2 }
\]
This provides a measure of how far, on average, the counterfactual trajectories deviate from the factual trajectory in the \( d \)-dimensional phase space, and it is calculated at each time step \( t \).
\subsection{Lorenz system}
\begin{align}
    \begin{cases}
         \dfrac{dx_1}{dt} = \sigma\left(x_2 - x_1\right)\vspace{0.2cm}\\
         \dfrac{dx_2}{dt} = x_1\left(\rho - x_3\right) - x_2, \quad \vspace{0.2cm}\\
         \dfrac{dx_3}{dt} = x_1 x_2 - \beta x_3
    \end{cases}
    \quad
    \bm{\theta} = \left( \begin{array}{c}
    \sigma \\
    \rho \\
    \beta
    \end{array} \right)
\end{align}
\subsection{Rossler system}
\begin{align}
    \begin{cases}
         \dfrac{dx_1}{dt} = - x_2 - x_3 \vspace{0.2cm}\\
         \dfrac{dx_2}{dt} = x_1 + ax_2\quad \vspace{0.2cm}\\
         \dfrac{dx_3}{dt} = b + x_3(x_1 - c)
    \end{cases}
    \quad
    \bm{\theta} = \left( \begin{array}{c}
    a \\
    b \\
    c
    \end{array} \right)
\end{align}

\subsection{Full Results}
\begin{figure}[h]
    \centering
    \begin{minipage}[b]{0.45\textwidth}
        \centering
        \includegraphics[width=\textwidth]{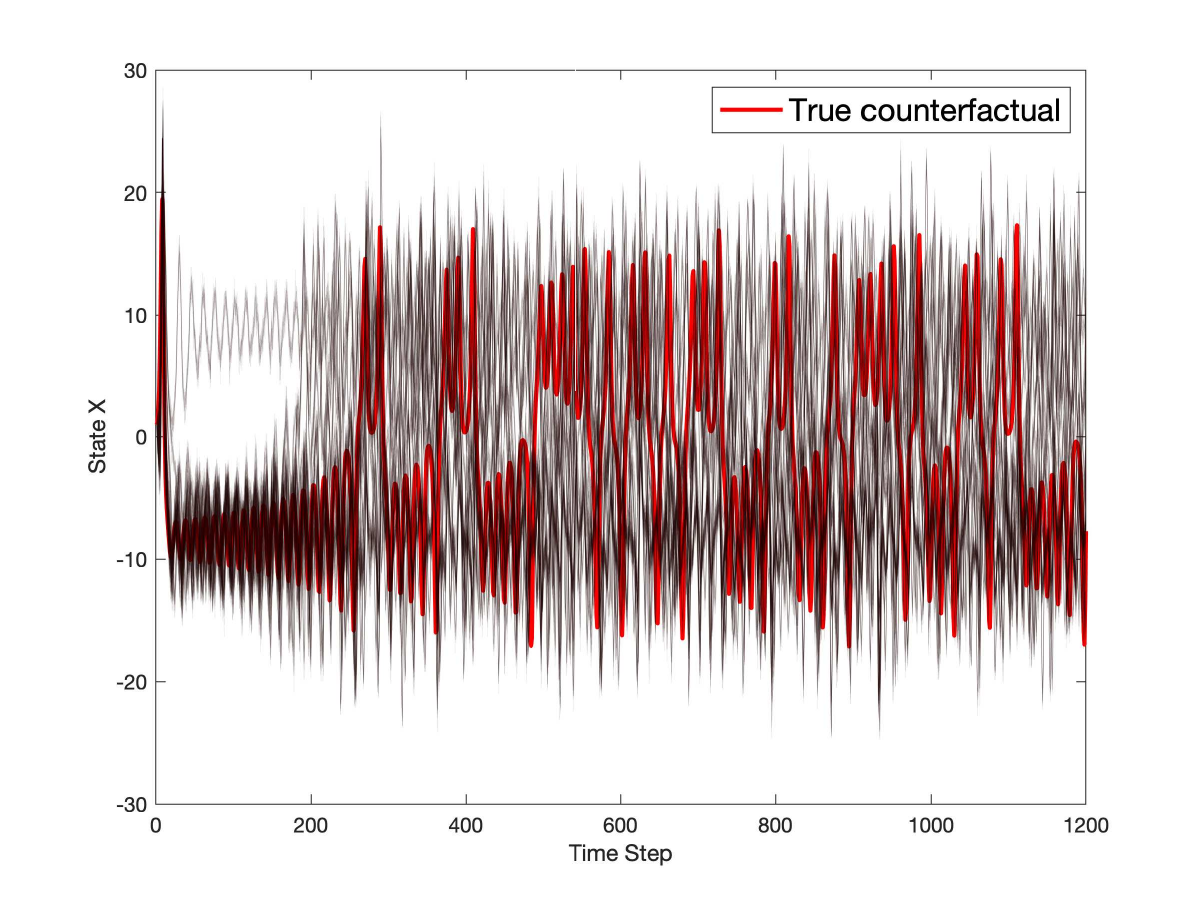} 
    \end{minipage}
    \begin{minipage}[b]{0.45\textwidth}
        \centering
        \includegraphics[width=\textwidth]{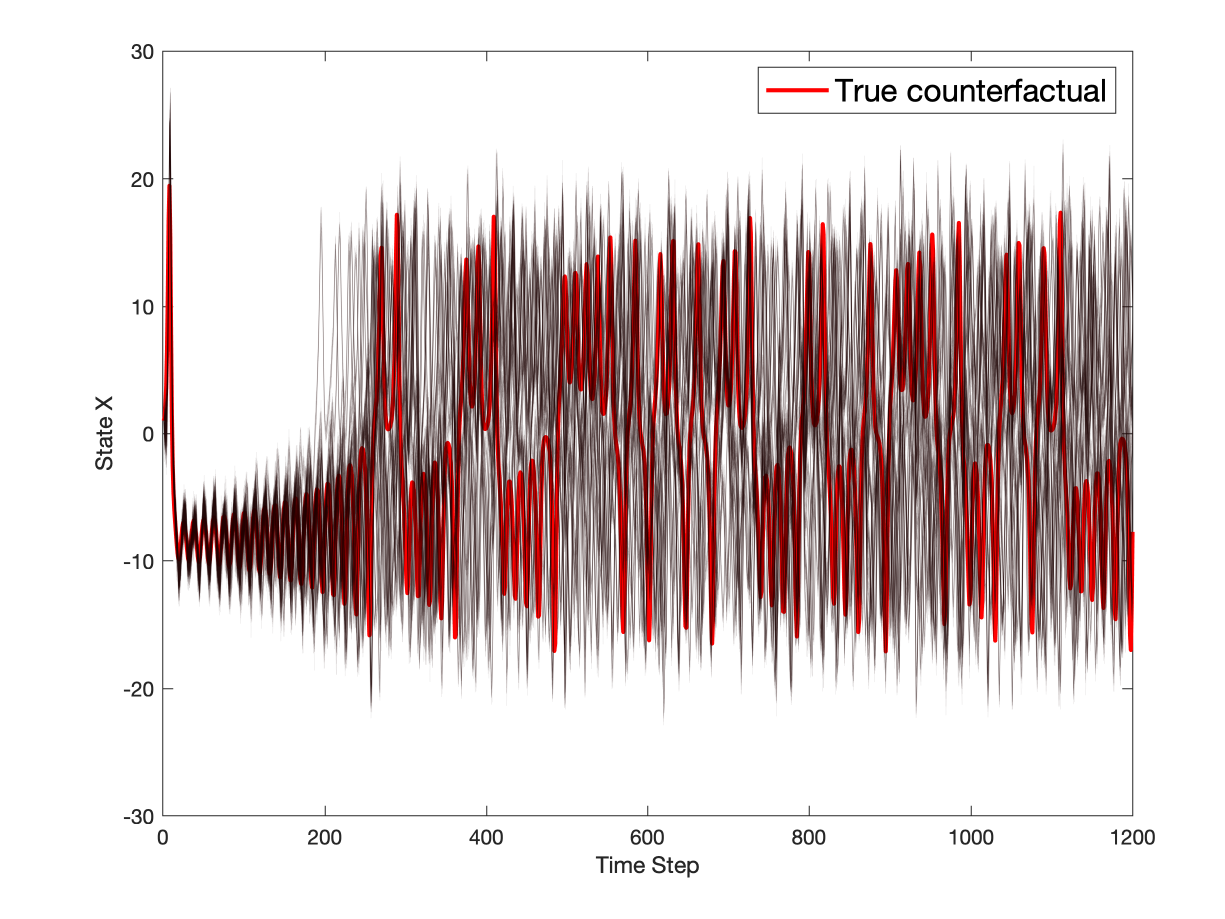} 
    \end{minipage}
    \begin{minipage}[b]{0.45\textwidth}
        \centering
        \includegraphics[width=\textwidth]{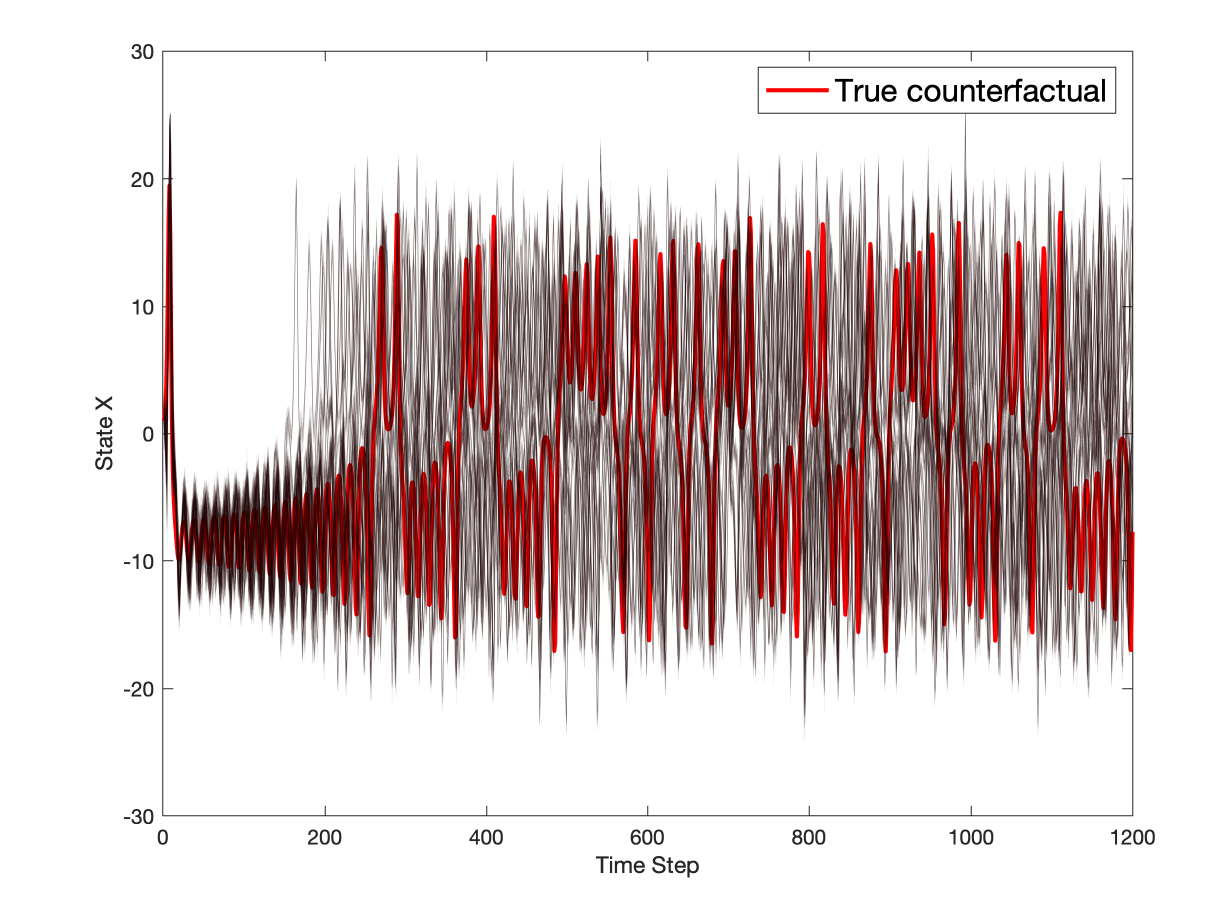} 
    \end{minipage}
        \begin{minipage}[b]{0.45\textwidth}
        \centering
        \includegraphics[width=\textwidth]{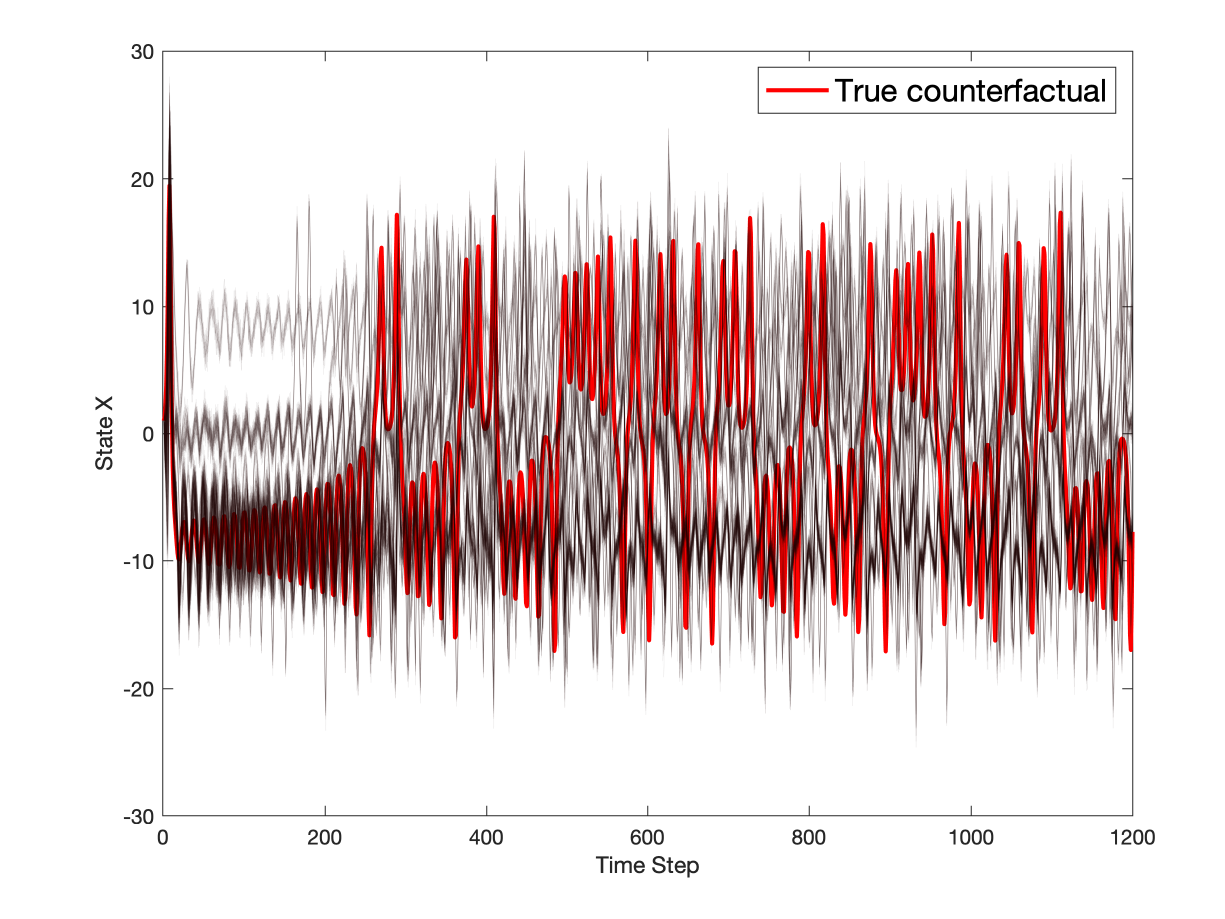} 
    \end{minipage}
    \caption{True counterfactual trajectory and generated counterfactual trajectories (in black) for Lorenz corresponding to $(\sigma_{\mathbf{U}}, \sigma_{\mathbf{W}}) = (4,1)$, $(1,2)$, $(0.01, 4)$, and $(0.01, 9)$ in the step left, top right, bottom left and bottom right, respectively. The counterfactuals are generated by sampling values of the parameters from the posterior distribution.}
    \label{fig:exp03_line_lorenz}
\end{figure}
\begin{figure}[h]
    \centering
    \begin{minipage}[b]{0.45\textwidth}
        \centering
        \includegraphics[width=\textwidth]{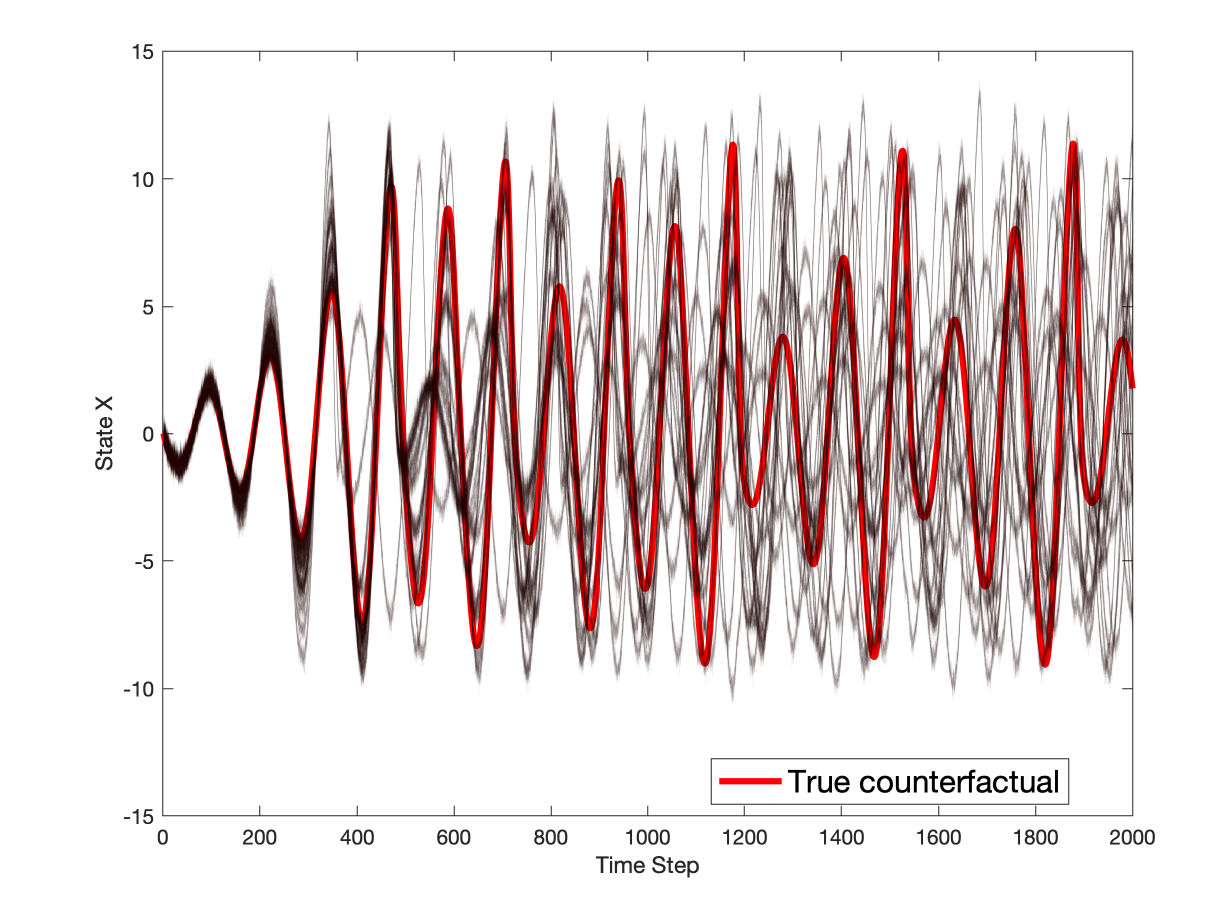} 
    \end{minipage}
    \begin{minipage}[b]{0.45\textwidth}
        \centering
        \includegraphics[width=\textwidth]{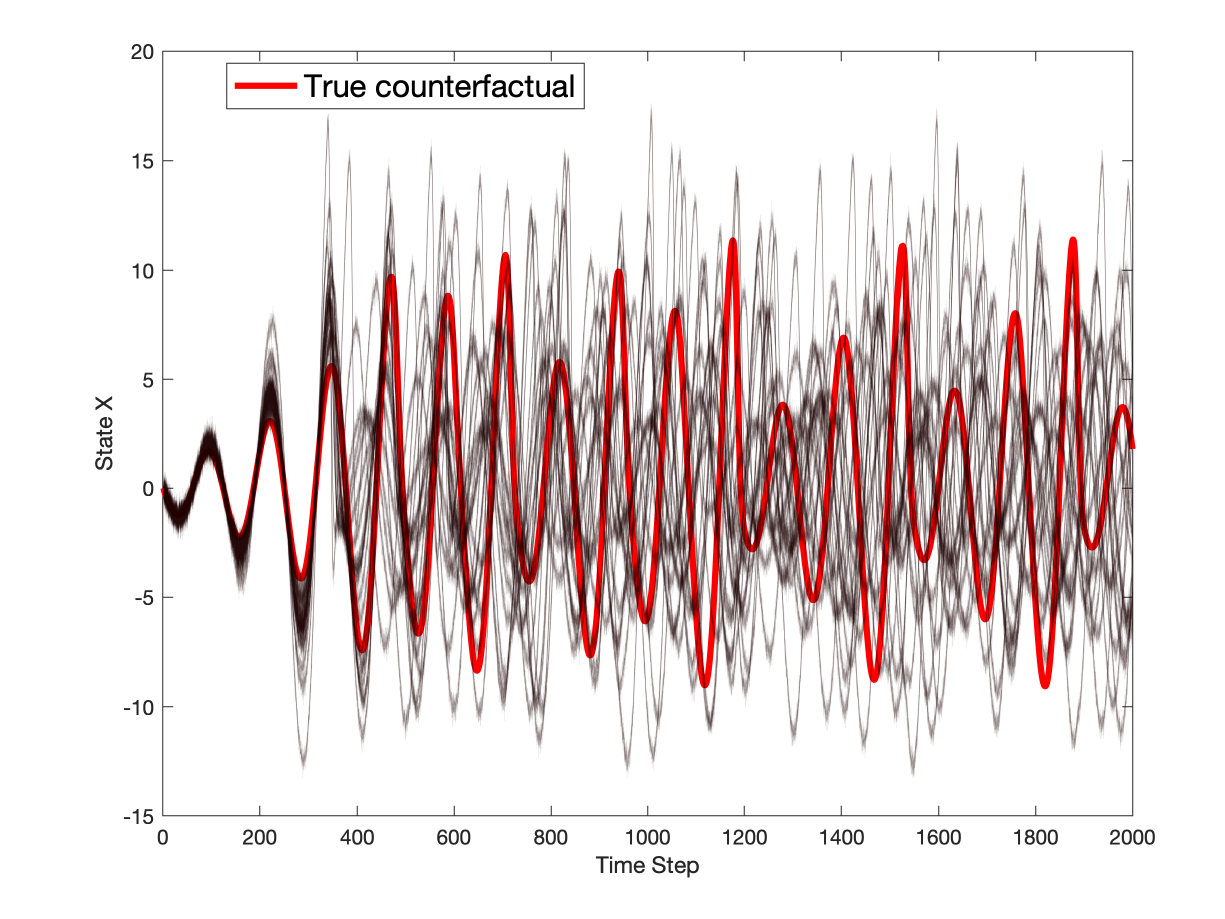} 
    \end{minipage}
    \begin{minipage}[b]{0.45\textwidth}
        \centering
        \includegraphics[width=\textwidth]{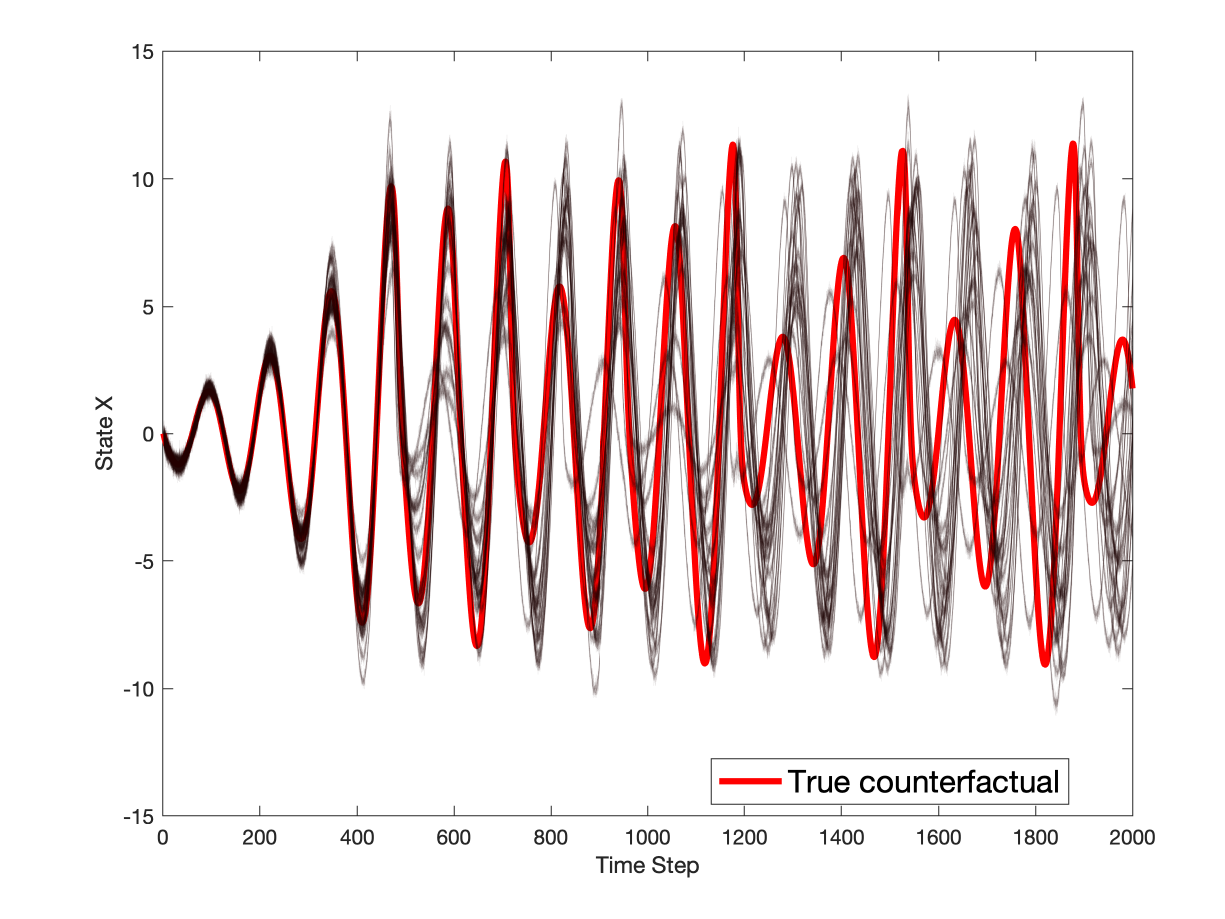} 
    \end{minipage}
        \begin{minipage}[b]{0.45\textwidth}
        \centering
        \includegraphics[width=\textwidth]{figure_09.pdf} 
    \end{minipage}
    \caption{True counterfactual trajectory and generated counterfactual trajectories (in black) for R\"{o}ssler, corresponding to $(\sigma_{\mathbf{U}}, \sigma_{\mathbf{W}}) = (4,1)$, $(1,2)$, $(0.01, 4)$, and $(0.01, 9)$ in the step left, top right, bottom left and bottom right, respectively. The counterfactuals are generated by sampling values of the parameters from the posterior distribution.}
    \label{fig:exp03_line_Rossler}
\end{figure}

\end{document}